# LLM Agents Grounded in Self-Reports Enable General-Purpose Simulation of Individuals

**Authors:** Joon Sung Park*[1], Carolyn Q. Zou[1,2], Jonne Kamphorst*[7], Niles Egan[1], Aaron Shaw[2], Benjamin Mako Hill[3], Carrie Cai[4], Meredith Ringel Morris[5], Percy Liang[1], Robb Willer[6], Michael S. Bernstein[1]

## Affiliations:

[1]Computer Science Department, Stanford University; Stanford, CA, 94305, USA.

[2]Department of Communication Studies, Northwestern University; Evanston, IL, 60208, USA.

[3]Department of Communication, University of Washington; Seattle, WA 98195, USA.

[4]Google DeepMind; Mountain View, CA 94043, USA.

[5]Google DeepMind; Seattle, WA 98195, USA.

[6]Department of Sociology, Stanford University; Stanford, CA, 94305, USA.

[7]CDSP and CEE, Sciences Po; Paris, 75007, France.

*Corresponding authors

**Abstract:**

Machine learning can predict human behavior well when substantial structured data are available for well-defined outcomes. Such models are typically outcome-specific, however, requiring training data for each target outcome, limiting their applicability to new domains. We test whether large language models (LLMs) can relax these requirements by using self-report data to build attitudinal and behavioral simulations, or "generative agents," that can predict responses across outcomes without outcome-specific training data. Using data from a diverse national sample of 1,052 Americans, we built agents from (i) two-hour, semi-structured interviews elicited using the American Voices Project interview schedule, (ii) structured surveys including General Social Survey items and the Big Five personality inventory, or (iii) both sources combined. On held-out General Social Survey items, interview-only, survey-only, and combined agents achieved accuracies equal to 83%, 82%, and 86% of participants' own two-week test-retest consistency benchmark, respectively, compared with 74% for demographics-only agents. Combining interviews and surveys produced the highest accuracy, though gains over either source alone were modest, suggesting that predictive benefits from data begin to asymptote once the model has observed sufficient evidence within a domain. We find that these agents also predict personality traits, economic-game behavior, and experimental responses, while reducing accuracy disparities across racial and ideological groups relative to demographics-only agents. Together, these results show that LLM agents grounded in qualitative or quantitative self-reports can support general-purpose simulation of individuals across outcomes, without requiring task-specific training data.

**Significance statement:**

Researchers increasingly use large language models to simulate human attitudes and behavior, but it remains unclear what information supports accurate predictive simulation. We show that generative agents built from individuals' own in-depth self-reports, either semi-structured qualitative interviews, structured surveys, or both, can predict held-out attitudes, personality traits, economic-game behavior, and experimental responses without task-specific training data. Accuracy varied substantially by agent specification, indicating that sparsely specified agents provide a limited test of LLM-based individual simulation. These findings suggest that building agents from individuals' own self-reports provides a practical route to individual-level simulations that generalize across outcomes and reduce accuracy disparities relative to demographic-only prompting.

General-purpose simulation of human attitudes and behavior – in which simulated individuals respond realistically across social, political, and informational contexts – promises a versatile tool for social and behavioral science *(1-10)*. Realizing these potential applications, however, requires agent architectures that can produce human-like attitudes and behaviors while remaining flexible and applicable across a range of domains, tasks, and settings. Recent machine-learning approaches to predicting human behaviors and outcomes have achieved strong performance in domains with abundant structured data (e.g., *11-13*). Yet these approaches depend on extensive, in-domain structured data that are costly to collect and tailored to a specific outcome, making them difficult to adapt to new contexts or outcomes. These constraints place important limits on person-level predictive modeling for many of the settings social scientists and policymakers care about, where the background information on individuals that is needed is unavailable.

Large language models (LLMs) offer a new path forward because they provide a domain-general interface linking person-specific information and a diverse range of application contexts (14-16). Given a description of an individual, the same underlying model can be queried across many outcomes without the need to train new outcome-specific predictive models *(17)*. The success of this approach depends on providing the model with reliable information about the individuals it is meant to simulate. Social scientists most often elicit background information on individuals via self-reports, using one of two forms: open-ended interviews and structured survey batteries. Each method has distinct advantages and limitations for informing prediction (2). Here, we compare the predictive performance of three distinct approaches to grounding LLM-based generative agents: (a) *interview agents*, prompted via responses to in-depth, semi-structured interviews, collected via an AI, voice-to-voice interview agent developed for this project that deployed the American Voices Project interview schedule; (b) *survey agents*, prompted via responses to survey batteries drawn from the General Social Survey and the Big Five personality inventory; and (c) *survey-plus-interview agents,* prompted with both of these self-report data sources.

We evaluate these generative agents at the relevant unit of analysis, the individual, comparing predicted responses with participants' actual responses across a range of outcomes. Our framework is designed to reduce LLMs' established tendency to rely on demographic stereotypes by grounding predictions in individuals' self-reports *(18-20)*. We assess performance across canonical social-science instruments and decision contexts, including survey measures of attitudes and behaviors, personality inventories, behavioral games, and behavior in randomized controlled trials. Because most human attitudes and behaviors are measured with nontrivial within-person noise (21), raw prediction accuracy is unlikely to reach 100%, and is not directly comparable across different instruments. We therefore benchmark agent performance against each participant's two-week test-retest consistency on the same measure, meaning that a perfect agent response is interpreted as "the agents replicate participants' responses as well as participants replicate their own responses two weeks later." Across evaluations spanning attitudes, personality, and incentivized choices, agents grounded in participants' self-reports – particularly those grounded in both interview and survey responses – are able to predict held-out individual responses substantially better than prompting with demographic data alone, often approaching participants' own test-retest consistency, though performance varies meaningfully by outcome reliability and task structure.

In sum, we show that LLM agents grounded in individuals' own self-reports can predict held-out person-level responses across multiple social-science domains without training outcome-specific models for each task. We also introduce an evaluation framework that benchmarks agent performance against participants' two-week test-retest consistency, enabling comparisons across instruments with different reliability. Moreover, by comparing agents prompted with interview responses, survey responses, or both, we are able to explore which forms of self-report information most improve accuracy of individual-level, LLM-driven simulation. Finally, beyond evaluating the agents themselves, we show that AI-based voice interviews offer a scalable way to collect the rich self-report data needed to ground person-level simulations. Our results support a range of near-term research applications, including the exploration of effect heterogeneity before conducting field work to identify sub-populations for whom an intervention might not be effective, refining measurement tools such as survey questions, experimental designs that combine simulated with human data to increase power *(22),* and pilot testing of surveys and experiments.

## Using Individuals' Self-Reports to Create Generative Agents

To gather in-depth information on individuals, we draw on two long-standing methodological traditions in the social and behavioral sciences. First, structured survey questionnaires provide standardized measures designed to capture the topics social scientists care about, using carefully formulated wording and response categories that support reliable comparison across respondents and, in turn, population-level inference. Surveys are typically constructed to cover a broad set of conceptually distinct constructs efficiently, with items that are intended to tap non-overlapping dimensions of attitudes, beliefs, and behaviors. Second, we draw on in-depth, semi-structured interviews, which excel at capturing the idiosyncratic factors and lived constraints that shape how people think and act. Semi-structured interviews pair pre-specified prompts with adaptive follow-up questions that respond to participants' own answers, allowing respondents to foreground what they view as most consequential for understanding them *(23-25)*. Open-ended narratives can encode multiple features in a single response, linking facts, constraints, and trajectories that would otherwise require numerous closed-form items. For example, the statement "Since my back injury, I can only work part-time" simultaneously conveys health constraints, employment capacity, and anticipated prospects – information that is difficult to recover from standard demographic instruments with predefined response categories. By combining pre-specified prompts with responsive probing, interviews expand what can be measured in the first place, not merely how answers are expressed *(24, 25)*. To scale this semi-structured component, we developed an AI interviewer that autonomously conducts in-depth interviews, pairing a fixed protocol with adaptive, real-time follow-up questions to elicit the kind of open-ended narratives that have previously required trained human interviewers (SM 2).

We recruited a stratified sample ($N$ = 1,052) that approximated U.S. adult distributions on age, gender, race, region, education, and party identification. Each participant completed a voice-to-voice interview in English with the AI interviewer, averaging about two hours and producing transcripts with an average length of 6,491 words per participant (std = 2,541; SM 1). To avoid inadvertently tailoring the interview protocol to our evaluation metrics, we sought an existing interview protocol that aimed for broad topical coverage. We adopted the interview protocol developed for the American Voices Project *(26)*. The semi-structured interview protocol included both fixed questions on a wide range of topics from participants' life stories (e.g., "Tell me the story of your life – from your childhood, to education, to family and relationships, and to any major life events you may have had”) to their views on current societal issues (e.g., “Some people tell us about experiences of arrest – of loved ones, family members, friends, or themselves. How about for you?”; SM 10, Table 7) – and also follow-up questions generated by the AI interviewer in real-time. Its broad scope, and its design independent of any of our evaluation tasks, strengthens results if high performance is achieved.

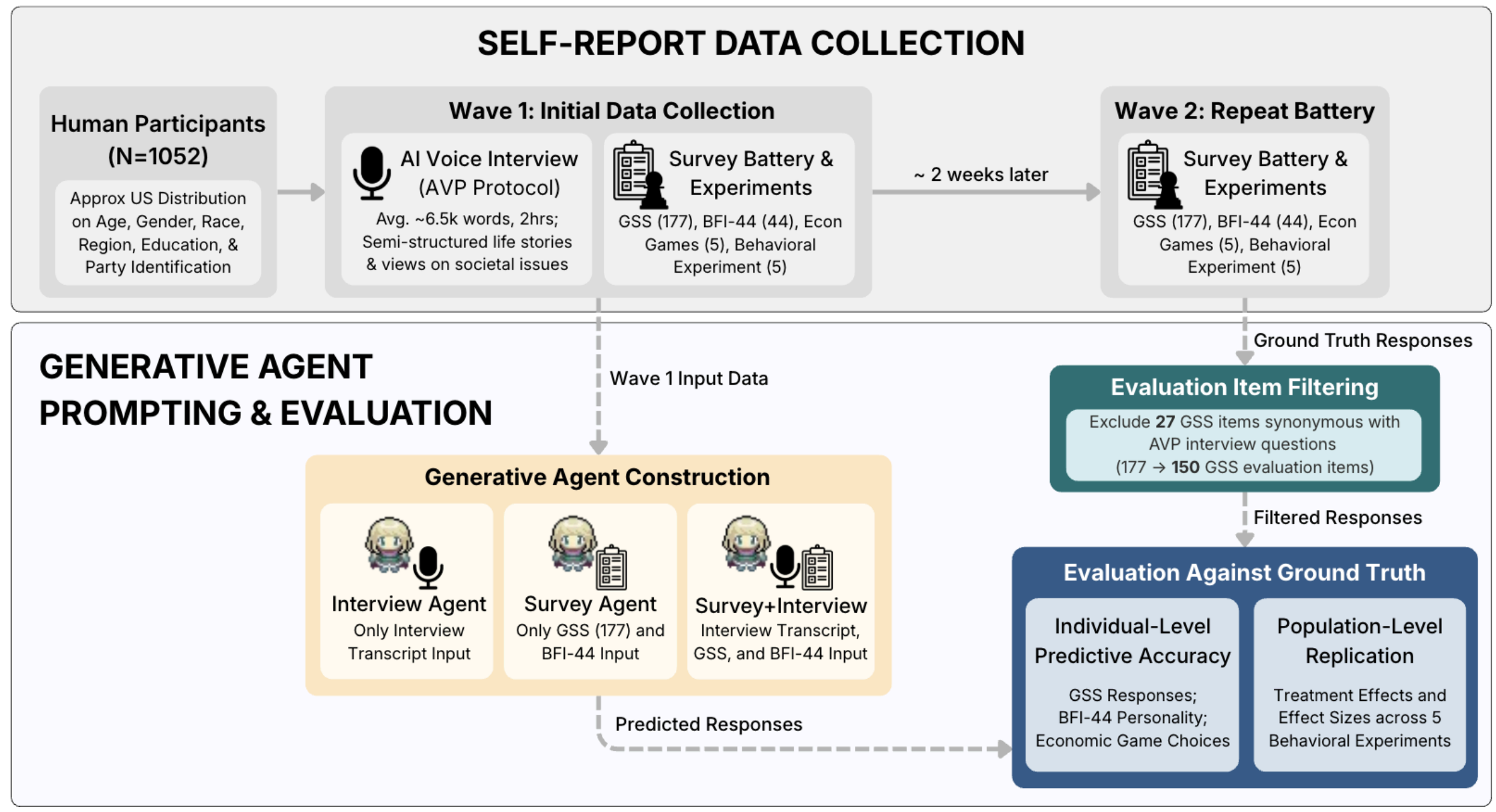

**Figure 1.** We recruit a stratified U.S. sample of 1,052 individuals by age, gender, race, region, education, and party identification. Participants complete a two-hour audio interview with our AI interviewer, followed by surveys and experiments. For each participant, we create three generative agents: interview agents (transcript only), survey agents (structured responses only), and survey-plus-interview agents (both). Agents and participants complete the same surveys and experiments; participants retake them two weeks later. We assess individual-level accuracy by comparing agent responses to participants' Wave 2 responses, normalized by each participant's self-consistency. Solid arrows denote the chronological progression of human participants across experimental waves; dashed arrows denote algorithmic and evaluative data flows.

After the interview, respondents completed the core module of the General Social Survey (GSS; *27*), a widely used survey across the social sciences, and the 44-item Big Five Inventory (BFI-44; *28*), the most commonly used personality battery from psychology. Additionally, we elicited participants’ choices in five well-known behavioral economic games with real stakes: the dictator game, the first and second player Trust Games, public goods game, and prisoner’s dilemma (*29-32*). Finally, we evaluated individual choices in social science experimental settings by collecting participants’ choices in a series of five experiments with control and treatment conditions *(33-37)* drawn from a recent large-scale replication effort *(38)*. This included every text-based experiment for which 1,000 participants would provide adequate statistical power from within this group of replicated experiments, which allows for exploration of potential use of generative agents to simulate experimental replications (SM 4).

To create the generative agents *(17,39)*, we developed an agent architecture that leverages different types of information about respondents: (i) interview agents rely on the semi-structured interviews (SM 3), (ii) survey agents on the structured surveys (the General Social Survey and Big Five personality inventory), and (iii) survey-plus-interview agents combine both sources. When an agent is queried, the relevant input information – either the full interview transcript, the structured survey responses, or both – is injected into the model prompt, instructing the model to imitate the individual based on their self-reported data. Each agent also stores a set of expert reflections: short, model-generated notes written from the perspective of four social-science experts (a psychologist, behavioral economist, political scientist, and demographer) that draw out implications the participant did not state directly. At prediction time the model selects the relevant expert's reflections, appends them to the self-report data, and reasons step by step to an answer (see SI appendix 3 for prompt details). We use GPT-4o for all agent types. Re-running the GSS task on a 50-agent subsample across newer OpenAI models produced very similar accuracy, and a 2025 run of GPT-4o reproduced the 2024 result almost exactly (SI Appendix, section 9). The resulting agents can respond to any textual stimulus, including forced-choice prompts, surveys, and interactional settings.

We evaluated the generative agents' ability to predict their source participants' responses on the four surveys and experiments commonly used across social science disciplines. We used the GSS, Big Five, and a set of economic games to measure the accuracy of the generative agents in predicting individual attitudes, traits, and behaviors, while the replication studies assessed their ability to predict population-level treatment effects and effect sizes in a replication. To measure varying levels in respondents' self-consistency, we asked each participant to complete the four batteries of items twice, two weeks apart. For all evaluations, we rely on the second wave of data collection. To prevent direct answer retrieval for the agents relying on survey data in their input, we exclude from the survey agents' input the questions that we evaluate on for the GSS, and for the Big 5 all question-answer pairs from the same category as the question being predicted (categories were defined by the creators of each instrument). In addition, we implemented a procedure to flag and remove GSS questions from the evaluation set that overlapped substantively with either other GSS questions or interview questions in the American Voices Project (details in Materials and Methods). The result of this process was an input for each agent that was confirmed distinct from the evaluation task. Our metrics and core analyses using the interview-based, demographic, and persona agents were pre-registered (SM 5).[1]

A key methodological benefit of simulating specific individuals is the ability to evaluate our architecture by comparing how accurately each agent replicates the attitudes and behaviors of its source individual. For the GSS, where responses are categorical, we measure accuracy and correlation based on whether the agent selects the same survey response as the individual. For the BFI-44 and economic games, which involve continuous responses, we assess the correlation and, in the SM, mean absolute error (MAE). Since individuals often exhibit inconsistency or drift in their responses over time in both survey and behavioral studies *(11,23, 40)*, we use participants' own attitudinal and behavioral consistency as a normalization factor: the probability of accurately simulating an individual's attitudes or behaviors depends on how consistent those attitudes and behaviors are over time.

Our main dependent variable is *normalized accuracy*, calculated as the agent's accuracy in predicting the individual's responses divided by the individual's own replication accuracy. A normalized accuracy of 1 indicates that the generative agent predicts the individual's responses as accurately as the individual replicates their own responses two weeks later. For continuous outcomes, we calculate *normalized correlations* instead.

To assess the contribution of interviews, surveys, or both sources combined, we compared the performance of these generative agents with two baselines that replace this information with alternative forms of description, reflecting how language models have been used to proxy human behaviors in prior studies: one using demographic attributes *(1, 2)*, and the other using a paragraph summarizing the target person's profile *(39)*. For the 'demographic-based' generative agents, we used participants' responses to GSS questions to capture individuals' age, gender, race, and political ideology – demographic attributes commonly used in previous studies *(2)*. For the "persona-based" generative agents, we asked participants to write a brief paragraph about themselves after the interview, including their personal background, personality, and demographic details, similar to material used in prior work *(39)*.

[1] Pre-registration materials: https://osf.io/mexkf/?view_only=375fe67b9a3e48afa7c3684c9d344da4

## Predicting Individuals' Attitudes and Behaviors

Self-report-grounded agents generally outperformed demographic and persona baselines. Our main results are shown in Figure 2. Our evaluation focused on 150 core GSS questions, which we used to establish a benchmark for measuring the agents' predictive accuracy. Each question had an average of 3.31 response options (std = 1.58), yielding a random chance prediction accuracy of .30.

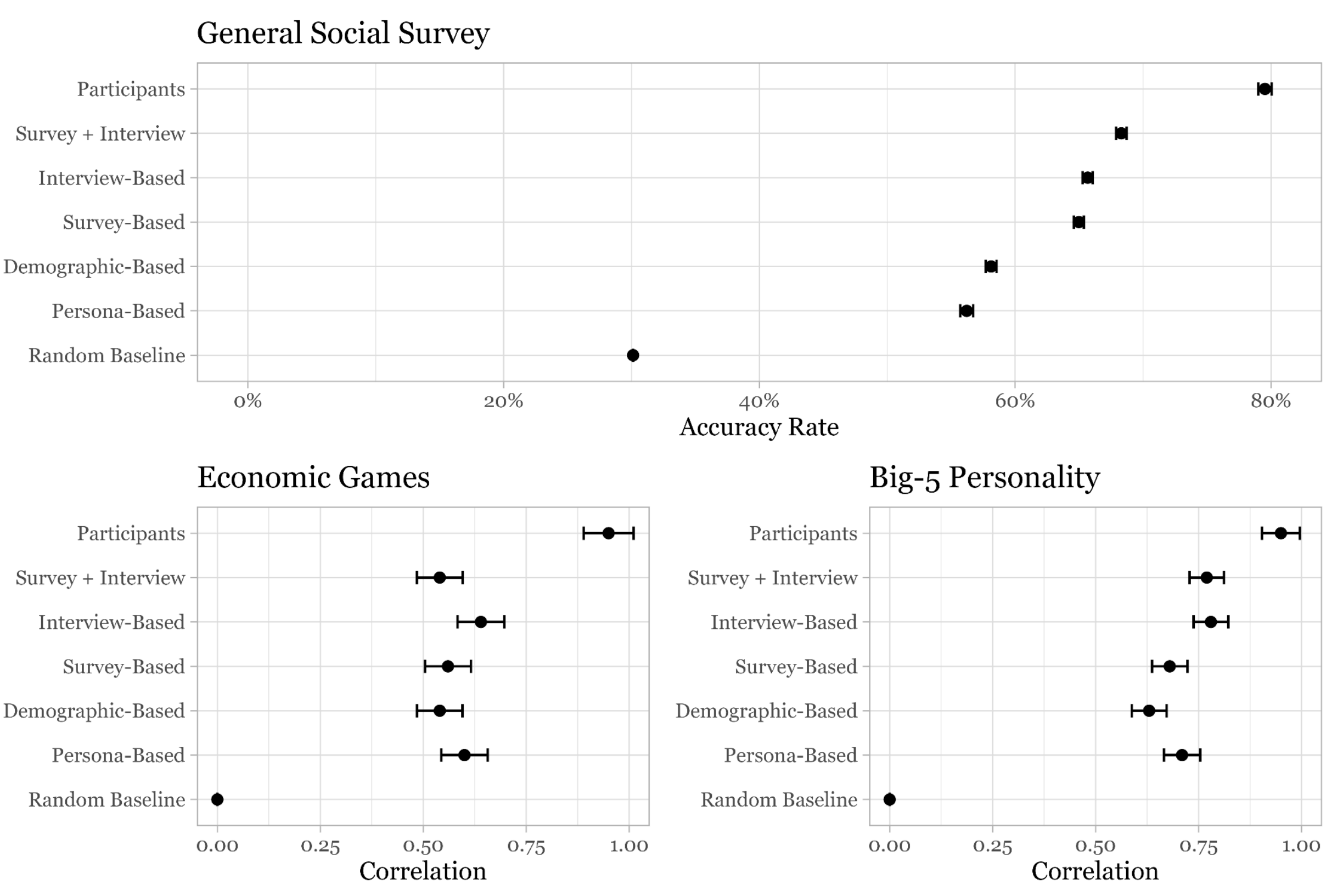

**Figure 2.** Generative agents' and participants' predictive performance and 95% confidence intervals. For the General Social Survey (GSS), accuracy is reported due to its categorical response types, while the Big Five personality traits and economic games report correlations because of the numerical response type. The non-normalized results are reported in the figure.

For the GSS, the interview-based generative agents predicted participants' responses with an average normalized accuracy of 0.83 (std = 0.11), calculated from a raw accuracy of 65.67% (std = 6.51) divided by participants' self-consistency of 79.53% (std = 8.65). The survey agents had an average normalized accuracy of 0.82 (std = 0.11), while the survey-plus-interview agents scored 0.86 (std = 0.09). These three agent types significantly outperformed both demographic-based and persona-based agents (Figure 2). The demographic-based generative agents achieved a normalized accuracy of 0.74 (std = 0.12), while persona-based agents reached 0.71 (std = 0.12). A one-way ANOVA across all five agent types confirmed significant differences in GSS accuracy ($F(4, 5255) = 577.83$, $p < .001$). Post-hoc Tukey comparisons showed that interview, survey, and survey-plus-interview agents each significantly outperformed both demographic- and persona-based agents (all $p < .001$). Interview and survey agents did not differ from one another ($p = .16$), while survey-plus-interview agents significantly outperformed both interview-only and survey-only agents (both $p < .001$).

The second component of our evaluation focused on predicting participants' Big Five personality traits using the BFI-44, which assesses five personality dimensions: openness, conscientiousness, extraversion, agreeableness, and neuroticism *(28)*. Each dimension is calculated from eight to ten Likert scale questions. Our generative agents predicted participants' responses to the individual items, which were then used to compute the predicted aggregate scores for each personality dimension. Because these are continuous measures, we calculated correlation coefficients and normalized correlations, which again divide the agents' correlation by the participants' correlation with their own responses two weeks later. For the Big Five, the interview-based generative agents achieved a normalized correlation of 0.80 (std = 1.88), based on a raw correlation of $r = 0.78$ (std = 0.70) divided by the 'replication correlation' (the correlation between participants' wave 1 and wave 2 answers) of 0.95 (std = 0.76). As with the GSS, the interview-based generative agents outperformed both demographic-based (normalized correlation = 0.61) and persona-based (normalized correlation = 0.75) agents. Survey agents performed comparatively worse than for the GSS (normalized correlation = 0.65), and survey-plus-interview agents were similar to interview only (normalized correlation = 0.77). A one-way ANOVA comparing correlations across all five agent types confirmed significant differences in Big Five correlation ($F(4, 5255) = 26.66$, $p < 0.001$). Interview-only and survey-plus-interview agents significantly outperformed both baselines (all $p < 0.001$) and also outperformed survey-only agents (both $p < 0.001$). Survey-only agents, by contrast, did not significantly differ from either baseline (vs. demographic $p = 0.21$; vs. persona $p = 0.44$), and combining sources did not improve on interview-only agents (survey-plus-interview vs. interview, $p = 0.99$).

The third component involved participants' decisions in five well-known economic games, designed to elicit participants' behaviors in decision-making contexts with real stakes. These included the Dictator Game, the first and second player Trust Games, a Public Goods Game, and the Prisoner's Dilemma *(29-32)*. To ensure genuine engagement, participants were offered monetary incentives. Participants' responses for each game were standardized to scale from 0 to 1. We then compared the generative agents' predicted values to the actual values obtained from participants. Since the outcomes in these economic games are continuous measures, we calculated correlation coefficients and normalized correlations. On average, the interview-based generative agents achieved a normalized correlation of 0.66 (std = 2.83). Survey agents achieved a normalized correlation of 0.38 (std = 3.01), survey-plus-interview agents 0.49 (std = 2.79), demographic agents 0.48 (std = 2.90 ), and persona agents 0.57 (std = 2.84 ). A one-way ANOVA comparing correlations across all five agent types found no significant differences in economic-game performance ($F(4, 5255) = 1.63$, $p = 0.16$), and no pairwise contrast was significant.

In exploratory analyses, we conducted additional tests by ablating portions of the interview-based generative agents' interviews to examine the impact of interview content volume and style (see SI Table 6). First, we randomly removed 80% of the interview transcript (equivalent to removing 96 minutes of the 120-minute interview) and found that the interview-based agents still achieved an average normalized accuracy of 0.79 (std = 0.11) on the GSS and normalized correlation of 0.73 (std = 0.74) on Big Five. Second, to investigate how much the predictive power of interviews came from linguistic cues embedded in the language used by interviewees or the informational content of the interviews, we created "interview-summary" generative agents by prompting the large language model to convert interview transcripts into bullet-pointed summaries of key response pairs, capturing the factual content while removing the original linguistic features. These agents achieved a normalized accuracy of 0.81 (std = 0.13) on the GSS and a normalized correlation of 0.70 (std = 0.67) on Big Five.

## Predicting Experimental Replications

Participants took part in five social science experiments to assess whether generative agents can predict treatment effects in experimental settings commonly used by social scientists. These were drawn from a collection of published studies included in a large-scale replication effort (*33-38*; SM 4), including investigations of how perceived intent affects blame assignment (*33*) and how fairness influences emotional responses (*34*). Both human participants in our work and generative agents completed all five studies, with *p*-values and treatment effect sizes calculated using the statistical methods from the original studies. While our five-study design is underpowered for comparing the predicted and replicated effect sizes, the effect sizes estimated from the different types of generative agents were larger than and highly correlated with those of the participants (ranging from $r = 0.91$ to $r = 0.99$), as shown in Table 1. The overall pattern is similar to that observed previously *(1)*.

| | **Human replication** | | **Agent prediction** | | | | | | | | | |
|---|---|---|---|---|---|---|---|---|---|---|---|---|
| | *Participants* | | *Interview* | | *Survey* | | *Survey + interview* | | *Demographic* | | *Persona* | |
| **Replication Studies** | ***p*** | **Effect size** | ***p*** | **Effect size** | ***p*** | **Effect size** | ***p*** | **Effect size** | ***p*** | **Effect size** | ***p*** | **Effect size** |
| **Ames & Fiske 2015** | *** | 8.03 | *** | 12.59 | *** | 21.22 | *** | 17.35 | *** | 13.43 | *** | 10.30 |
| **Cooney et al. 2016** | *** | 0.40 | *** | 1.34 | *** | 1.61 | *** | 1.43 | *** | 1.37 | *** | 1.27 |
| **Halevy & Halali 2015** | *** | 0.93 | *** | 2.98 | *** | 3.95 | *** | 3.06 | *** | 4.22 | *** | 3.35 |
| **Rai et al. 2017** | | 0.06 | | 0.094 | * | 0.16 | | 0.09 | *** | 0.21 | | 0.078 |
| **Schilke et al. 2015** | *** | 0.36 | *** | 2.97 | *** | 9.64 | *** | 3.18 | *** | 5.52 | *** | 3.74 |
| ***Effect size correlation w/ human rep.*** | | | | **Correlation** $r = 0.98$ 95% CI [0.75, 1.00] | | **Correlation** $r = 0.91$ 95% CI [0.15, 0.99] | | **Correlation** $r = 0.99$ 95% CI [0.87, 1.00] | | **Correlation** $r = 0.93$ 95% CI [0.26, 1.00] | | **Correlation** $r = 0.94$ 95% CI [0.38, 1.00] |

**Table 1.** Results of replication studies by human participants and generative agents. We report the p-values (***: < 0.001, **: < 0.01, *: < 0.05) and effect sizes.

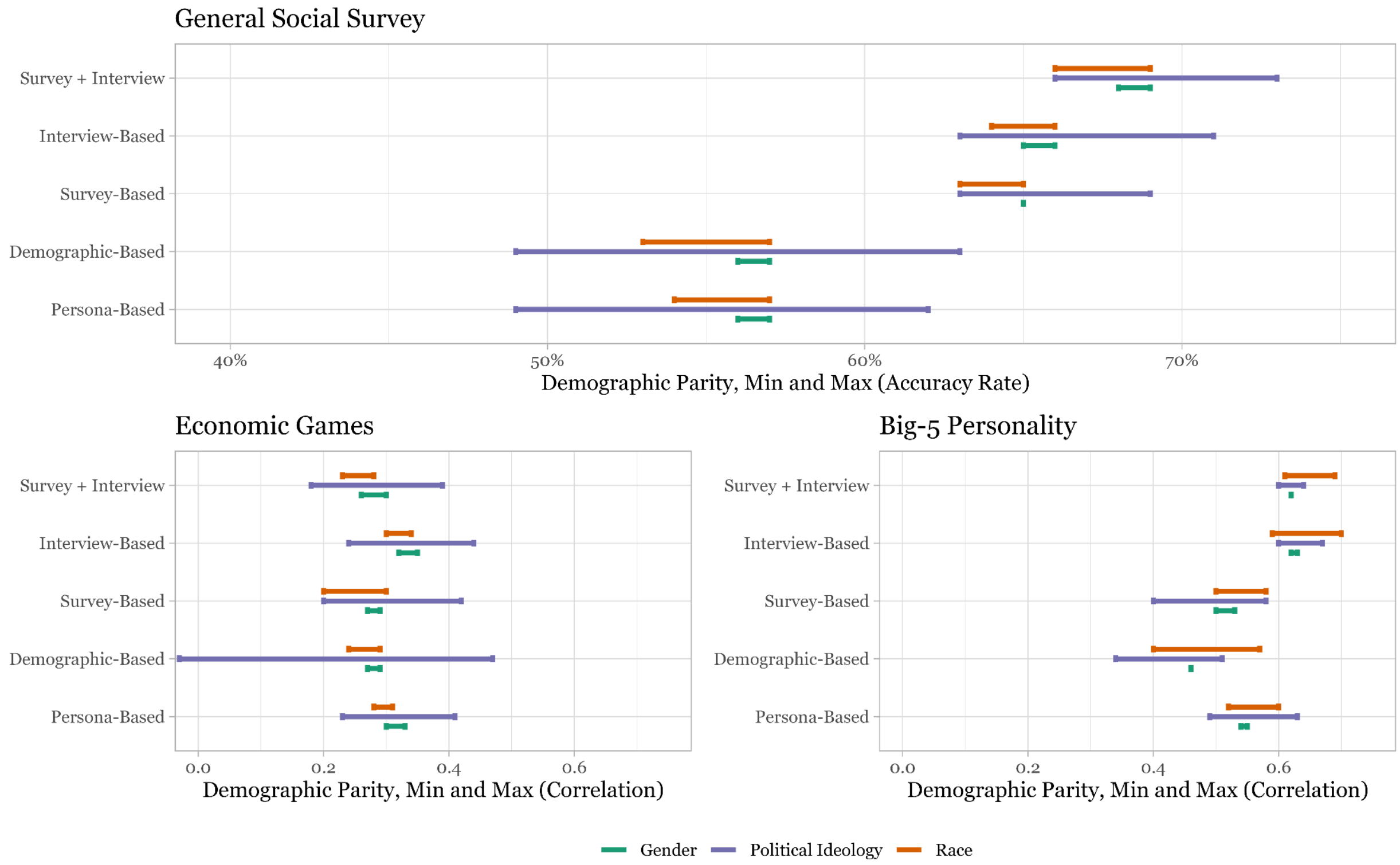

**Figure 3.** Difference in performance between the best-performing and worst-performing demographic groups (DPD), as captured by the highest and lowest point of each bar, for generative agents across political ideology, race, and gender subgroups on three tasks: GSS (reported in percentages), Big Five Personality Inventory, and the economic games (in correlation coefficients). Generative agents using interviews or survey data generally show lower DPDs compared to those using demographic information or persona descriptions, suggesting that interview and survey based generative agents mitigate bias more effectively. Gender-based DPDs remain relatively low and consistent across all agent types.

## Richer Background Data Reduces Bias in Generative Agent Accuracy

AI systems less accurately predict the attitudes and behaviors of members of underrepresented populations *(41)*. To address this concern, we conducted a subgroup analysis focusing on political ideology, race, and gender – dimensions of particular interest in past research *(1, 2, 18, 42, 43)*. We aimed to assess whether the detailed self-report data in interviews and surveys could mitigate biases compared to methods using demographic prompts, which exhibited stereotyping in prior research *(18, 42, 43)*. We quantified group-level bias using the Demographic Parity Difference (DPD), which measures the difference in performance between the best-performing and worst-performing demographic groups *(44, 45)*. For the GSS, we report DPD in percentages; for Big Five and economic games, in correlation coefficients. Subgroups were defined by participants' responses to GSS items (details in SM 5). Figure 3 reports these results.

Agents that rely on the survey and/or interview information generally reduced DPD, compared to demographic-based and persona agents. For political ideology in the GSS, DPD fell from 13.75% with demographic-based agents to 8.60% with interview-based agents, 6.22% for survey agents, and 7.09% for survey-plus-interview agents. In the Big Five personality task, DPD was 0.166 (demographic), 0.063 (interview), 0.176 (survey), and 0.048 for survey-plus-interview agents. Similarly, in economic games, DPD decreased from 0.498 (demographic) to 0.194 (interview), with survey and survey-plus-interview agents scores being higher (0.216 and 0.214, respectively). Racial subgroup DPD also showed reductions: in the GSS, DPD declined from 3.39% (demographic) to 2.22% (interview) and further to 1.95% with survey agents, while survey-plus-interview agents achieved 2.49%. For the Big Five, DPD decreased from 0.173 (demographic) to 0.111 (interview), 0.088 (survey), and 0.081 (survey-plus-interview); in economic games, from 0.043 (demographic) to 0.040 (interview), 0.100 (survey), and 0.046 (survey-plus-interview). Gender-based DPD remained uniformly low across all agent types and tasks, with values typically under 1%, likely due to its already low level of discrepancy.

**Discussion**

Across a diverse national sample, LLM-based generative agents simulated specific individuals substantially more accurately when grounded in people's own self-reports - semi-structured interviews, structured surveys, or both - than when prompted only with demographic or persona descriptions, the dominant approach in prior work (1,2). These gains were not only relative. On the GSS, self-report agents recovered 82 to 86 percent of participants' own two-week test-retest consistency, the practical ceiling for predicting any one person's responses. The advantage over demographic and persona baselines also held for personality traits, and the same agents tracked the direction and relative size of effects across five experimental replications. For the incentivized economic games, all agent types performed similarly and none significantly beat the demographic baseline, so we treat the games as a boundary case rather than a domain where self-reports clearly help. Across these outcomes, self-report agents generally narrowed, though did not fully eliminate, accuracy disparities across racial and ideological groups relative to demographics-only agents. The central contribution of this work is thus to show that rich individual self-reports provide a practical, general-purpose foundation for simulating specific people in settings where structured outcome data are unavailable. Realizing this at scale also required a methodological advance: we designed and fielded an AI-driven, voice-to-voice interviewer that conducted in-depth, semi-structured interviews with a large and diverse national sample, a form of data collection that has previously relied on large teams of human interviewers.

This work has several limitations, each pointing to a direction for future research. First, we do not yet fully understand why grounding agents in self-reports is so predictive, and the limits of this prediction. There is an evolving literature identifying errors and biases in LLM behavioral simulation (46-48). In our work, interview-only agents performed nearly as well as agents combining both - the modest gains from combining them pointing to diminishing returns once a person is described in sufficient depth - even though the two sources draw on different information: structured surveys provide standardized, high-signal measures of attitudes and traits, while semi-structured interviews capture idiosyncratic facts, constraints, life trajectories, and self-understandings that are difficult to encode in fixed response categories. We hypothesize that, once the model has construct-relevant information, further details do not dramatically shift its predictions. Appendix S8 offers a first answer to how the model translates information from the interviews into accurate predictions, identifying at least two processes at work in interview-based agents: 'direct retrieval', in which the model locates an answer already embedded in a transcript even when the original question was on a different topic, and 'inference', in which the model combines what a participant reports with its general knowledge of the world to deduce a likely answer that is never stated. For example, a participant who mentions being enrolled in school provides no direct answer to a survey item about having a workplace supervisor, yet the model can infer that a full-time student is unlikely to be employed and therefore unlikely to report one. It is important to note here that the point is not that interviews restate survey questions - we ensured duplicated questions were removed (SI Appendix, sections S7) - it is that the model recovers answers from the open-ended narratives.

Second, our evaluation centers on individual-level accuracy and does not establish whether the agents reproduce the correlational structure of attitudes across a sample. An agent can predict a given respondent's answers well yet still misrepresent how attitudes covary at the population level. Indeed, prior work finds that LLM-generated survey responses can show inter-item correlations stronger than those in human data (49). Future work should test whether interview, survey, and survey-plus-interview agents recover sample-level correlational structure, not only individual responses. Third, our experimental-replication analysis, while suggestive, is limited by the small number of studies: agent-predicted effect sizes were highly correlated with human effect sizes across the five experiments, but the design is underpowered for strong conclusions about average treatment-effect prediction. Fourth, we evaluated a single model family. Results were stable across OpenAI model tiers and across a year of model updates (SI Appendix, section 9), but we did not test open-weight models such as Llama or Qwen, and whether the architecture transfers to them is an open question. Finally, our data were collected entirely through English-language surveys and interviews, leaving open the question whether self-report grounding is equally effective in other linguistic and cultural contexts. The results are promising in this regard, as demographic-only prompting has been shown to perform worse in cultural contexts more distant from the US, and providing more information about the people being simulated may be a solution (50).

Despite these limitations, the findings reframe what accurate simulation of people requires. As LLMs are increasingly used as stand-ins for human respondents, our results indicate that the quality of individual simulation depends far less on model scale or synthetic persona engineering than on the depth and reliability of the data an agent is built from.

## Materials and Methods Summary

We contracted with the survey firm Bovitz *(51)* to recruit a target sample of approximately 1,000 U.S. adults, with composition matched to nationally representative census quotas on age, gender, race, region, education, and party identification. Comparing the achieved sample against these targets (Table S9), recruitment tracked closely for race and ethnicity and for the youngest age groups, but the sample is slightly more educated than the population (we under-recruited respondents without a high school degree, 2.3% versus a 10.6% target), somewhat more female (56.4% versus 50.9%), more Democratic (35.3% versus 27.0%), and over-represents the Midwest (26.8% vs 20.9%) while under-representing the South (30.2% versus 38.0%). The final analytic sample comprised 1,052 participants who completed both survey waves, ending up with more respondents than we targeted because we overestimated the attrition rate between wave 1 and 2.

Each participant completed, in sequence, a voice-to-voice interview, a battery of survey questions, and a series of real-stakes behavioral economic games (29-32). To collect the interviews at scale, we designed and fielded an AI-driven interviewer that administered the American Voices Project protocol, a semi-structured interview script, posing fixed questions and generating adaptive follow-up questions in real time. We used OpenAI's Audio model, a text-to-speech model that generates voice audio from textual input, to read respondents the interview questions. Participant speech was then transcribed using Whisper, and a GPT-4o model decided whether to move on to the next question or ask a follow-up (see SM 2 for the full architecture). The survey battery comprised the General Social Survey (GSS) and the 44-item Big Five Inventory (BFI-44): from the GSS we used all "core" module questions except conditional items that depend on answers to other questions and those with more than 25 response options, yielding 177 questions alongside six numerical items, and we administered the BFI-44 in full. The behavioral games - a dictator game, a trust game (first- and second-mover roles), a prisoner's dilemma, and a public goods game - gave participants the opportunity to earn additional income. The median response time in the survey components for wave 1 was 55 minutes (*mean*=152, std=727) and 50 minutes for wave 2 (*mean*=120, std=427).

Participants were then randomly assigned to one condition in each of five experiments, which we selected from a recent large-scale replication effort (38) using three inclusion criteria: the study had to be interpretable to a language model using only text or images; the replication effort's power analysis had to indicate that effects would be detectable with 1,000 or fewer participants; and, to maximize the number of studies we could include, we preferred those requiring shorter participation time. The selected studies (33-37) spanned a diversity of phenomena, including the influence of perceived intent on judgments of harm, emotional reactions to actions perceived to be unfair, the perceived benefits of conflict intervention, the willingness to harm dehumanized others, and how the truster's power influences trust. Approximately two weeks later, participants completed a follow-up wave re-administering all measures except the interview, which we use to estimate each participant's test-retest consistency, our benchmark for predictive accuracy. We report the agent-comparison ANOVAs on the correlation and accuracy metrics, secondary metrics specified in our pre-analysis plan, because they make comparisons between outcome types possible; the pre-registered mean absolute error (MAE) scores are reported in SI Table 8, and the conclusions remain unchanged when using this metric.

To rule out the possibility that interview-based agents performed well on the GSS simply because some interview questions closely resembled GSS items - in which case an agent could reproduce an answer it had effectively already been given rather than genuinely simulate the participant - we ran a robustness check screening all 54,694 possible pairings of GSS items and interview questions for duplicate and near-duplicate wording, first with a GPT-4.1 classifier (see SI section 7 for how the classifier was prompted and validated) and then by human review. The procedure flagged 27 GSS items, which we removed. Applying the same procedure to identify GSS questions synonymous with one another returned none.

To balance scientific value with privacy, we provide open access to aggregated responses to the fixed survey instruments used to build the survey-based agents (e.g., the GSS and Big Five Inventory). The code for generating all the different agents is available in an open-source repository here: https://osf.io/t6g7k/files/osfstorage The code to create the interview agents, including the AI interviewer, can be found here: https://github.com/joonspk-research/generative_agent. Researchers interested in constructing agents from their own data can use these resources.

**Acknowledgements**

We thank Huanxing Chen, Garbo Chung, David Grusky, Luke Hewitt, and Chrystal Redekopp for their contributions to the project. This work was supported by Google, Toyota Research Institute, Ford, Stanford's Institute for Human-Centered Artificial Intelligence, American Express, AXA, Banco Itaú, Hanwha, and SCBX.

**Contribution Statement**

J.S.P., M.S.B., and P.L. conceptualized the project and developed the technology. J.S.P., M.S.B., R.W., A.S. and B.M.H. developed the participant elicitation and evaluation methodologies. J.S.P. and C.Q.Z. engineered the technology, including the AI interviewer and simulation system. J.S.P., J.K., C.Q.Z., and N.E. collected the data and conducted the formal analyses. J.S.P. led the investigation. J.S.P., R.W., M.S.B. and J.K. wrote the article. A.S., B.M.H., P.L., M.R.M. and C.C. reviewed and edited the manuscript. M.S.B., P.L., R.W., A.S., B.M.H., and M.R.M. supervised the research.

**Replication materials**

Replication files can be found on the OSF repository: https://osf.io/t6g7k/files/osfstorage

Supplementary Materials for

# LLM Agents Grounded in Self-Reports Enable General-Purpose Simulation of Individuals

Joon Sung Park, Carolyn Q. Zou, Jonne Kamphorst, Niles Egan, Aaron Shaw, Benjamin Mako Hill, Carrie Cai, Meredith Ringel Morris, Percy Liang, Robb Willer, Michael S. Bernstein

**The PDF file includes:**

## 1. Constructing the Agent Bank

We created over 1,000 generative agents, each modeling a real individual in the U.S., collectively forming a representative sample of the U.S. population. To achieve this, we recruited a stratified sample of 1,052 individuals from the U.S. and conducted two hour voice-to-voice interviews using an AI interviewer (SM 2). In addition, we collected each participant's responses to a series of surveys and behavioral experiments. The interview transcripts and responses to surveys formed a knowledge base about participants that we use to condition agent behaviors (SM 3). We then combine survey responses and experiments to assess the fidelity of the resulting agents. This section details the participant data collection procedure, including participant recruitment and flow, demographic distributions, and informed consent.

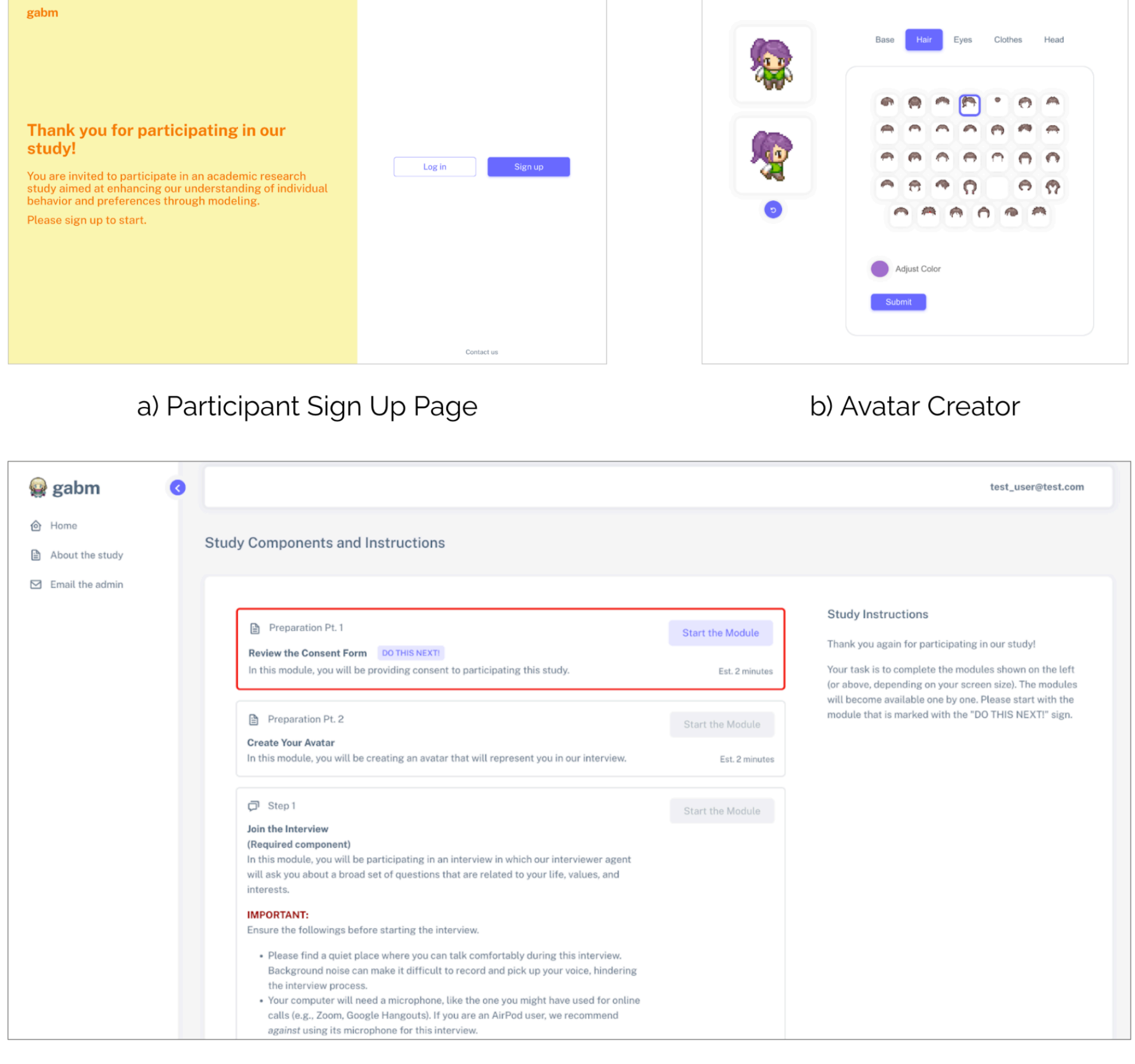

a) Participant Sign Up Page

b) Avatar Creator

c) Main Interface Displaying the Study Components

**Figure 1**. The study platform and interface. Once recruited, our participants are routed to our custom-built platform. The interface includes several components: a) Participant sign-up page: Participants sign up with an ID and password of their choice. b) Avatar creator: Participants consent and create a 2-D sprite avatar to represent them in the study platform. c) Main interface displaying the study components: The modules include: 1) study consent, 2) avatar creation, 3) interview, 4) surveys and experiments, 5) self-consistency retake of the surveys and experiments. The modules only become available in order; the button to start a module becomes clickable once the participants have completed all previous modules. The self-consistency survey and experiment module only becomes available two weeks after the participants have completed the previous modules.

Data Collection Procedure

Data collection took place on our custom-built platform, where participants signed up with an ID and a password of their choosing (see Figure 1 for visual records of the study platform). This process was conducted in two phases. In the first phase, participants were informed about the study's goals, scholarly benefits, and potential risks, and provided informed consent. Once our participants provided consent to the study, they created a custom, 2-D sprite avatar using our avatar creator to visually represent themselves on the study platform. They then completed a two-hour interview with our AI interviewer, followed by a series of surveys and experiments. The surveys and experiments were administered in the following order: the General Social Survey (GSS; *27*), the 44-item Big Five Inventory (BFI-44; *28*), five behavioral economic games *(29-32)*, and five behavioral experiments *(33-38)*. Within the economic games and the replication studies, the order of the subcomponent studies was randomized for counterbalancing purposes, and similarly for the BFI-44, the order of questions was randomized. The GSS adhered to the sequence recommended by its documentation *(64)*. Details of all surveys and experiments appear in SM 4.

For the second phase, participants joined a follow-up study two weeks after their first phase participation. In this phase, participants completed the same surveys and experiments as in the first phase, except for the interview. This was done to account for any inconsistency in responses to surveys and experiments, allowing us to measure the internal consistency of participants' responses over a two-week period.

Recruitment and Demographics

We aimed to enroll a total of 1,000 participants who would complete all study components, including the second phase participation scheduled two weeks after the initial involvement. The sample size of 1,000 was determined to ensure that we could replicate the five behavioral experiments in our study with appropriate statistical power. Anticipating an attrition rate of approximately 20% for the re-taking session, we recruited 1,300 participants for the first phase of the study. The participation rate for the self-consistency  phase was higher than expected; to maintain the representativeness of our sample, we retained 1,052 participants in the final pool.

Participants were recruited through Bovitz, a study recruitment firm *(51)*. Our stratification strategy aimed to recruit a nationally representative sample based on age, race, gender, region of residence, educational attainment, and political ideology. The only inclusion criteria were that all participants must be at least 18 years old and currently living in the U.S. Participants were paid $60 for agreeing to participate in the first phase, which included the interview and the initial round of surveys and experiments. They were paid an additional $30 for joining the second phase of the study. Additionally, for both phases, participants were eligible for a bonus payment ($0 to $10) depending on their choices in the economic games. The mean age of our participants was 47.55 (std = 15.93), with maximum age being 84 and minimum 18. With respect to gender, 593 participants identified as female, and 459 as male. With respect to education, 283 participants held a bachelor's degree, 151 had a higher degree, 185 had an associate's degree, and the rest had a high school diploma or some high school-level education. With respect to ethnicity, 833 of our participants identified as white/Caucasian, 154 as black/African American, 53 as Asian, and 95 as other. Note that for ethnicity, the participants could choose more than one option. The full demographic breakdown is described in Table 1.

Participant Consent

The data we collect in this study, particularly the qualitative interview data, is difficult to anonymize and poses a risk due to the potentially sensitive nature of the interview content. Therefore, in addition to following best practices and employing precautionary strategies for providing the agent bank access to the scientific community, we placed significant emphasis on the consent procedure. We worked with our Institutional Review Board (IRB) for over six months to ensure that participants maintain autonomy and provide informed consent. Participants were made aware that, despite efforts to de-identify their data by programmatically replacing all occurrences of their names with pseudonyms, there is still a possibility that the information they provide—such as demographic details, personal history, and political views—may be inadvertently shared as researchers use the Agent Bank. They were informed that their data would be used to develop AI models simulating human behaviors, which "aim to simulate how [they] might behave in specific situations or respond to certain survey questions," and that these agents and data might become available to other researchers strictly for academic purposes.

Additionally, participants were informed that they have the right to withdraw their consent at any time, even after completing their participation. Requests for data removal will be honored for the first 25 years following the completion of the study to the best of our ability. Participants will also be kept informed of significant changes in the model's capabilities that may affect their privacy, with the assurance that privacy risks will be reassessed as necessary. Despite our best efforts, participants are aware of the inherent risks involved in the collection of personal information, acknowledging that "achieving complete anonymity remains challenging." They are also aware that "the models we construct may become increasingly powerful over time," potentially inferring more information than currently feasible.

## 2. Creating the AI Interviewer Agent

To ensure high quality and consistency of the interview data, we developed an AI interviewer agent to conduct semi-structured interviews with study participants. Conducting large-scale interviews using human interviewers poses significant challenges, including threats to data quality, consistency, and scalability *(65)*. By employing an AI interviewer-agent built with a variant of the generative agent architecture *(17)*, we aimed to ensure uniformity in the style and quality of interviewer interactions across all participants. Additionally, this approach allowed us to scale up our data collection to over 1,000 participants.

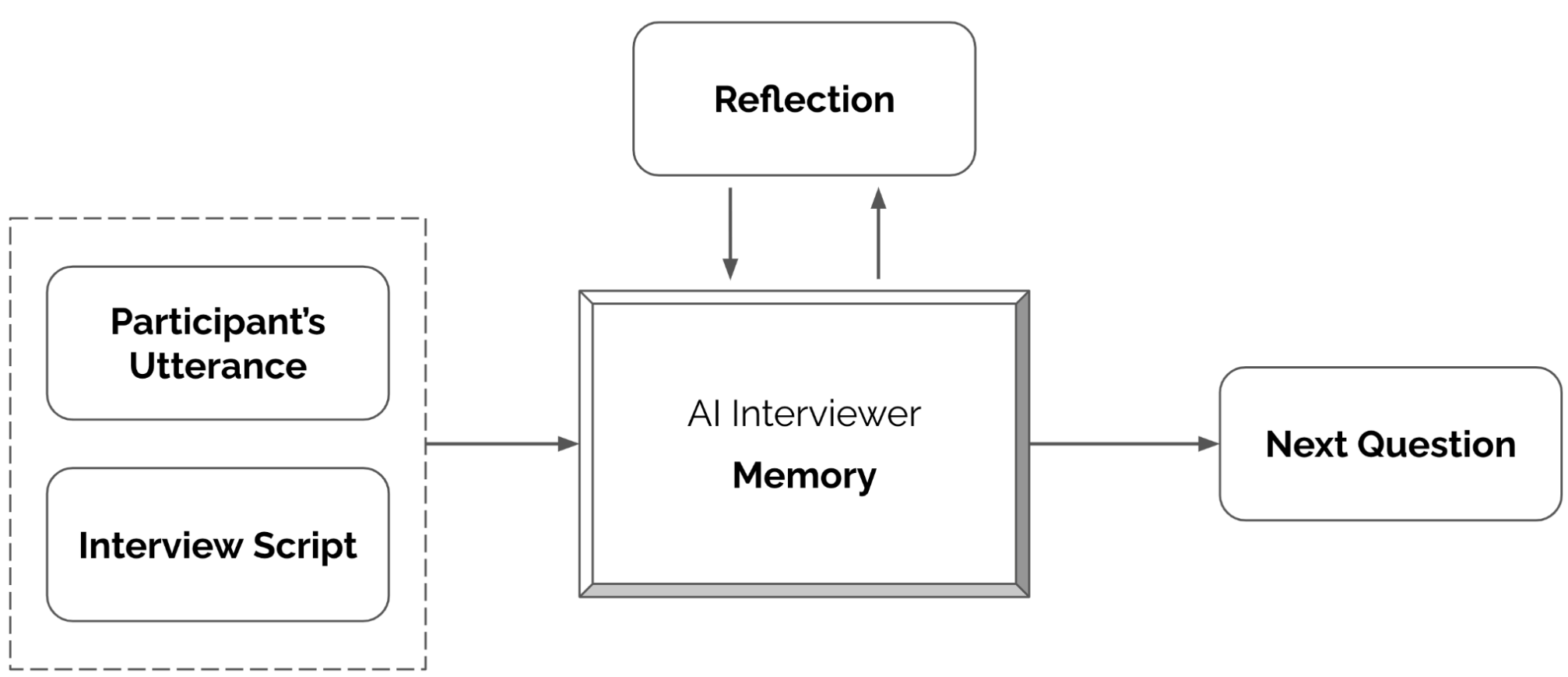

**Figure 2.** The architecture of the interviewer agent. It takes as input the participants' utterances and the interview script, generating the next action in the form of follow-up questions or deciding to move on to the next question module using a language model. A reflection module helps the architecture succinctly summarize and infer insights from the ongoing interview, enabling the agent to more effectively generate follow-up questions.

AI Interviewer Agent Architecture

A trained human interviewer knows when and how to ask meaningful follow-up questions, balancing the need to adhere to a well-designed interview script while allowing for detours that help participants open up and share aspects they may have initially forgotten or not thought to share *(24, 25, 66)*. To instill this capability in an AI interviewer agent, we needed to design an interviewer architecture that affords the researchers control over the overarching content and structure of the interview while allowing the interviewer agent a degree of freedom to explore follow-up questions that are hard-coded in the interview script. This served as our design goal for the AI interviewer agent.

Our interviewer architecture takes an interview protocol and the most recent utterances from the interviewee as inputs and outputs an action to either: 1) move on to the next question in the script, or 2) ask a follow-up question based on the conversation so far. The interview script is an ordered list of questions, with each question associated with a field indicating the amount of time to be spent on that particular question. At the start of a new question block in the interview script, the AI interviewer begins by asking the scripted question verbatim. As participants respond, the AI interviewer uses a language model to make dynamic decisions about the best next step within the time limit set for the question block. For instance, when asking a participant about their childhood, if the response includes a remark like, "I was born in New Hampshire… I really enjoyed nature there," but without specifics about what they loved about the place in their childhood, the interviewer would generate and ask a follow-up question such as, "Are there any particular trails or outdoor places you liked in New Hampshire, or had memorable experiences in as a child?" Conversely, when asking the participant to state their profession, if the participant responds, "I am a dentist," the interviewer would determine that the question was completely answered and move on to the next question.

The reasoning and generation of the follow-up questions were done by prompting a language model. However, to generate effective actions for the interviewer, the language model needed to remember and reason over the prior conversational turns to ask meaningful follow-up questions that are informed and relevant in the context of what the participants have already shared. While modern language models have become increasingly proficient at reasoning, they still struggle to consider every piece of information in the prompt if it is too long *(67)*. Thus, indiscriminately including everything from the interview up to that point risks gradually degrading the performance of the interviewer in generating effective follow-up questions or decisions to move on.

To overcome this, our interviewer architecture includes a reflection module that dynamically synthesizes the conversation so far and outputs a summary note describing inferences the interviewer can make about the participants. For instance, for the participant mentioned earlier, it would generate reflections such as:

```
{
  "place of birth": "New Hampshire"
  "outdoorsy vs. indoorsy": "outdoorsy with potentially a lot of
  time spent outdoors"
}
```

Then, when prompting the language model to generate the interviewer's actions, instead of including the full interview transcript, we included the much more concise but descriptive reflection notes the interviewer had accumulated up to that point and the most recent 5,000 characters from the interview transcript (Figure 2).

Interview Script

With the design of the interview script fed to our interviewer agent, we aimed to satisfy two goals. The first goal, shared with qualitative research, is that a well-designed script with questions that inspire meaningful answers is crucial for the study's objective of creating generative agents that encapsulate a nuanced portrait of the individuals we are modeling. The second goal is more unique to our study: we wanted an interview script that was designed independently of our evaluation metrics, by researchers outside our team. This approach ensures that we do not unfairly tailor the content of the interview script to favor or align with predicting participants' responses to the specific surveys and experiments included in our study.

To conduct the interviews, we employed a slightly abbreviated version of the interview script developed and used by the American Voices Project *(26)*, which we include in Table 7. The American Voices Project initiative involves recruiting a representative sample of the U.S. population for in-depth, approximately three-hour interviews. During these interviews, participants are questioned about their life experiences, including their life stories and perspectives on various social, political, and value-related topics. For instance, the interview script starts with an open-ended and broad question such as, "To start, I would like to begin with a big question: tell me the story of your life. Start from the beginning—from your childhood, to education, to family and relationships, and to any major life events you may have had." It then proceeds to more topical questions, such as "How have you been thinking about race in the U.S. recently?" We selected this interview script due to its broad coverage that explores the lived experiences of the interviewees. However, the script is extensive and includes specific questions that delve into intricate details that we considered too specific for many potential use cases of our agents, such as individuals' financial spending in various categories. Therefore, to make the interview manageable in a two-hour session, we omitted parts of the script (e.g., rapid-fire questions that delve into specific details of the participants' spending habits, or COVID-era life pattern changes) during our interviews with the participants.

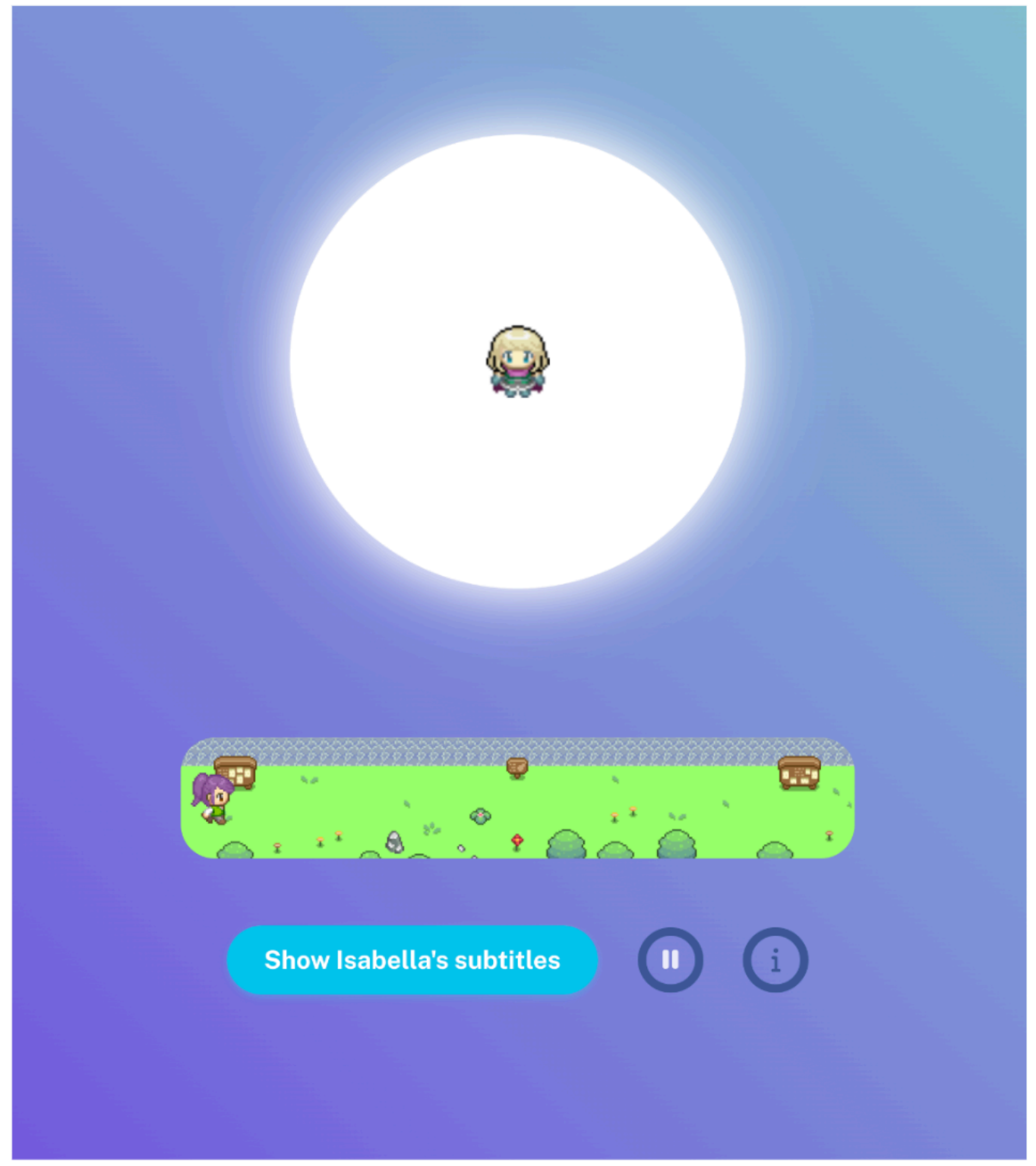

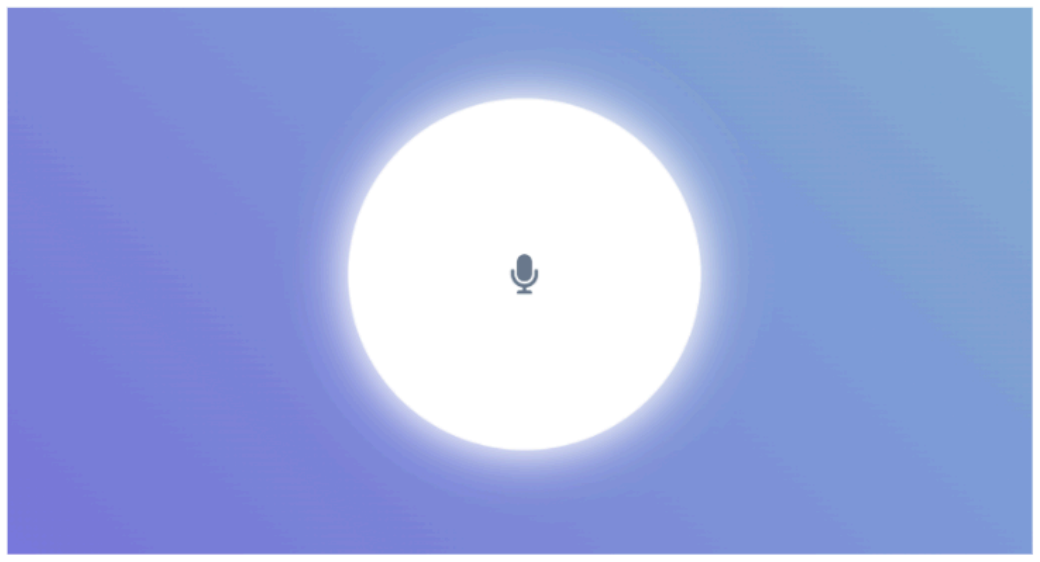

b) A microphone icon prompts the participants to speak

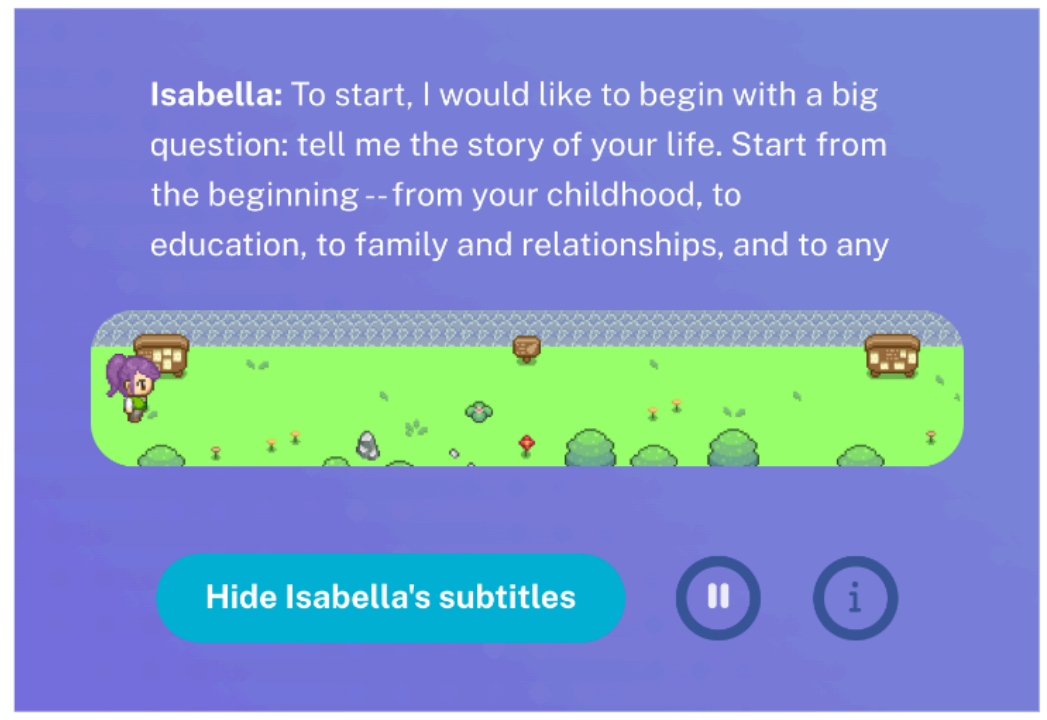

a) The main interview interface

c) A progress bar and optional subtitles

**Figure 3**. The interview interface. a) The main interview interface: A 2-D sprite representing the AI interviewer agent is displayed in a white circle that pulsates to match the level of the audio, visualizing the interviewer agent's speech during the AI interviewer's turn. b) Participant's response: The 2-D sprite of the AI interviewer agent changes into a microphone emoji when it is the participant's turn to respond, with the white circle pulsating to match the level of the participant's audio being captured. c) Progress bar and subtitles: A 2-D sprite map shows the participant's visual avatar traveling from one end point to the other in a straight line, indicating progress. The interface also features options to display subtitles or pause the interview.

<u>Implementation</u>

We implemented the interviewer agent as a web application in our study platform, providing voice-to-voice interactions with audio and microphone capabilities through an audio Zoom-like interface. Low-latency voice-to-voice interviews were crucial for giving participants the feeling of actually talking to an interviewer and helping the AI interviewer agent form rapport with the interviewee *(68)*. Before the interview, our platform disclosed that our interviewer is an AI, and conducted an audio calibration by asking participants to read aloud the first two lines of *The Great Gatsby* by F. Scott Fitzgerald.

The interview interface displayed the 2-D sprite avatar representing the interviewer agent at the center, with the participant's avatar shown at the bottom, walking towards a goal post to indicate progress (see Figure 3). When the AI interviewer agent was speaking, it was signaled by a pulsing animation of the center circle with the interviewer avatar. When it was the participant's turn to speak, the interviewer avatar changed to a microphone emoji and pulsed to match the audio being recorded, indicating the sound level registered from the participant. If the participant stopped speaking and silence lasted for longer than 4 seconds, the circle gradually faded, and the audio recording for the participant's utterance stopped. At this point, a loading animation appeared while generating the AI interviewer agent's next utterance. Our interviewer agent generally responded within 4 seconds—reasoning, generating, and returning its generated voice responses within this time frame—to maintain a smooth interview flow. The AI interviewer agent automatically started speaking when it was ready, with the interviewer avatar being displayed in the circle again.

The interview script is communicated to the AI interviewer agent as a JSON file containing an ordered list of questions. Each question is paired with a metadata field indicating a manually set time limit, suggesting the amount of time to be spent on each question so that the interview can conclude within 2 hours. Every question in the script, along with the follow-up questions, is read aloud using OpenAI's Audio model, a text-to-speech model that generates voice audio from textual input *(69)*. The participants' voice responses were transcribed using OpenAI's Whisper model, a speech-to-text model that converts voice audio into text *(70)*. This transcription allows us to use the interview transcript to prompt the language models to determine the next conversational move. Then, to dynamically generate reflections for the participants' responses to the current question, we prompted OpenAI's GPT-4o language model *(71)* with the following prompt (with input fields dynamically filled in):

```
Here is a conversation between an interviewer and an interviewee.
<INPUT: The transcript of the most recent part of the
conversation>

Task: Succinctly summarize the facts about the interviewee based
on the conversation above in a few bullet points -- again, think
short, concise bullet points.
```

And to dynamically generate new questions, we prompted GPT-4o with a prompt that looks as follows:

```
Meta info:
Language: English
Description of the interviewer (Isabella): friendly and curious
Notes on the interviewee: <INPUT: Reflection notes about the
participant>

Context:
This is a hypothetical interview between the interviewer and an
interviewee. In this conversation, the interviewer is trying to
```

```
ask the following question: "<INPUT: The question in the
interview script>"

Current conversation:
<INPUT: The transcript of the most recent part of the
conversation>
=*=*=
Task Description:
Interview objective: By the end of this conversation, the
interviewer has to learn the following: <INPUT: Repeat of the
question in the interview script, paraphrased as a learning
objective>
Safety note: In an extreme case where the interviewee
*explicitly* refuses to answer the question for privacy reasons,
do not force the interviewee to answer by pivoting to other
relevant topics.

Output the following:
1) Assess the interview progress by reasoning step by step --
what did the interviewee say so far, and in your view, what would
count as the interview objective being achieved? Write a short
(3~4 sentences) assessment on whether the interview objective is
being achieved. While staying on the current topic, what kind of
follow-up questions should the interviewer further ask the
interviewee to better achieve your interview objective?
2) Author the interviewer's next utterance. To not go too far
astray from the interview objective, author a follow-up question
that would better achieve the interview objective.
```

On average, with this implementation, our AI interviewer agent spoke 5372.59 (std=2406.12) words during the interview, asking on average 81.71 (std=54.39) follow-up questions from 99 scripted questions, to which our participants responded with on average 6491.19 words (std=2540.56).

### Development Process and Evaluation

We iteratively refined and assessed the performance of the AI interviewer agent using a sample of 3 waves of 3 pilot interviews conducted between August 2023 to November 2023. Of the pilot interviews, 10 were conducted by human interviewers and 25 by the AI interviewer agent. The resulting interview transcripts were evaluated by members of our research team who were trained in the social sciences and assessed for their performance as training data for generative agents on the same set of attitudinal and behavioral tasks presented in the main results of this article. We compared the quality of the AI Interviewer interview transcripts to transcripts of interviews performed by expert human interviewers as part of the American Voices Project. By the end of this pilot stage, our team concluded that the quality of the transcripts produced by the AI interviewer agent compared well with those produced by human interviewers.

Additionally, in our pilot studies the performance of the generative agents of real people created using transcripts collected by the AI interviewer agent matched those created using transcripts collected by human interviewers when predicting GSS responses. For instance, our deployed AI interviewer agent could maintain a smooth conversation, ask relevant follow-up questions when participants' responses were terse, and navigate sympathetically after hearing about a difficult experience. This can be seen from an excerpt of an interview for the first question of the interview—only the first interviewer question is scripted, while the subsequent utterances from the interviewer are generated:

> **[Interviewee]:** I was raised in a rural town in [Redacted] where there are actually more cows than people. I'm [Redacted] years old, so I am a child of the [Redacted]. I wound up being a latchkey kid and was largely raised either by myself or by daycare and babysitters. Childhood was not really great and I was really glad to move on to high school.
>
> **[Interviewer]:** I'm sorry to hear that childhood wasn't great for you. Could you tell me more about your experiences in high school? What was it like for you there, and were there any particular events or people that had a significant impact on your life during that time?
>
> **[Interviewee]:** So I had a head injury while I was in elementary school and the full weight of that untreated injury came to bear when I was in about 8th grade and I started having these really bad mood swings. So I spent most of my high school years in deep depression and no one got me any help for it. It was pretty awful. I excelled in English, in theater, did pretty well in science, won a lot of awards at the end of the school time when it was time to graduate. Didn't have the greatest GPA because I had a habit of skipping classes and not doing any homework but acing finals. Particular events or people that had a significant impact, I would say that the album Pink Floyd The Wall pretty much helped me with my really terrible mood. I felt understood by that music.
>
> **[Interviewer]:** Thank you for sharing that with me. It sounds like high school was a particularly challenging yet formative time for you. After high school, what path did you take with your education? Did you go to college, or did you jump straight into a career?
>
> **[Interviewee]:** After high school I went off to college originally to study criminal justice, but I changed over to English literature with a concentration in behavioral studies. After college, well, during college I had a bunch of different jobs. I worked with radioactive waste disposal, I worked at a woman's bookstore, I worked at a nightclub. I went off to law school and was there until I had to take a leave of absence. Years later, like 20 years later, I wound up going back to school again and got a business degree, mainly because I thought it would be fun to do. In the meantime, between law school and going back to business school (...). Then I took over my parents' business as a manager. After I worked as a manager there and I moved out of state in order to get married, I wound up buying into that business, so now I'm a part owner of the business where I'd started working when I was 15, so it's kind of gone full circle.

## 3. Generative Agent Architecture

Generative agents are software systems that simulate human behavior, powered by a language model augmented with a set of memories to define their behaviors *(39, 17)*. These memories, stored in a database (or "memory stream") in text form, are retrieved as needed to generate the agent's behaviors using a language model. This is paired with a reflection module that synthesizes these memories into reflections, selecting portions or all of the text in the agents' memories to prompt a language model to infer useful insights, thereby enhancing the believability of the agents' behaviors. While traditional agents in agent-based models rely on manually articulated behavior in specific scenarios, generative agents leverage language models to produce human-like responses that reflect the personas described in their memories across a wide range of circumstances. In this work, we aimed to build generative agents that accurately predict individuals' attitudes and behaviors by using detailed information from participants' interviews and/or survey responses to seed the agents' memories, effectively tasking generative agents to role-play as the individuals that they represent.

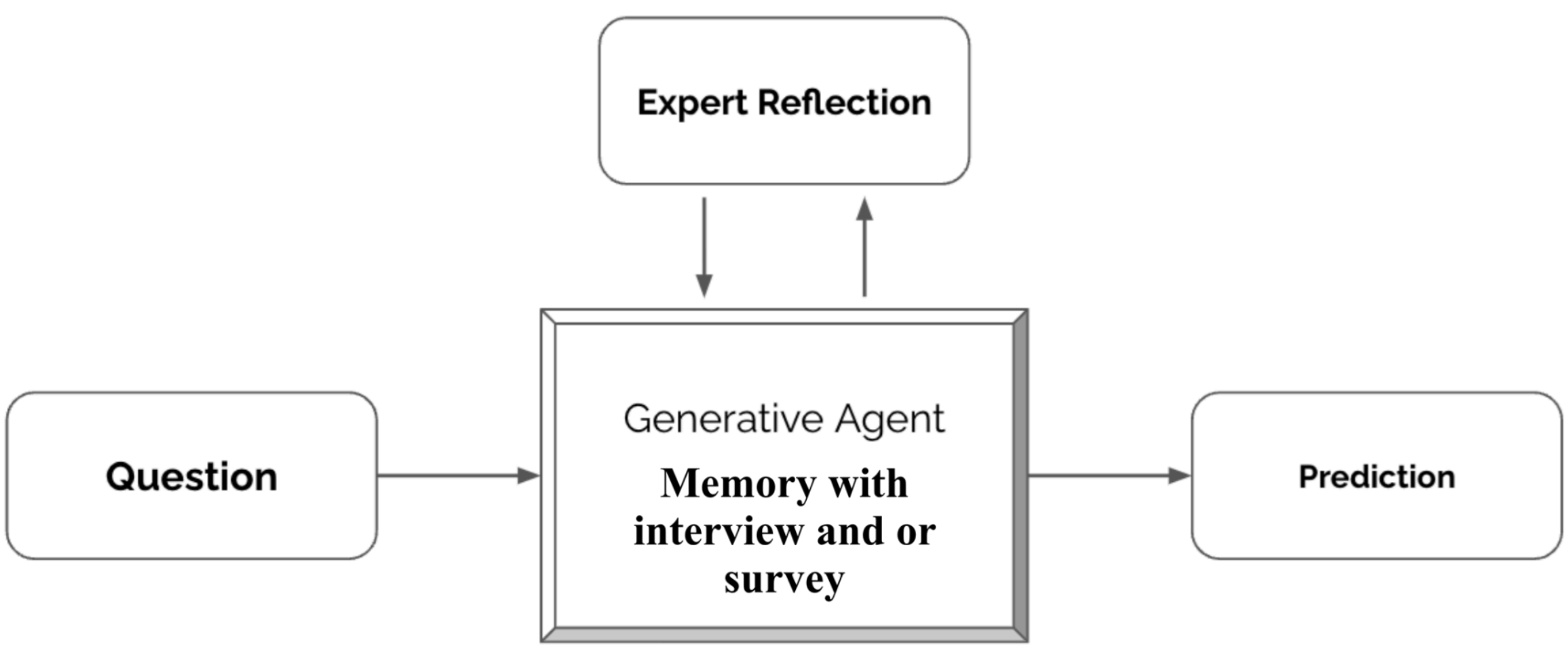

**Figure 4**. The architecture of our generative agents involves taking a question as input and outputting a prediction of how the source participant might respond, using a language model. Each agent's memory comprises the self-report data and the outputs of expert reflections over that data. These reflections are short syntheses generated using a language model, designed to infer insights about the participants that might not be explicitly stated. The personas of expert social scientists (e.g., psychologist, behavioral economist) guide these reflections.

Expert Reflection

Prompting the language model with participants' self-report data to predict their responses in a single chain of thought may cause the model to overlook latent information not explicitly stated by the participant. To explicitly infer high-level, more abstract insights about the participants embedded in the self-report data, we introduced a variant of generative agents' reflection module called "*expert reflection*." In this module, we prompt the model to generate reflections on a participant's data, but instead of simply asking the model to infer insights from the data, we ask it to adopt the persona of a domain expert. Specifically, we ask the model to generate four sets of reflections, each time taking on the persona of a different domain expert from four branches of social sciences: psychologist, behavioral economist, political scientist, and demographer. These sets of reflections synthesize insights relevant to the domain represented by each expert. For instance, for one interview transcript, the expert personas generated different insights:

> Psychologist: “[Redacted] values his independence and expresses a clear preference for autonomy, particularly highlighted by his enjoyment of traveling for his job and his frustration with his mother's overprotectiveness. This suggests a strong desire for personal freedom and self-determination.”
>
> Behavioral Economist: “[Redacted]’s aspiration to save for a relaxing vacation and possibly advance to a managerial position indicates a blending of practical financial goals with personal leisure aspirations, emphasizing balanced life satisfaction.”
>
> Political Scientist: “[Redacted] identifies as a Republican and espouses strong support for the party's views, particularly around immigration and drug policy. However, he also expresses specific support for traditionally Democratic positions on issues like abortion rights and the legalization of marijuana, suggesting a blend of ideologies.”
>
> Demographer: “[Redacted] works as an inventory specialist and earns between $3,000 to $5,000 monthly, contributing to a household income of around $7,000 per month. He works primarily at Home Depots but has a varied work schedule, indicating some job stability and flexibility.”

For every participant, we generated these four sets of reflections by prompting GPT-4o with the participants’ self-report data and asking it to generate up to 20 observations or reflections for each of the four experts. The prompt, tailored for each expert, was similar to the following (for the demographer expert):

```
Imagine you are an expert demographer (with a PhD) taking notes
while observing this interview. Write observations/reflections
about the interviewee's demographic traits and social status.
(You should make more than 5 observations and fewer than 20.
Choose the number that makes sense given the depth of the
interview content above.)
```

We generated these reflections once and saved them in the agents’ memory. Whenever we needed to predict the participants’ responses to a question, we first classified, by prompting the language model, which domain expert (demographer, psychologist, behavioral economist, or political scientist) would best answer the question. We then retrieved all reflections generated by that particular expert. Along with the self-report data, these sets of reflections informed the language model’s generation of predictions for the participants’ responses. After retrieval, we appended the reflections to the participants’ data and used this to prompt GPT-4o to generate responses.

<u>Generating a Prediction With Generative Agents</u>

Our prompting strategy leveraged the chain-of-thought prompting approach. For example, when using an interview transcript, we used the following prompt:

```
<INPUT: Participant’s interview transcript and relevant expert
reflections>

=====

Task: What you see above is an interview transcript. Based on the
interview transcript, I want you to predict the participant's
survey responses. All questions are multiple choice, and you must
guess from one of the options presented.

As you answer, I want you to take the following steps:
Step 1) Describe in a few sentences the kind of person that would
choose each of the response options. ("Option Interpretation")
Step 2) For each response option, reason about why the
Participant might answer with that particular option. ("Option
Choice")
Step 3) Write a few sentences reasoning on which of the options
best predicts the participant's response. ("Reasoning")
Step 4) Predict how the participant will actually respond in the
survey. Predict based on the interview and your thoughts.
("Response")

Here are the questions:

<INPUT: Question we are trying to respond to>
```

To predict numerical responses, we modified the ending and prompted:

```
[... Same as the categorical response prompt]

As you answer, I want you to take the following steps:
Step 1) Describe in a few sentences the kind of person that would
choose each end of the range. ("Range Interpretation")
Step 2) Write a few sentences reasoning on which option best
predicts the participant's response. ("Reasoning")
Step 3) Predict how the participant will actually respond.
Predict based on the interview and your thoughts. ("Response")

Here are the questions:

<INPUT: Question we are trying to respond to>
```

Finally, if the agents needed to maintain context from the experimental stimuli for the behavioral experiments, we appended the agents' received stimuli and prior actions in the experiments at the end of the self-report data and reflections in natural language form.

## 4. Surveys and Experimental Constructs

To evaluate the fidelity of our generative agents, we aimed to assess their predictive accuracy regarding the attitudes and behaviors of the underlying sample across surveys and experimental constructs from a broad array of social scientific disciplines and methods. To operationalize this, we identified four existing constructs commonly deployed in the social sciences. In this section, we describe them. As discussed in the manuscript, when we evaluate on constructs that are used in the input for the agents, we ensure to remove the question we predict on from the input (for the GSS), and all questions in the same question block (for the Big5).

### The General Social Survey

The General Social Survey (GSS) is a long-running, widely used sociological survey administered biannually to representative cross sections of U.S. adults to collect information encompassing demographic details and respondents' viewpoints on issues such as government spending, race relations, and beliefs concerning the existence and nature of God *(27, 72)*. The survey consists of a repeated "core" module, run every cycle, that covers the more timeless elements of the survey, along with additional modules that are swapped in and out to meet the needs of the year when the survey is administered. Traditionally, the survey was conducted via voice, with the surveyor asking the questions in person or on the phone and later coding the participants' responses into discrete survey question and answer pairs. In recent years (since the COVID-19 pandemic, when in-person contact was challenging), the GSS has been implemented and administered online *(73).* The ability to predict GSS responses signifies a comprehensive understanding of individuals, particularly in areas of interest to social scientists *(72)*. It also signifies the ability of our agents to predict the participants' responses to survey constructs on topics related to societal issues and personal views.

In our study, we focused on the GSS Core as it represents the most enduring and important set of survey questions within the GSS. While some questions within the GSS allow for qualitative responses or freeform input, our study specifically focused on questions that ask for structured responses in the form of categorical or numerical answers. These questions can be quantitatively assessed, making them the primary focus of our evaluation efforts. Consequently, we excluded: 1) conditional questions that depend on answers to other questions, 2) questions with more than 25 option categories, and (3) questions requiring free-form responses. This refinement process resulted in a final analytic sample of 177 categorical questions and 6 numerical questions. Following recent best practice *(73)*, we administered these questions online through a custom-built Qualtrics survey linked from our study platform.

### Big Five Personality Traits

The Big Five personality traits are a widely recognized framework in psychology for understanding human personality *(28)*. The construct encompasses five broad dimensions that capture substantial variability in individuals' personalities: openness to experience, conscientiousness, extraversion, agreeableness, and neuroticism. Each trait represents a range, with individuals receiving a score for each trait. The Big Five traits have been validated across diverse cultures and are used to predict a wide range of behaviors and life outcomes, from academic and job performance to social relationships and mental health *(28)*. Traditionally, these traits are assessed using self-report questionnaires where respondents rate their agreement with each statement on a Likert scale. The ability to accurately predict an individual's Big Five personality traits is central for psychologists and researchers *(74, 76)*, as it provides insights into human behavior, informs interventions, enhances personal development, and contributes to the understanding of social dynamics.

In our study, we used the 44-item version of the questionnaire for testing the Big Five personality traits (BFI-44), developed by Oliver John and Sanjay Srivastava in 1998 *(28)*. We administered all 44 questions in the BFI-44 through a custom-built Qualtrics form linked from our study platform. Using the participants' responses to these questions, we calculated the scores following the aggregation methods suggested in the original work.

Behavioral Economic Games

Behavioral economic games are a set of experimental tools to study decision-making and social behavior under real stakes. Each game is designed to reveal aspects of human social behavior such as altruism, trust, cooperation, and competition. Participants engage in these games with real financial incentives. In our study, we offered participants a bonus payment based on their choices in the games. This incentive helps ensure that participants' choices reflect genuine preferences and strategies. These economic games are pivotal in understanding the underlying motivations and behavioral patterns in economic decision-making, providing valuable data for social scientists in multiple fields *(74)*.

In our study, we included the following five economic games, chosen for their prominence in the academic community:

- Dictator Game *(29)*: This game measures altruism by allowing one player (the dictator) to decide how to split a sum of money between themselves and another participant. The participants are initially given $5 and decide how to split the $5 between themselves and one other participant. The other participant cannot affect the outcome chosen.
- The Trust Game (First Mover) *(30)*: This game assesses trust, with the First Mover deciding how much of an initial endowment to send to the Second Mover. Player 1 is initially given $3 and then chooses how many dollars, if any, to send to Player 2, another participant in the study. Unlike the Dictator Game, the sum sent is tripled, so for every $1 sent, Player 2 receives $3.
- The Trust Game (Second Mover) *(30)*: This game assesses reciprocity, with the Second Mover deciding how much of the tripled amount received from the First Mover to return. This game continues from the actions of the First Mover in the Trust Game, as described above.
- The Public Goods Game *(31)*: This game explores cooperative behavior by having participants decide how much of their private endowment to contribute to a common pool. Participants are randomly assigned to interact with three other participants, with everyone receiving the same instructions. Each person in the group is given $4 for this interaction. They each decide how much of the $4 to keep for themselves and how much (if any) to contribute to the group's common task. All money contributed to the common task is doubled and then split among the four group members. The money is equally distributed among all players, regardless of individual contributions.

- The Prisoner's Dilemma *(32)*: This game examines the tension between cooperation and self-interest. Each of two participants chooses whether to cooperate or defect. If both cooperate, they receive a moderate payoff of $6. If one defects while the other cooperates, the defector receives a high payoff of $8, and the cooperator receives a low payoff of $2. If both defect, they receive a low payoff of $4.

We administered all five games through a custom-built Qualtrics form linked from our study platform and calculated the bonus amount afterward. At the start of the five games, we informed participants that we would randomly select one of the five studies to calculate their bonuses, but we did not tell them which game would be selected. Each game offered a maximum bonus of $8 to $10. After the study, we randomly paired participants and used the Dictator Game to calculate the bonus amount—the selection of the Dictator Game was randomly determined by us prior to the study. The bonus amount was based on participants' actual earnings in that game.

Replication Studies of Experimental Treatment Effects

Experiments involving randomized controlled trials (RCTs) of interventions ("treatments") with human subjects are standard across the social sciences. In a typical case, study participants are randomly assigned to either a treatment group or a control group, ensuring that outcome differences can be attributed to the intervention and minimizing confounding variables. Somewhat more formally, in this basic example, random assignment ensures that treatment assignment ($Z$) is independent of the observed outcomes ($Y(1)$ for treatment or $Y(0)$ for control) conditional on any set of observed or unobserved covariates ($X$) (i.e., $Z \perp \{Y(0), Y(1)\} | X$). Researchers often estimate treatment effects by calculating the difference between the average (mean) outcome in the treatment group and control group. Estimates of treatment effects from rigorous, well-replicated studies support evidence-based practice in social sciences like psychology and guide policy decisions and clinical recommendations *(77, 78)*. By predicting such effects with generative agents, we evaluate whether individual agent behaviors also exhibit (aggregate) responses to interventions in ways that accurately simulate human samples.

*Our analytic sample of the replication studies.* In our study, we selected a sample of human behavioral experiments from a recent large-scale replication effort of experimental studies that were published in the *Proceedings of the National Academy of Sciences* as curated by Camerer et al. *(38)*. Each study had at least one clear hypothesis and a significant reported effect. Sampling from the studies replicated by Camerer et al. ensured that 1) we did not subconsciously choose studies more favorable to the generative agents; 2) all estimates of treatment effects among human study participants had already been replicated in pre-registered studies conducted by independent research teams and subjected to peer review; and 3) the interventions we tested were drawn from multiple social scientific disciplines.

The Camerer et al. project replicated 41 studies in total. Among these, we selected studies based on two criteria: first, the study had to be describable in natural language (optionally with images) for processing by a language model; second, the power analysis from the replication effort suggested that the effects would be observable with 1,000 or fewer participants. The criteria ensured that our sample of 1,000 human participants and the corresponding 1,000 generative agents could replicate the effect if present. These filters resulted in the following five studies in our sample of experiments:

- Ames & Fiske, 2015 *(33)*: This study examines how perceived intent affects the evaluation of harm. Participants read a vignette about a nursing home employee who switched patients' medications. One group was told the switch was intentional, while the other was told it was unintentional. After reading the vignette, participants were asked to complete their choice of five tasks, such as providing opinions about how the nurse should be blamed and punished or taking a short quiz about the cost of healthcare in the U.S. The study found that those who read the intentional scenario were more likely to choose tasks related to assigning blame and punishment compared to those who read the unintentional scenario.
- Cooney et al., 2016 *(34)*: This study explores how perceived fairness affects emotional responses. In a modified dictator game, participants believed they were receivers and predicted whether they would feel less upset about not receiving a bonus if the decision was made fairly (by a coin flip) rather than unfairly (by personal choice). The study found that participants expected fairness to influence their feelings, anticipating less upset when the decision was perceived as fair.
- Halevy & Halali, 2015 *(35)*: This study examines the perceived benefits and costs of intervening in conflicts. Participants recalled their personal experience of either intervening or not intervening in a conflict between friends and were asked to assess how beneficial it was to intervene in the conflict. The results showed that those who recalled intervening perceived the intervention as more beneficial and less costly than those who did not intervene.
- Rai et al., 2017 *(36)*: This study explores how dehumanization affects participants' willingness to harm others. In a vignette-based experiment, participants were given a description of a person in a dehumanized manner, simply as a "man," or a description of a person in a humanized manner with details about the person such as "John is a 29-year-old man with brown hair and brown eyes. People who know him would describe him as ambitious and imaginative..." Participants were then asked whether they would be willing to harm the person for monetary compensation. The study found that participants were more willing to harm a stranger described in dehumanized terms compared to one described in humanized terms when motivated by financial gain.
- Schilke et al., 2015 *(37)*: This study investigates how power influences trust in social exchanges. Participants, imagining themselves as typists, were divided into high-power and low-power groups based on financial need and job availability. In the high-power condition, participants' service was essential for their clients, and they were offering the service to make extra spending money. In the low-power condition, participants' service was non-essential for their clients, and they were offering the service to make ends meet. Trust was measured by the willingness to provide a free sample of their service to a potential client. The results showed that participants in the high-power condition were less willing to offer a free sample.

Our platform randomly assigned the participants to a condition for each of the five studies and we administered all five experiments through a custom-built Qualtrics form linked from our study platform.

## 5. Evaluation Methods

Given our survey and experimental constructs, we set out to evaluate the predictive power of the generative agents. In this section, we describe the metrics and evaluation methods used for this purpose. The individual subsection headers in this section are organized to match the presentation in the main document.

### Predicting Individuals' Attitudes and Behaviors

To determine whether generative agents of the 1,000 human participants accurately predict their respective individuals' behaviors and attitudes, we utilized the GSS, BFI-44, and five economic games. We deployed our generative agents to predict their respective individuals' responses to questions in these surveys and behavioral constructs. However, this measurement poses challenges due to variability in human participants' responses (*40, 11*). To address this, our evaluation employs the following strategy:

1) We use participants' responses from the first phase of participation to assess the accuracy rate of our agents' predictions—the number of answers predicted correctly over the total number of questions.
2) We use the second phase of participation to assess individuals' rate of internal consistency—the participants' rate of prediction on the battery of surveys and experiments used in this study.
3) We then calculate the *normalized accuracy* as follows:

$$\text{normalized accuracy} = \frac{\text{agent's prediction accuracy}}{\text{internal consistency}}$$

Conceptually, a normalized accuracy of 1.0 means that the generative agent predicts the individual's responses as accurately as the person replicates their own responses two weeks later.

The diverse response types in our surveys and experimental constructs present a challenge in determining a single metric for assessing our agents' predictive accuracy. For instance, while the categorical-ordinal responses in the GSS are well-suited to accuracy rates, numerical responses in other constructs are better evaluated using Mean Absolute Error (MAE) or correlation coefficients. To address this, we developed a reporting approach that satisfies the following criteria:

1) Report metrics appropriate for each response type (e.g., accuracy rate for categorical, MAE for numerical).
2) Ensure metric interpretability across different community norms (e.g., accuracy/MAE in machine learning literature, correlation in social science literature).
3) Provide a metric allowing comparison across different constructs.

To meet these criteria, we report accuracy rates for evaluation constructs with categorical-ordinal response types, MAE for numerical response types, and Pearson correlation coefficient as a metric comparable across constructs. Additionally, we present these metrics alongside the normalized accuracy to provide a comprehensive evaluation of our agents' performance.

Our analysis primarily focuses on *individual-level analyses*, conducted for each participant and then aggregated across the population. We calculate accuracy measurements for each individual and average these values across all participants. Additionally, we report and discuss *construct-level analyses*, which involve calculating accuracy measurements for the population and measuring the average predictive accuracy for individual questions, dimensions, or games in the constructs, averaged across all agents in the agent bank. In essence, the individual-level analysis answers "For the average person, how accurate are the generative agents?", whereas the construct-level analysis answers "For a specific item in one construct, how accurate are the generative agents?" In the main body of the article, we report the individual-level analysis, as our goal is to generalize over the individuals in our population rather than over items.

Averaging correlation coefficients presents a challenge due to their non-linear nature. To address this, we employ Fisher's z-transformation, which converts correlation coefficients to a scale where they can be averaged linearly (*79*). The process involves:

1) Applying Fisher's z-transformation to each correlation coefficient: $z = \frac{1}{2} ln(\frac{1+r}{1-r})$
2) Calculating the average of the z-values.
3) Applying the inverse Fisher's z-transformation to the average z-value: $r = tanh(z)$

Below, we describe in more detail the evaluation methods and our reporting strategies for the individual constructs.

*The General Social Survey.* The subset of the core module of the GSS that fits our inclusion criteria, as described in the prior section, includes 183 questions, most of which—177—are categorical or ordinal (*categorical-ordinal)* response types, with the remaining 6 being numerical. As such, we report the accuracy rate and correlation coefficient for the categorical response type variables in the GSS, and separately report the MAE and correlation coefficient for the 6 numerical response type variables.

To calculate the accuracy of the categorical-ordinal GSS questions, we consider our prediction accurate and the participant consistent if the responses match exactly. The accuracy rate is the ratio of correctly predicted items divided by the total number of items. For correlations, we translated categorical and ordinal questions into numerical forms. For categorical variables, each response option is first transformed into a separate binary variable. For example, if a categorical variable has five response options, we create five binary variables, each indicating whether the participant selected that option. These binary variables are coded as 0 or 1. However, to prevent these binary variables from disproportionately influencing the overall correlation due to their higher count, we adjust their weights. Each binary variable is assigned a weight such that the total weight for all variables combined equals that of the original categorical variable, ensuring fair representation. In this example, each binary variable would receive a weight of 0.25. For ordinal questions, we normalize the responses to a range between 0 and 1 by evenly spacing the options. This ensures that the ordinal responses maintain their inherent order and distance between options while fitting into a numerical scale. We then apply the same weighting principles when calculating the correlation between the participants' original responses and their later self-consistency responses to ensure conformity across variable types.

To calculate the MAE for the numerical GSS questions, we first normalize the participants' responses to a 0 to 1 scale relative to the range of historical responses to the respective question as indicated on the official NORC site for the GSS. For instance, for the question "Your age," if the historical minimum value was 18 and the maximum was 89, an age of 30 would be normalized to 0.17 accordingly. We use these normalized values to calculate the MAE across the questions and responses. Similarly, we calculate the MAE between the participants' responses and predictions, and between the participants' responses and their self-consistency responses.

Using these measurements for our predictions and the participants' internal consistency, we calculate the normalized accuracy. However, we note that normalized accuracy cannot be computed for MAE, as some rows contain internal consistency values of 0 (when there is no variation, meaning the participant gave the same response in both the test and retest), making the denominator 0. Therefore, we report the normalized accuracy only for accuracy and the correlation coefficient.

When we use as input any GSS questions from the first survey wave while predicting on the GSS in the second wave, we ensure that the question we are predicting on is not in the input for the agents.

*BFI-44.* Each dimension in the five-dimension measurement of the BFI-44 is provided as a scale ranging from 1 to 5. These scales are calculated based on the participants' responses to the 44 questions in BFI-44. To calculate these scales, we first reverse-code certain items as specified in the original work. After reverse coding, we aggregate the responses by taking the average for each dimension. Then, we use these averaged scales to calculate and report the MAE and correlation coefficient of the five scales. As with the GSS, we only report the normalized accuracy for the correlation coefficient. When we use as input any Big5 questions from the first survey wave while predicting on the Big5 in the second wave, we ensure that the question we are predicting or any other questions in the same personality grouping on are not in the input for the agents.

*Economic games.* The five economic games include numerical (dictator game, trust games, public goods) and dichotomous (prisoner's dilemma) response types. To normalize these into one construct, we first normalize the responses for each game to a 0 to 1 scale using the minimum and maximum values offered to the participants as min and max. Then, we use the normalized values to calculate the MAE and correlation coefficient. Again, as with the GSS, we only report the normalized accuracy for the correlation coefficient.

## Replicating Experiments With Generative Agents

Do generative agents predict the behavior of sample average treatment effects? To study this, we use the responses from the human participants in our study to directly run a replication of the five sampled experiments, rather than using the published results. This direct replication is important because our sample might be different, and some studies may not replicate. To the extent that some of these studies are not replicated by the agents, this approach allows us to distinguish whether the failed replication was due to the original study not replicating, or whether the failure to replicate instead lay with the agents.

In this evaluation, we are interested in three related outcomes: whether the effect direction and significance replicate, and whether there is a correlation between effect sizes. The five studies have different statistics, so we use the same statistical methods as in the original papers to derive the p-values for the effects' significance. We briefly describe these methods here. Consistent with the original replication studies conducted by Camerer et al., we calculate and compare all effect sizes in terms of Cohen's *d*.

- Ames & Fiske, 2015 *(33)*: A chi-square test ($\chi$2-test) of equal proportions to evaluate the hypothesis by comparing the number of participants choosing the blame task between the two experimental conditions.
- Cooney et al., 2016 *(34)*: An F-test on the Outcome x Procedure interaction within a 2x2 ANOVA design, which evaluates how the predicted feelings about fairness (receiving or not receiving a bonus) interact with the allocation procedure (fair or unfair).
- Halevy & Halali, 2015 *(35)*: An independent samples t-test to compare the perceived benefits of intervening in a conflict between two friends across two treatment groups (those who did intervene and those who did not).
- Rai et al., 2017 *(36)*: An independent samples t-test to compare participants' willingness to harm a stranger described in humanized versus dehumanized terms under instrumental motives.
- Schilke et al., 2015 *(37)*: A chi-square test ($\chi$2-test) to compare the levels of trust between participants with high structural power and those with low structural power.

We report which of the studies our human participants and simulated predictions replicated with significant results. To understand how well the effect sizes are correlated, we calculate the Pearson correlation coefficient between the effect sizes from the human participants and simulated participants.

<u>Interviews Improve Agents' Prediction Accuracy</u>

Interviews provide qualitative data expressed in participants' own words, but to what extent do interviews improve agents' predictive capabilities? To address this question, we present two sets of analyses:

1) A main analysis where we compare the interview agents we presented in the main paper to agents created using the known practices from recent literature that studied human behavioral simulations with language models.
2) Exploratory analyses conducted on a random subset of 100 agents in the agent bank, investigating a broader range of design spaces. This includes examining generative agents with interview lesions and agents informed by survey data instead of interview data.

The main analysis aims to establish a baseline for predictive performance grounded in prior literature and evaluate whether our agent architecture surpasses this benchmark.

*Main analysis.* Alongside interview based, survey based, and survey + interview based, we evaluate two alternative agent descriptors: participants' demographic information and persona descriptions. To operationalize demographic information, we reconstruct agent descriptions following a method representative of approaches used in recent literature that employ language models to simulate human behavior. This method, similar to that presented by Argyle et al., constructs a first-person descriptor including political ideology, race, gender, and age (*34*). For example:

```
Ideologically, I describe myself as conservative. Politically, I
am a strong Republican. Racially, I am white. I am male. In terms
of age, I am 50 years old.
```

We reconstruct these descriptors for our participants using their responses to the GSS. Then, to prompt the language model with this data, we replace any other self-report data with the demographic descriptors.

To operationalize persona descriptions, we asked participants to write a short paragraph about themselves at the end of their phase 1 participation in surveys and experiments. They were instructed to describe themselves as they might to a stranger, including information about their personal background, personality, and demographics. For instance:

```
I am a 20 year old from new york city. I come from a working
middle class family, I like sports and nature a lot. I also like
to travel and experience new restaurants and cafes. I love to
spend time with my family and I value them a lot, I also have a
lot of friends which are very kind and caring. I live in the city
but I love to spend time outdoors and in nature. I live with both
parents and my younger sister.
```

Given this, we leverage both demographic agents and persona agents to predict participants' responses at the individual level. To compare performance differences between these agent versions, we conducted an ANOVA with post hoc Tukey tests, examining differences between architectures on the main evaluation metrics. This comparison was made between agents created with interviews, demographic information, persona descriptions, and survey data.

*Exploratory robustness analysis.* The primary goal for this exploratory analysis is to understand, in a more exhaustive manner, why the interview and survey based agents are performant. We conducted this analysis on a random sample of 100 agents from our agent bank, comparing agents informed by different data sources. Each comparison aims to shed light on a particular question about the efficacy of agents with richer background information in generating predictive agents. We compare the following agent types in this robustness check (results shown in SM table 6):

- **Survey Agents**. To investigate whether interviews are a uniquely powerful medium or if similar information can be captured through surveys, we created composite descriptions by compiling participants' responses to two components used in our study—the GSS and Big Five personality test. This approach aligns with traditional social science methods, which have developed and validated structured surveys to efficiently gather information from participants. We aimed to determine if these established methods capture comparable information to our interview-based agents. In other words, do the interviews capture the same predictive power, more, or less, than these common validated instruments?

Survey agents were constructed using respondents' wave 1 answers to the GSS and Big-5, and evaluated on their wave 2 responses. For the GSS, we test two different strategies for constructing the agents. First, we remove only the outcome question we predict from the input data for the agent. By doing so, the agent will have access to information related to the outcome questions. For example, when predicting how an agent will respond to: "If a man and woman have sex relations before marriage, do you think it is always wrong, almost always wrong, wrong only sometimes, or not wrong at all?", the agent will know the answer to "What if they are in their early teens, say 14 to 16 years old? In that case, do you think sex relations before marriage are always wrong, almost always wrong, wrong only sometimes, or not wrong at all?"

Second, we hold-out the whole GSS module a given outcome question is in. This approach means the survey agents have fewer access to similar questions than the interview agents have, whereas the first strategy ensures the survey agents have access to more similar questions than the interview agents. For the Big-5 we always hold-out the whole block of questions asking about a particular personality trait when predicting an outcome question within that trait.

Compared to the interview agents, the first approach gives the survey agents more access to questions similar to the one being predicted: questions within the same GSS block tap similar topics and are asked in a similar close-ended fashion, whereas American Voices Project and GSS use different questioning styles. In the second approach, survey based agents have access to fewer similar questions, as there is little thematic overlap between GSS modules, while the American Voices Project and GSS do touch upon similar topics.

- **Survey agents with interviews.** To assess whether the information captured in interviews fully encompasses that from surveys and experiments, we add the interviews to the survey agents.

- **Maximal Agents**. Maximal agents combine all the data we have by adding the economic games alongside the interviews to the survey agents. By equipping these agents with access to all data sources, we aim to determine our current performance ceiling: do interviews alone capture the most relevant information, or is there added value in combining methods?

- **Summary Agents.** We investigated whether the predictive power of interviews stems from linguistic cues in the transcript, or from the information in the transcript . To explore this, we created summary agents by prompting GPT-4o to convert interview transcripts into bullet-pointed dictionaries of key response pairs, capturing the factual content while removing most linguistic features (e.g., `{ "childhood_town": "Small town", "siblings": "Only child", "marital_history": "Married twice", "children": "Two children, but they are not living with the interviewee"...}`). By isolating the factual knowledge from the unique linguistic elements, we aimed to determine whether the predictive accuracy of the agents relies on those linguistic nuances. If the summary agents perform worse than the full interview-based agents, it would suggest that linguistic features play a key role in enhancing prediction accuracy.

- **Random Lesion Interview Agents**. We investigated the efficiency of interviews in providing insights compared to shorter data collection methods. While surveys such as the GSS Core typically take about 30 minutes, our interviews span two hours. To assess how much interview content is necessary to convey meaningful insights, we developed random lesion agents. These agents were created by progressively shortening interviews, randomly removing 0%, 20%, 40%, 60%, and 80% of the question-response pairs. This approach allowed us to identify the threshold at which the interview becomes too sparse to retain valuable information.

Demographic Bias in Agent Predictions

How do individuals' demographic attributes interact with the generative agents' ability to predict their respective behaviors and attitudes? Here, we investigated how individuals' demographic attributes interact with the generative agents' ability to predict their behaviors and attitudes. Specifically, we aimed to assess whether creating individualized models based on self-report data reduces performance gaps across demographic groups, compared to relying solely on demographic attributes or personas. To quantify these differences, we employed the Demographic Parity Difference (DPD), a metric used in machine learning fairness literature *(44, 45)*. DPD measures the difference in prediction rates between demographic groups, with lower values indicating more equitable predictions.

We calculated DPD for three constructs—General Social Survey (GSS), Big Five personality traits, and economic games—across five agent conditions: interview-based, demographic-based, persona-based, survey+interview based, and survey based. The interview-based condition utilized our novel approach, while the demographic and persona conditions served as baselines from prior literature. Using demographic attributes provided by our recruitment vendor, Bovitz, we calculated accuracy metrics for subpopulations across Studies 1 and 2. We compared these metrics across the five agent conditions to determine if certain populations experienced worse predictive accuracy, to quantify the extent of disparities, and to examine how these differences interacted with the agent condition. DPD values were reported alongside individual accuracy results for different population subgroups.

In addition, to assess the statistical significance of the observed differences, we conducted regression analyses. These analyses explored the impact of demographic variables on the predictive performance of generative agents, with separate regression models run for each demographic variable, using predictive performance as the dependent variable.

## 6. Supplementary Results

In this section, we present a higher-level interpretation of supplementary results that, while not central to our main findings, offer valuable insights. The methods are as outlined in the previous section, and detailed tables of these results can be found in Section 10.

Numerical GSS Prediction Results

In addition to our predictive performance on the primary categorical questions of the General Social Survey (GSS), we also evaluated our model's ability to predict responses to the six GSS core numerical questions. The results show similarly strong predictive accuracy, with an average correlation of r = 0.97 (std = 0.82) and an average mean absolute error (MAE) of 0.14 (std = 0.15). These metrics indicate a high degree of alignment between our predictions and the actual GSS responses for the numerical questions.

Exploratory Robustness Analysis Results

We present supplementary findings from the exploratory robustness analysis, organized by the materials used to inform agent behavior. Detailed results of this analysis are provided in Table 6.

- *Interview agents:* On the specific subsample of 100 interview agents we used for these robustness checks, the interview agents score a normalized accuracy of 0.82 (std = 0.13) on the GSS.
- *Survey Agents:* When excluding question-answer pairs from the same category as the predicted question, survey agents performed similarly compared to interview-based agents. Specifically, the survey agents reached a normalized accuracy on the GSS of 0.82 (std = 0.18). When removing the whole GSS module a given outcome question is in from the input when predicting on that question, survey agents are worse and achieved an average normalized accuracy of 0.77 (std = 0.12) on the GSS.
- *Survey plus interviews:* When including the interview transcripts to survey agents with only the GSS outcome question removed from the input, performance increases (0.85, std = 0.19). This finding suggests that adding more information does add additional predictive power above and beyond the interviews.
- *Maximal Agents.* Maximal agents, which incorporated information from surveys, experiments, and interviews, achieved slightly better performance than interview-based agents, with a normalized accuracy of 0.87 (std = 0.13) on the GSS. This finding reinforces that more information improves predictive performance.
- *Summary Agents.* The summary agents performed slightly below the interview agents, with a normalized accuracy of 0.81 (std = 0.13) on the GSS. This indicates that while some information may be lost from the linguistic cues during summarization, much of the performance is due to the information in the interview rather than the low level language that participants use.
- *Random Lesion Interview Agents.* Performance declined linearly as we removed increasing portions of the interview data. Starting with a normalized accuracy of 0.82 (std = 0.13) when no information was removed, accuracy dropped to 0.79 (std = 0.12) when 80% of the utterances were excluded. This suggests that although performance decreases as interview length is reduced, even a short interview contains sufficient richness to outperform agents informed solely by surveys and experiments, highlighting the efficiency of interviews in identifying valuable insights.

Demographic Bias in Agent Predictions

We present supplementary findings on bias in agent predictions. The complete results of our regression analysis can be found in Table 4, with demographic parity differences (DPD) detailed in Table 5. In addition to the DPD findings outlined in the main document, our regression analysis indicates that significant demographic performance discrepancies across constructs were relatively rare. However, notable exceptions emerged for political ideology, party affiliation, and sexual orientation when predicting GSS responses. Specifically, predictive performance was noticeably stronger for participants identifying as strong liberals, strong Democrats, and non-heterosexual/straight, compared to those identifying as more conservative, Republican, or heterosexual/straight.

## 7. Deviations from the Pre-Analysis-Plan for the GSS Outcome Questions

We reduced the set of GSS outcome questions by removing those GSS questions for which there is a synonymous question in the American Voices Project (AVP). While we pre-registered that we would use the full GSS survey to evaluate the agents' predictions, this left open the possibility that agents with access to the AVP interview (interview and survey-plus-interview agents) could achieve high accuracy simply because some GSS questions were nearly identical to those asked in the interviews. To address this potential concern, we developed a procedure to identify and remove from the input prompt such overlapping questions from the evaluation set. This procedure involved three steps.

First, we created all possible pairs of GSS and interview questions, resulting in 54,694 unique pairs. We then used a GPT-4.1 model to classify whether each pair of questions is synonymous. Research has highlighted that GPT is good at such text classification tasks *(80)*. The model was instructed to provide a binary judgment of whether two question pairs were synonymous or not and an associated score capturing how similar the questions are, which we used as a confidence score.[1] After manually confirming the questions the model flagged as similar following the same coding criteria as the GPT model, this process identified 24 GSS outcome questions for which

[1] The exact prompting instructions read: "
'[survey question]'

Task: What you see above is a survey question. Based on the survey question, I want you to assess whether the survey question is essentially the same as the interview question listed below.

As you answer, I want you to take the following steps:
Step 1) Return a score from 0 to 100 indicating how similar the two questions are, where 0 is not at all similar and 100 is 'they are the same question'.
Step 2) Return an answer 'Same' if the question asks about the same information or 'Different' otherwise.

Here is the interview question: '[interview question]'"

there was an identical or nearly identical question in the interview transcripts. For example: "Did your mother ever work for pay for as long as a year, while you were growing up?" and "Did your mom work for pay for at least a year while you were growing up?"

Of these 24 questions, 15 ask about the respondent's own demographic characteristics (e.g., employment, gender, education, religion), five concern parental education or employment, and the remaining questions address aspects such as spousal traits, and partisanship.

Second, we assessed the risk of false negatives—cases where the model failed to flag identical or nearly identical questions. To do this we: 1) manually reviewed all cases the model marked as "unsure," as captured by a score different from 0, and manually assessed whether question pairs were synonymous, following the same decision criteria as we instructed the model to follow. This process led us to identify two additional demographic questions to remove. 2) We hired a human coder to classify 500 question pairs (GSS and interview questions). This included all model-flagged cases and a stratified random sample of unflagged ones, oversampling cases where the model indicated low confidence by a factor of ten to create a set of questions that were most likely to include false negatives. The human coder confirmed 21 of the 24 questions flagged by the model and identified only one additional missed question (from the set of 500 pairs), which we then also removed. Identifying one false negative out of five hundred suggests that the model is highly robust to false negatives and that we successfully removed the right set of GSS questions. To be precise, treating the 1-in-500 miss rate as if it came from a simple random sample yields a margin of error of ±0.4 percentage points, so even an extreme upper bound puts the false-negative rate comfortably below 1%. To reiterate, we also manually checked the 24 questions the model flagged to ensure there are no false-positives.

Third, we excluded the final set of 27 questions from our GSS outcome analysis.[2] This revised set, reported in the main paper, is less likely to give rise to inflated agent accuracy rates due to direct retrieval of answers to questions. For completeness, the pre-registered models using the full set of outcome questions are reported below in figure 5. Accuracy is 0.02 points higher when including GSS questions that are identical to American Voices Project questions.

Crucially, we repeated the same question-identification procedure to ensure that survey agents do not have questions in the input that are identical to those in the output. We found that no two questions in the GSS are synonymous with each other, which is to be expected considering the optimization for survey time with this instrument.

[2] Specifically, we remove the following GSS outcome questions:
BORN, DEGREE, DWELOWN, EDUC, FAMDIF16, HISPANIC, MADEG, MAEDUC, MARITAL, MARTYPE, MAWRKGRW, PADEG, PAEDUC, PARTYID, RACE, REG16, RELIG, RVISITOR, RELPERSN, SEX, SPEDUC, SPRTPRSN, SPWRKSTA, VETYEARS, VISITORS, WIDOWED, ZODIAC

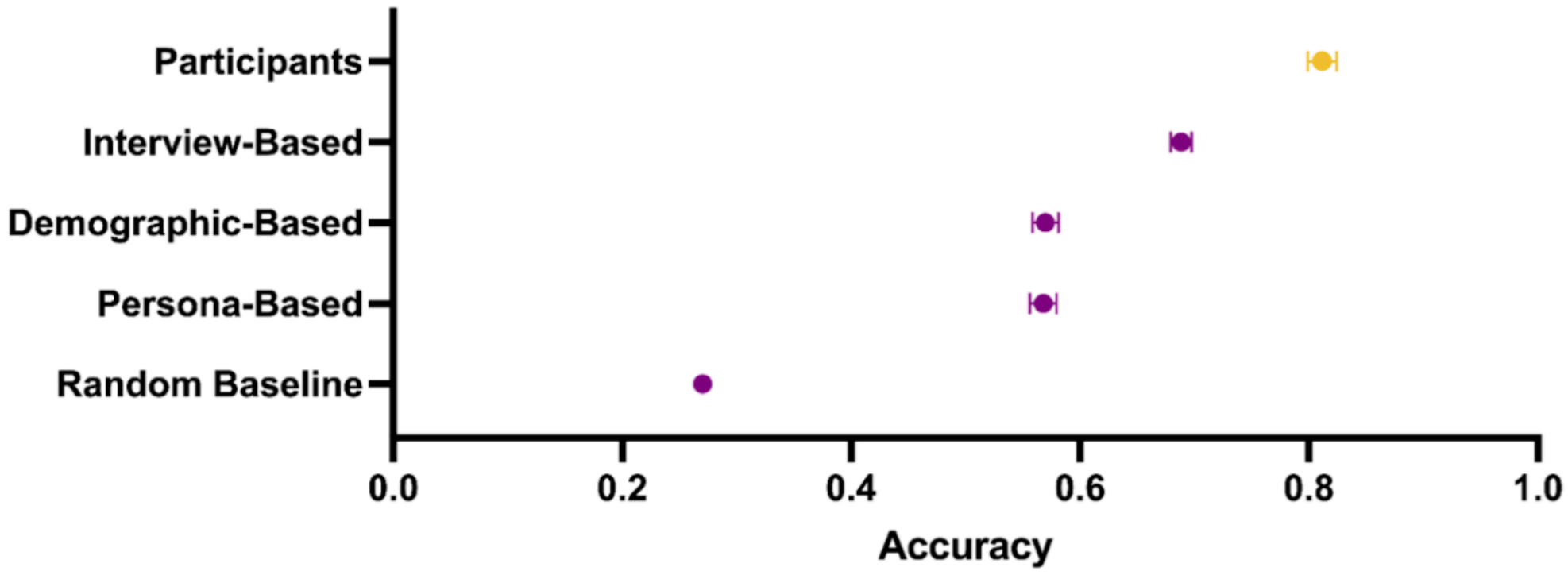

**Figure 5.** GSS results from figure 2 in the main paper yet with all GSS outcome questions (as pre-registered).

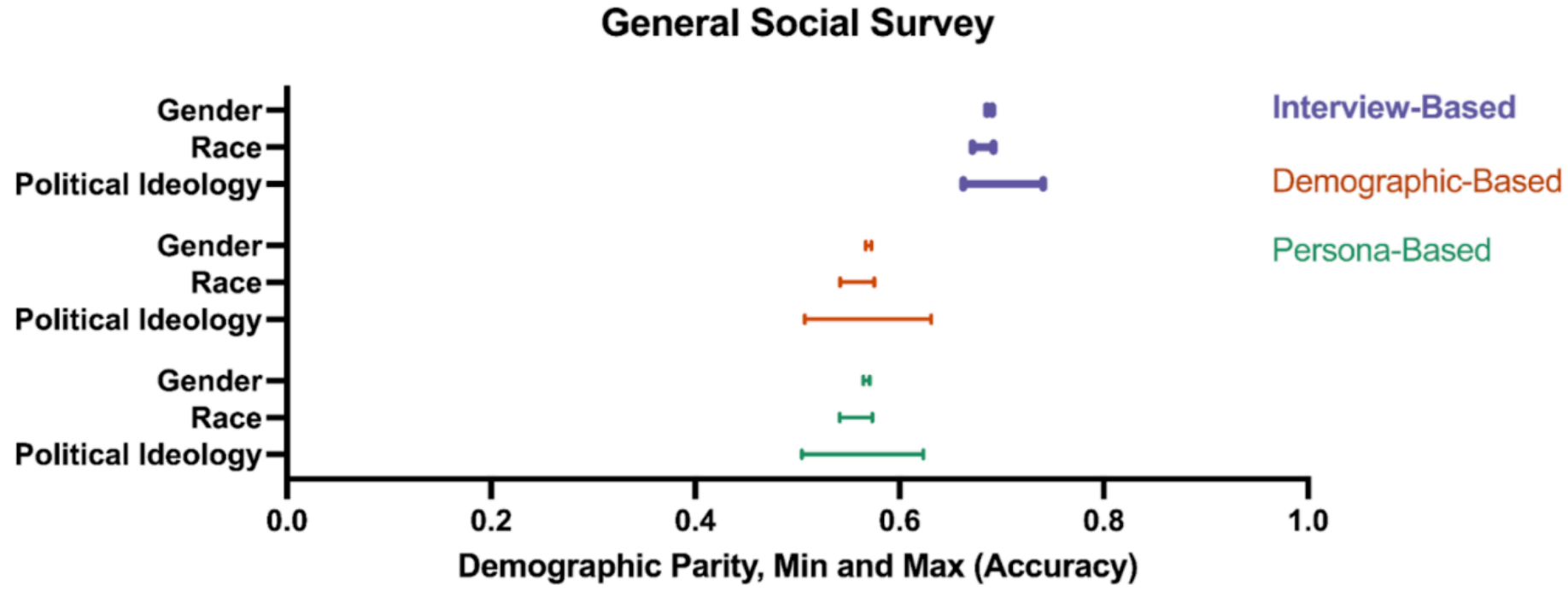

**Figure 6.** GSS results from Figure 3 in the main paper yet with all GSS outcome questions (as pre-registered).

The complete list of 27 question pairs for which we removed the GSS question from the evaluation set is below:

| GSS question text | AVP question text |
| --- | --- |
| Were you born in this country (U.S.)? | Were you born in the United States? |
| Your highest degree received | What is the highest degree or grade you've completed? |
| Do you/Does your family own your home/apartment, pay rent, or what? | Does your household own or rent this home? |
| What is the highest grade in elementary school or high school that y ou finished and got credit for? | What is the highest degree or grade you've completed? |
| Were you living with both your own mother and father around the time you were 16? If not living with both own mother and father: what happened? | Did you live with both your own mother and father when you were 16? |
| Are you Spanish, Hispanic, or Latino/Latina? | Are you of Hispanic, Latinx, or Spanish origin? |
| Your mother's highest degree received | What's the highest degree or grade your mom [OR ADULT FEMALE LIVED WITH AT 16] completed? |
| What is the highest grade in elementary school or high school that your mother finished and got credit for? | What's the highest degree or grade your mom [OR ADULT FEMALE LIVED WITH AT 16] completed? |
| Are you currently -- married, widowed, divorced, separated, or have you never been married? | Are you now married, widowed, divorced, separated, or have you never been married? |
| Your marital type | Are you now married, widowed, divorced, separated, or have you never been married? |
| Did your mother ever work for pay for as long as a year, while you were growing up? | Did your mom [FIGURE AT AGE 16] work for pay for at least a year while you were growing up? |
| Your father's highest degree received | What's the highest degree or grade your dad [OR ADULT MALE LIVED WITH AT 16] completed? |
| What is the highest grade in elementary school or high school that your father finished and got credit for? | What's the highest degree or grade your dad [OR ADULT MALE LIVED WITH AT 16] completed? |
| Generally speaking, do you usually think of yourself as a Republican, Democrat, Independent, or what? | Generally speaking, do you usually think of yourself as a Democrat, a Republican, an Independent, or what? |
| What race do you consider yourself? | What race or races do you identify with? |
| In what state or foreign country were you living when you were 16 years old? | What was the address that you lived at when you were 16 years old? |
| What is your religious preference? Is it Protestant, Catholic, Jewish, some other religion, or no religion? | What religion do you identify with, if any? |
| To what extent do you consider yourself a religious person? Are you . . . | Some people say that religion or spirituality is important in their lives, for others not so much. How about you? |
| Your sex | Which gender do you identify with? |
| What is the highest grade in elementary school or high school that your spouse finished and got credit for? | What about [PERSON X]? [Figure out the education of every person in someone's household] |
| To what extent do you consider yourself a spiritual person? Are you . . . | Some people say that religion or spirituality is important in their lives, for others not so much. How about you? |

| Last week, was your spouse working full time, part time, going to school, keeping house, or what? | [IF APPLICABLE] Tell me all about any work [SPOUSE/PARTNER NAME] has right now. [REPEAT PROBES ABOVE] |
|---|---|
| Have you ever been on active duty for military training or service for two consecutive months or more? If yes: what was your total time on active duty? | Have you ever served on active duty in the U.S. Armed Forces, Reserves, or National Guard? |
| How many visitors are there in your household? & Are you a visitor in your household? | Tell me about the people who live with you right now, even people who are staying here temporarily. |
| Have you ever been widowed? | Are you now married, widowed, divorced, separated, or have you never been married? |
| Your astrological sign | [removed because we ask for birthday] |

## 8. Why do interview-based generative agents work?

Prior work provides frameworks for understanding the limits of survey data's predictive accuracy and has found similar performance of LLM-based predictions on hold-out samples of survey questions[3]. A key open question remains: why does the interview data contribute additional accuracy gains? To investigate why interview-based generative agents outperform agents prompted only with demographic information, we run a series of robustness checks, each designed to confirm or rule out a specific mechanism behind interview-based agents' superior accuracy. We outline these tests below.

*Mechanism 1: Direct retrieval of answers.*
The most straightforward possibility is that the model can directly locate verbatim answers to outcome questions within the interview transcript. Qualitative interviews are designed to elicit rich narratives, which may incidentally contain the factual content queried by, for example, the GSS. This can occur even when the interview and GSS questions differ in wording (a divergence we confirmed; see §7 of the SI). This feature sets qualitative interviews apart from fixed-format questionnaires, which typically aim for concise responses to narrowly framed questions. Historically, leveraging the rich textual data from interviews for structured quantitative analysis required labor-intensive hand coding, but LLMs facilitate scalable processing of such texts. In short, interview-based agents may perform well because they can retrieve direct answers from the transcript. Consider the following example, based on the GSS item "Are you employed?" In response to an interview question about health, a participant might say: "I am disabled. My health is so bad that I can't even work." Despite the difference in questions, this answer can be directly retrieved as a valid response to the GSS item.

[3] I. Lundberg et al., The origins of unpredictability in life outcome prediction tasks. *Proc. Natl. Acad. Sci.* U.S.A. 121, e2322973121 (2024).; O. Toubia, G. Z. Gui, T. Peng, D. J. Merlau, A. Li, H. Chen, Twin-2K-500: A data set for building digital twins of over 2,000 people based on their answers to over 500 questions. *Marketing Science* 44, 1446-1455 (2025).

To test this mechanism, we implement a two-step procedure. First, for a random subset of 59 agents, we generate all possible pairs between GSS outcome questions and interview-question answers, resulting in an average of 21,713 pairs per agent.[4] For each pair, we prompt a GPT-4o-mini model to classify whether the interview answer contains the information needed to answer the GSS question. This allows us to compute, for each GSS question, the proportion of agents for which the question was flagged as directly answerable. Second, we calculate each agent's normalized accuracy on the GSS while progressively removing outcome questions, starting with those most likely to have a verbatim answer in the transcript. We then compare the performance of full interview agents to that of demographic and persona agents, who lack access to the transcript. If direct retrieval plays a role, we expect that interview agents will lose accuracy as questions with retrievable answers are removed, while demographic and persona agents will remain unaffected.

[4] Note that for this subset of 59 agents we ran 1,281,040 API queries. To minimize cost, we do not run this analysis on the full sample. We ended up with an odd number because our API query for 60 agents was interrupted and needed to be re-started. As the costs of this analysis are so high, and the larger the subsample the more representative of the full sample it is, we did not want to throw this data away.

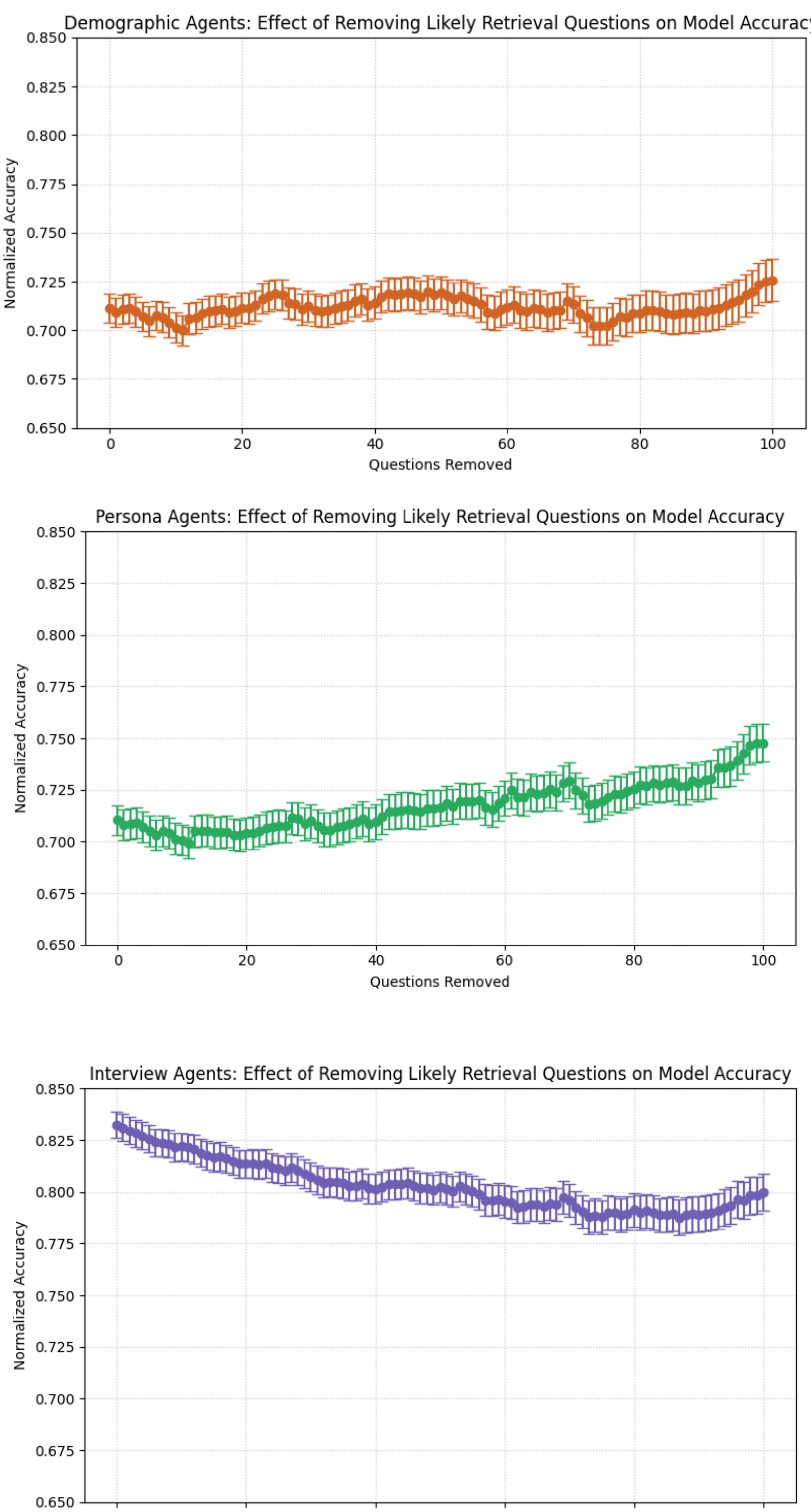

**Figure 7**. Demographic agents (top), persona agents (middle), interview agents (bottom). The GSS-outcome questions we remove (x-axis) when calculating the normalized accuracy (y-axis) are ranked from most to least likely to have a verbatim answer to them somewhere in the interview transcript.

The results of this analysis are shown in Figure 7. For the interview agents (bottom facet), we observe that dropping questions more likely to have a verbatim answer in the interview transcript (those on the left of the x-axis) leads to a decline in normalized accuracy (y-axis). This pattern does not hold for the demographic and persona agents, which do not have access to the interview transcript. This suggests that direct answer retrieval is a key mechanism behind the higher performance of interview-based agents. At the same time, full interview agents continue to outperform the other agents even after many GSS questions are removed. For example, when 80 questions are dropped, interview agents still achieve an accuracy of 0.79, compared to 0.71 for demographic agents. This sustained performance advantage indicates that other mechanisms, beyond direct retrieval, also contribute to the superior accuracy of interview agents.

*Mechanism 2: Reasoning based on other answers.*

Next, we examine a second mechanism: even if the answer to a GSS question is not explicitly stated in the interview transcript, full agents may still infer it from other information in the transcript. For example, consider the GSS question: "Do you have a supervisor on your job to whom you are directly responsible?" and the interview question and answer: "Q: Are you enrolled in school? A: Yes, I'm at school." In this case, the LLM may infer that someone who is currently enrolled in school is unlikely to be employed, and therefore unlikely to have a supervisor. In other words, the model's ability to reason based on qualitative interview content may be another key mechanism behind the superior performance of interview-based generative agents.

To test this mechanism, we follow a two-step procedure similar to that used for direct retrieval. First, using the same subset of 59 agents and the same set of GSS outcome questions and interview-answer pairs, we prompt a GPT-4o-mini model to classify whether it can answer the GSS question based on the given interview question and answer. As before, we compute, for each GSS question, the proportion of agents for which it was flagged as answerable. Compared to the previous test, we find that approximately three times as many questions are flagged. Second, we compute the normalized accuracy of all agents on the GSS while progressively removing outcome questions—starting with those most likely to be answerable through inference from the interview transcript. We then compare the performance of full agents to that of demographic and persona agents, which do not have access to interview transcripts. If inference plays a role, we expect interview agents to perform worse as more inferable questions are dropped, while the performance of other agents remains stable.

The results are shown in Figure 8. As with the direct retrieval test, we find that agents without access to interview transcripts perform similarly regardless of which GSS items are removed. Interview agents, by contrast, get progressively worse as we drop questions that are most likely to be answerable through inference.

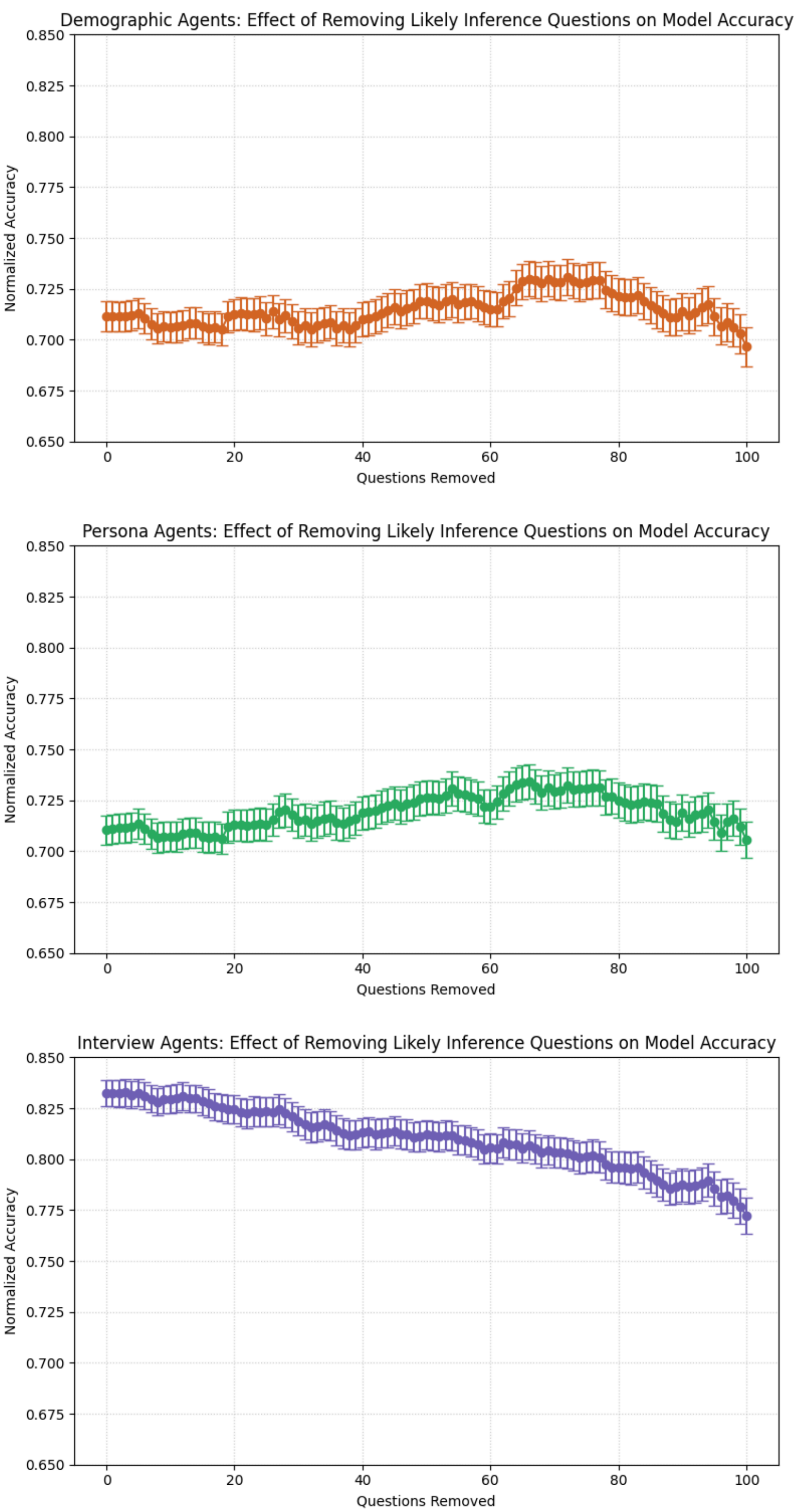

**Figure 8**. Demographic agents (top), persona agents (middle), interview agents (bottom). The GSS-outcome questions we remove (x-axis) when calculating the normalized accuracy (y-axis) are ranked from most to least likely to be able to be answered through inference based on information somewhere in the interview transcript.

One potential complication in interpreting these results is that (nearly) all instances of direct retrieval also allow for inference, as shown in the table below. To better isolate the role of inference relative to direct retrieval, we calculate normalized accuracy after removing the 30 GSS questions most likely to involve direct retrieval. We then compute accuracy again after removing both these 30 questions and an additional 30 questions most likely to be answerable through inference. This approach ensures that any observed drop in performance when inference questions are removed is not simply driven by overlap with direct retrieval cases. The results indicate that, after removing direct-retrieval questions, interview agents outperform demographic agents by 0.095 points. When inference questions are also removed, this gap shrinks to 0.080—suggesting that inference, independent of direct retrieval, contributes meaningfully to agent performance.

| | Inferred | Not inferred |
|---|---|---|
| Direct retrieval | 4649 | 116 |
| No direct retrieval | 44596 | 1,221,204 |

Overlap between GSS-outcome question and interview question pairs where a GPT-4o-mini model has classified that the answer to the GSS question can be directly retrieved or inferred from the answer to the interview question.

Taken together, these results suggest both mechanisms play a role in the predictive accuracy of generative agents built from in-depth interview transcripts, i.e., interview-based generative agents likely outperform demographic and persona agents because they uniquely combine the depth of semi-structured interviews with the reasoning and retrieval capabilities of large language models. Open-ended interviews elicit far richer and more nuanced information than closed-ended surveys, capturing not just life facts and attitudes in a structured format, but also individuals' reasoning behind such attitudes, long-tail life experiences that are not asked about in surveys, and the interwoven narratives of identity, belonging, and personal history that illuminate the complex foundations of people's attitudes. Large language models, in turn, are able to process this wealth of information to directly retrieve answers embedded anywhere in the text and to infer plausible answers even when they are not stated verbatim. Crucially, our results show that these two mechanisms—direct retrieval and inference—do not fully account for the superior performance of interview-based agents: even after excluding questions most susceptible to these mechanisms, interview-based agents still outperform the alternatives. This suggests that additional mechanisms remain to be identified, highlighting a valuable direction for future research.

## 9. Results using different models

Throughout the main text we relied on GPT-4o because, at the time of data collection, it offered the best balance between model capacity and computational cost. Four questions nevertheless remain:

1. **Model tier:** Would a more sophisticated—but also more expensive—reasoning model (e.g., o3, or o1) or GPT model (i.e. 4.1, 5) yield meaningfully higher accuracy?
2. **Model economy:** Does the richness of the 6 k–word interview context allow substantially cheaper models (e.g., GPT4o-mini) to approach GPT-4o performance?
3. **Model drift:** Are our findings specific to the exact instance of GPT-4o we used, or do they hold for newer instances released since?
4. **Finetuning:** Does fine-tuning a model on a test set improve how well it predicts on a training set?

To address points one through three we replicated the General Social Survey (GSS) prediction task on a subsample of 50 generative agents—we use 50 agents only because the costs to run the full interviews through some of these more advanced models are substantial. All prompt templates, chain-of-thought instructions, and temperature settings were held constant across models; only the model parameter changed. We report the non-normalized accuracy score. These results are shown in Figure 9.

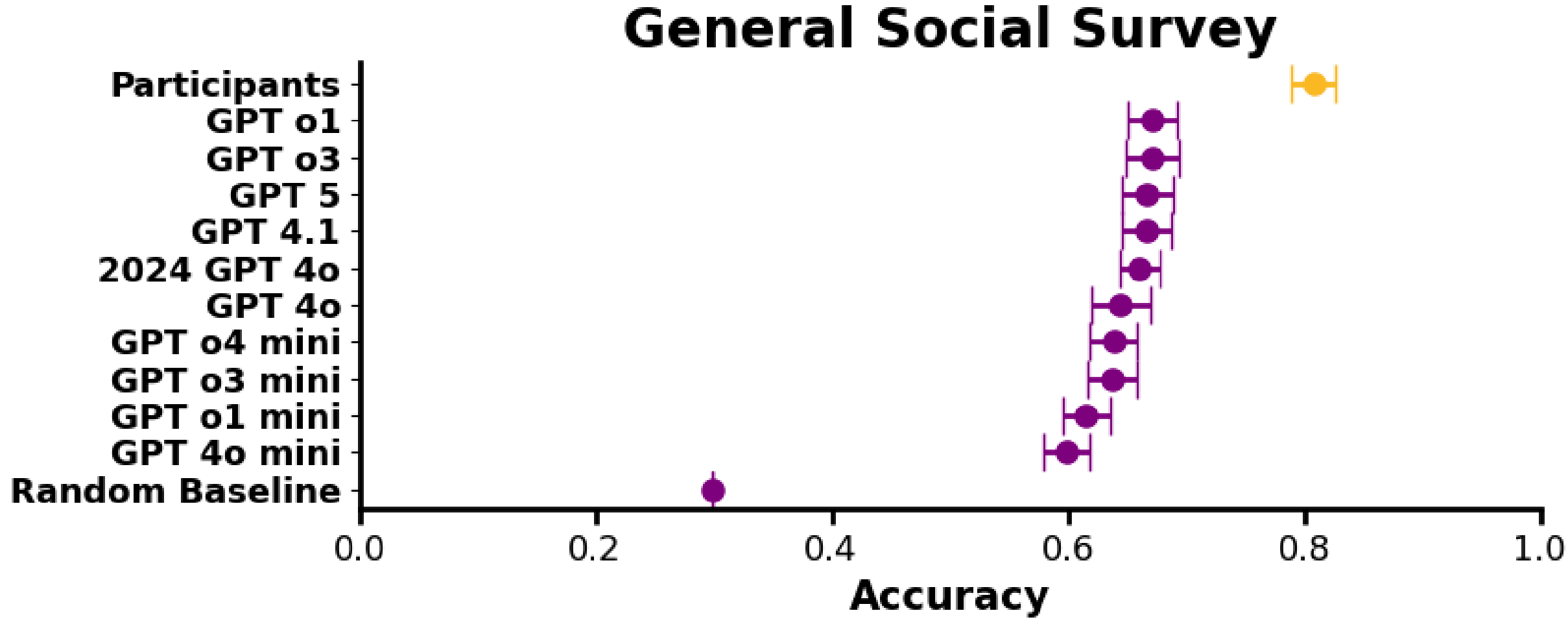

**Figure 9.** GSS (repeated questions omitted) results predicted results using 50 agents on different GPT models.

Taken together, the results paint a consistent picture. The more advanced (reasoning) models—GPT-5 (0.67), GPT-4.1 (0.67), GPT-o1 (0.67), and GPT-o3 (0.67)—outperform the 2024 GPT-4o baseline reported throughout the paper (0.66) in accuracy, but the advantage is small and statistically indistinguishable with just 50 agents. The lighter "mini" variants show the trade-off between cost and performance: GPT-4o-mini (0.60) and GPT-o1-mini (0.62) trail their full counterparts by roughly two to five points, while GPT-o3-mini (0.64) and GPT-o4-mini (0.64) nearly match the GPT-4o baseline. Should model prices drop further, the small—but consistent—accuracy gains of the advanced tiers could become attractive. Finally, model drift appears minimal: a run of GPT-4o in 2025 reproduces the original accuracy from 2024 almost identically, with an 0.02 point decrease in performance for the newer instance of 4o (0.64 vs 0.66).

Next, we explored the impact of fine-tuning focused specifically on the outcome questions from the GSS. We used OpenAI's out-of-the-box fine-tuning tool with one epoch and modified the prompt to remove the reasoning and reflection components from the output, as there is no ground truth for those. For 500 randomly selected agents, we constructed both the prompts and the correct answers to the GSS questions. We then fine-tuned GPT-4o on these 500 agents and evaluated it on the remaining 552 agents in the dataset. We compared the performance of these agents to the same 552 agents in the main GSS analyses (which included reasoning and reflection components in the prompt), as well as to a non-fine-tuned GPT-4o model run using the exact same prompt structure that we used for the fine-tuned model. The fine-tuned model achieved a normalized accuracy of 0.79 (std = 0.07) and a normalized correlation of 0.81 (std = 0.13). Without fine-tuning, using the same prompting strategy as in the main paper (with reasoning and reflection), the model achieved a normalized accuracy of 0.84 (std = 0.11) and a normalized correlation of 0.87 (std = 0.15). Without fine-tuning, but using the exact same prompts as those used to test the fine-tuned model, we found a normalized accuracy of 0.82 (std = 0.08) and a normalized correlation of 0.85 (std = 0.13). Overall, fine-tuning does not appear to improve performance.

## 10. Supplementary Tables

| Age | 18 to 24 | 25 to 34 | 35 to 44 | 45 to 54 | 55 to 64 | 65 to 74 |
|---|---|---|---|---|---|---|
| | 11.03% | 13.88% | 17.49% | 19.77% | 21.48% | 13.50% |
| | 75 or more | | | | | |
| | 2.85% | | | | | |
| **Census division** | New England | Middle Atlantic | E.N. Central | W.N. Central | South Atlantic | E.S. Central |
| | 6.65% | 12.83% | 18.73% | 8.08% | 10.08% | 11.5% |
| | W.S. Central | Mountain | Pacific | Foreign | | |
| | 8.65% | 5.13% | 15.78% | 2.57% | | |
| **Education** | Less than high school graduate | High school graduate | Associate/junior college | Bachelor's degree | Graduate degree | |
| | 2.28% | 38.88% | 17.59% | 26.9% | 14.35% | |
| **Race** | White | Black | Other | | | |
| | 75.95% | 14.26% | 9.79% | | | |
| **Ethnicity** | White/Caucasian | Black/African American | Asian | Native Hawaiian or Pacific Isl. | American Indian or Alaskan nat. | Other ethnicity |
| | 79.18% | 14.64% | 5.04% | 0.48% | 2.76% | 5.80% |
| **Sexuality** | Heterosexual /straight | Gay or lesbian | Bisexual | Asexual | Pansexual | Other sexual orientation |
| | 82.2% | 4.36% | 8.04% | 1.72% | 2.41% | 1.15% |
| **Gender** | Female | Male | | | | |
| | 56.37% | 43.63% | | | | |
| **Income** | Less than $25,000 | $25,000 to $34,999 | $35,000 to $49,999 | $50,000 to $74,999 | $75,000 to $99,999 | $100,000 to $124,999 |
| | 18.83% | 11.83% | 13.89% | 20.44% | 14.7% | 8.04% |
| | $125,000 to $149,999 | $150,000 to $174,999 | $175,000 to $199,999 | $200,000 to $249,999 | $250,000 or more | |
| | 5.05% | 2.18% | 1.61% | 1.38% | 2.18% | |
| **Neighborhood** | Urban | Suburban | Rural | | | |
| | 30.88% | 48.11% | 21.13% | | | |
| **Political ideology** | Extremely Liberal | Liberal | Slightly Liberal | Moderate | Slightly conservative | Conservative |
| | 11.31% | 19.01% | 9.32% | 28.8% | 8.94% | 16.83% |
| | Extremely conservative | | | | | |
| | 5.8% | | | | | |
| **Political party preference** | Strong Democrat | Democrat | Independent, close to Dem. | Independent | Independent, close to Rep. | Republican |
| | 21.96% | 13.31% | 11.88% | 15.59% | 8.46% | 11.6% |
| | Strong Republican | Other | | | | |
| | 14.83% | 2.38% | | | | |

**Table 1.** Demographic distribution of our 1,052 participants. Collectively, they represent a stratified sample of the U.S. demographic across age, gender, race, region of residence, level of education, and political identity. Note that for ethnicity, the participants could choose more than one option.

**[General Social Survey:** Agent Architecture Comparison (ANOVA)**]**

| Source | Sum of Squares (SS) | df | F | p-value |
|---|---|---|---|---|
| Group | 11.406 | 4 | 577.83 | < .001 |
| Residual | 25.933 | 5255 | N/A | N/A |

| Group 1 | Group 2 | Mean Difference | p-value | Lower Bound | Upper Bound | Reject Null |
|---|---|---|---|---|---|---|
| Demographics | Interview | 0.0755 | < .001 | 0.0671 | 0.0838 | Yes |
| Demographics | Persona | -0.0192 | < .001 | -0.0275 | -0.0108 | Yes |
| Demographics | Survey | 0.0685 | < .001 | 0.0602 | 0.0769 | Yes |
| Interview | Persona | -0.0946 | < .001 | -0.1030 | -0.0863 | Yes |
| Interview | Survey | -0.0069 | 0.1585 | -0.0153 | 0.0014 | No |
| Persona | Survey | 0.0877 | < .001 | 0.0794 | 0.0961 | Yes |
| Composite | Demographics | -0.1017 | < .001 | -0.1101 | -0.0934 | Yes |
| Composite | Interview | -0.0263 | < .001 | -0.0346 | -0.0179 | Yes |
| Composite | Persona | -0.1209 | < .001 | -0.1293 | -0.1125 | Yes |
| Composite | Survey | -0.0332 | < .001 | -0.0415 | -0.0248 | Yes |

**[Big Five Personality Traits:** Agent Architecture Comparison (ANOVA)**]**

| Source | Sum of Squares (SS) | df | F | p-value |
|---|---|---|---|---|
| Group | 19.754 | 4 | 26.66 | < .001 |
| Residual | 973.603 | 5255 | N/A | N/A |

| Group 1 | Group 2 | Mean Difference | p-value | Lower Bound | Upper Bound | Reject Null |
|---|---|---|---|---|---|---|
| Demographics | Interview | 0.1577 | < .001 | 0.1065 | 0.2089 | Yes |
| Demographics | Persona | 0.0743 | 0.0013 | 0.0202 | 0.1227 | Yes |
| Demographics | Survey | 0.0397 | 0.2141 | -0.0115 | 0.0909 | No |
| Interview | Persona | -0.0862 | < .001 | -0.1374 | -0.0350 | Yes |
| Interview | Survey | -0.1180 | < .001 | -0.1692 | -0.0668 | Yes |
| Persona | Survey | -0.0318 | 0.4386 | -0.0830 | 0.0194 | No |
| Composite | Demographics | -0.1486 | < .001 | -0.1998 | -0.0974 | Yes |
| Composite | Interview | 0.0091 | 0.9889 | -0.0421 | 0.0603 | No |
| Composite | Persona | -0.0772 | < .001 | -0.1284 | -0.0259 | Yes |
| Composite | Survey | -0.1089 | < .001 | -0.1601 | -0.0577 | Yes |

**[Economic Games:** Agent Architecture Comparison (ANOVA)**]**

| **Source** | **Sum of Squares (SS)** | **df** | **F** | **p-value** |
|---|---|---|---|---|
| Group | 2.832 | 4 | 1.63 | 0.1637 |
| Residual | 2282.838 | 5255 | N/A | N/A |

| **Group 1** | **Group 2** | **Mean Difference** | **p-value** | **Lower Bound** | **Upper Bound** | **Reject Null** |
|---|---|---|---|---|---|---|
| Demographics | Interview | 0.0525 | 0.3582 | -0.0259 | 0.1309 | No |
| Demographics | Persona | 0.0273 | 0.8766 | -0.0511 | 0.1058 | No |
| Demographics | Survey | -0.0040 | 0.9999 | -0.0825 | 0.0744 | No |
| Interview | Persona | -0.0252 | 0.9061 | -0.1036 | 0.0533 | No |
| Interview | Survey | -0.0565 | 0.2822 | -0.1350 | 0.0219 | No |
| Persona | Survey | -0.0314 | 0.8107 | -0.1098 | 0.0470 | No |
| Composite | Demographics | 0.0086 | 0.9982 | -0.0698 | 0.0871 | No |
| Composite | Interview | 0.0611 | 0.2085 | -0.0173 | 0.1396 | No |
| Composite | Persona | 0.0360 | 0.7206 | -0.0424 | 0.1144 | No |
| Composite | Survey | 0.0046 | 0.9999 | -0.0738 | 0.0830 | No |

**Table 2.** Comparison of the five agent specifications (one-way ANOVA with Tukey HSD post-hoc tests; "Composite" denotes the survey-plus-interview agent, and mean differences are group 2 minus group 1, with N = 1,052 per agent type). For the General Social Survey (in accuracy), interview, survey, and Composite agents each significantly outperformed both demographic- and persona-based agents, and Composite agents additionally outperformed both single sources (all $p < 0.001$). For the Big Five personality traits (in correlation), interview-only and Composite agents outperformed both baselines (all $p < 0.001$), whereas survey-only agents did not differ significantly from either. For the economic games (in correlation), no agent specification differed significantly from another ($F(4, 5255) = 1.63$, $p = 0.16$).

## [**General Social Survey:** Construct-Level Analysis]

| General Social Survey | Accuracy | Normalized accuracy | Correlation | Normalized Correlation |
|---|---|---|---|---|
| **natspac/y** | 0.33 | 0.45 | 0.17 | 0.25 |
| **natenvir/y** | 0.6 | 0.74 | 0.51 | 0.69 |
| **natheal/y** | 0.7 | 0.91 | 0.36 | 0.73 |
| **natcity/y** | 0.48 | 0.75 | 0.32 | 0.62 |
| **natdrug/y** | 0.56 | 0.78 | 0.29 | 0.5 |
| **nateduc/y** | 0.68 | 0.84 | 0.36 | 0.54 |
| **natrace/y** | 0.64 | 0.84 | 0.67 | 0.87 |
| **natarms/y** | 0.56 | 0.75 | 0.46 | 0.63 |
| **nataid/y** | 0.7 | 0.85 | 0.34 | 0.53 |
| **natfare/y** | 0.67 | 0.84 | 0.5 | 0.71 |
| **natroad** | 0.44 | 0.61 | 0.14 | 0.24 |
| **natsoc** | 0.56 | 0.72 | 0.21 | 0.36 |
| **natmass** | 0.52 | 0.71 | 0.37 | 0.62 |
| **natpark** | 0.54 | 0.75 | 0.22 | 0.43 |
| **natchld** | 0.58 | 0.78 | 0.43 | 0.69 |
| **natsci** | 0.48 | 0.69 | 0.32 | 0.55 |
| **natenrgy** | 0.54 | 0.78 | 0.43 | 0.7 |
| **uswary** | 0.51 | 0.62 | 0.15 | 0.24 |
| **prayer** | 0.75 | 0.91 | 0.4 | 0.7 |
| **courts** | 0.56 | 0.74 | 0.48 | 0.7 |
| **discaffw** | 0.43 | 0.7 | 0.37 | 0.59 |
| **discaffm** | 0.39 | 0.67 | 0.33 | 0.6 |
| **fehire** | 0.33 | 0.58 | 0.48 | 0.71 |
| **fechld** | 0.48 | 0.75 | 0.29 | 0.48 |
| **fepresch** | 0.43 | 0.64 | 0.29 | 0.43 |
| **fefam** | 0.43 | 0.65 | 0.45 | 0.62 |
| **fepol** | 0.78 | 0.88 | 0.28 | 0.39 |
| **reg16** | 0.67 | 0.85 | 0.6 | 0.81 |
| **mobile16** | 0.75 | 0.85 | 0.72 | 0.86 |
| **famdif16** | 0.89 | 0.97 | 0.75 | 0.89 |
| **incom16** | 0.53 | 0.71 | 0.53 | 0.73 |
| **dwelown16** | 0.76 | 0.82 | 0.59 | 0.67 |
| **paeduc*** | 0.86 | 0.96 | 0.66 | 0.85 |
| **padeg*** | 0.8 | 0.91 | 0.74 | 0.87 |
| **maeduc*** | 0.91 | 0.98 | 0.72 | 0.92 |
| **madeg*** | 0.84 | 0.97 | 0.8 | 0.96 |
| **mawrkgrw** | 0.97 | 1.05 | 0.91 | 1.2 |
| **marital** | 0.97 | 1.02 | 0.96 | 1.03 |
| **widowed** | 0.98 | 1.0 | 0.84 | 0.95 |
| **divorced** | 0.94 | 0.97 | 0.86 | 0.92 |

| **martype*** | 0.84 | 0.97 | 0.78 | 0.94 |
|---|---|---|---|---|
| **posslq/y** | 0.95 | 1.02 | 0.93 | 1.03 |
| **wrkstat** | 0.76 | 0.94 | 0.76 | 0.94 |
| **evwork** | 0.96 | 1.01 | 0.66 | 0.94 |
| **wrkgovt1** | 0.88 | 0.95 | 0.56 | 0.74 |
| **wrkgovt2** | 0.74 | 0.89 | 0.3 | 0.51 |
| **partfull** | 0.81 | 0.99 | 0.69 | 0.93 |
| **wksub1** | 0.83 | 1.05 | 0.68 | 1.06 |
| **wksup1** | 0.83 | 1.0 | 0.69 | 0.95 |
| **conarmy** | 0.49 | 0.66 | 0.28 | 0.43 |
| **conbus** | 0.58 | 0.81 | 0.34 | 0.56 |
| **conclerg** | 0.63 | 0.83 | 0.6 | 0.83 |
| **coneduc** | 0.56 | 0.82 | 0.28 | 0.47 |
| **confed** | 0.5 | 0.7 | 0.23 | 0.38 |
| **confinan** | 0.52 | 0.72 | 0.2 | 0.31 |
| **conjudge** | 0.54 | 0.74 | 0.41 | 0.59 |
| **conlabor** | 0.54 | 0.78 | 0.38 | 0.64 |
| **conlegis** | 0.55 | 0.75 | 0.22 | 0.39 |
| **conmedic** | 0.6 | 0.85 | 0.43 | 0.67 |
| **conpress** | 0.62 | 0.84 | 0.42 | 0.64 |
| **consci** | 0.63 | 0.83 | 0.48 | 0.69 |
| **contv** | 0.58 | 0.87 | 0.33 | 0.65 |
| **vetyears** | 0.97 | 0.99 | 0.91 | 1.03 |
| **joblose** | 0.76 | 0.97 | 0.59 | 0.94 |
| **jobfind** | 0.59 | 0.84 | 0.32 | 0.67 |
| **happy** | 0.67 | 0.92 | 0.54 | 0.84 |
| **hapmar** | 0.77 | 0.94 | 0.5 | 0.81 |
| **satjob** | 0.67 | 0.96 | 0.43 | 0.83 |
| **speduc*** | 0.84 | 0.98 | 0.8 | 0.96 |
| **spdeg*** | 0.65 | 0.76 | 0.37 | 0.49 |
| **spwrksta** | 0.74 | 0.88 | 0.56 | 0.78 |
| **spjew** | 0.71 | 0.88 | 0.08 | 0.13 |
| **spfund** | 0.76 | 0.94 | 0.42 | 0.74 |
| **unemp** | 0.74 | 0.89 | 0.45 | 0.68 |
| **union1** | 0.94 | 0.99 | 0.57 | 0.83 |
| **spkath/y** | 0.74 | 0.87 | 0.23 | 0.44 |
| **colath** | 0.64 | 0.8 | 0.19 | 0.4 |
| **spkrac/y** | 0.56 | 0.71 | 0.04 | 0.07 |
| **colrac** | 0.71 | 0.9 | 0.17 | 0.34 |
| **librac/y** | 0.54 | 0.72 | 0.14 | 0.29 |
| **spkcom/y** | 0.68 | 0.8 | 0.21 | 0.32 |
| **colcom/y** | 0.67 | 0.85 | 0.2 | 0.39 |
| **libcom/y** | 0.75 | 0.89 | 0.19 | 0.34 |
| **spkhomo/y** | 0.9 | 0.98 | 0.28 | 0.65 |
| **colhomo** | 0.83 | 0.88 | 0.21 | 0.38 |
| **libhomo/y** | 0.85 | 0.96 | 0.39 | 0.67 |
| cappun | 0.67 | 0.77 | 0.37 | 0.51 |
| polhitok/y | 0.65 | 0.8 | 0.14 | 0.24 |
| polabuse/y | 0.94 | 1.0 | -0.01 | -0.02 |
| polattak/y | 0.67 | 0.86 | 0.08 | 0.16 |
| grass | 0.73 | 0.77 | 0.31 | 0.36 |
| gunlaw | 0.7 | 0.83 | 0.35 | 0.56 |
| owngun | 0.76 | 0.8 | 0.37 | 0.43 |
| hunt1 | 0.89 | 0.94 | 0.21 | 0.3 |
| class | 0.62 | 0.77 | 0.61 | 0.79 |
| satfin | 0.67 | 0.97 | 0.67 | 0.96 |
| finalter | 0.63 | 0.89 | 0.54 | 0.83 |
| finrela | 0.53 | 0.77 | 0.71 | 0.97 |
| race* | 0.93 | 0.95 | 0.92 | 0.95 |
| racdif1 | 0.81 | 0.95 | 0.61 | 0.86 |
| racdif2 | 0.93 | 1.0 | 0.08 | 0.17 |
| racdif3 | 0.71 | 0.89 | 0.42 | 0.7 |
| racdif4 | 0.75 | 0.9 | 0.38 | 0.63 |
| wlthwhts | 0.33 | 0.6 | 0.14 | 0.35 |
| wlthblks | 0.15 | 0.28 | 0.11 | 0.29 |
| wlthhsps | 0.31 | 0.62 | 0.06 | 0.15 |
| racwork | 0.62 | 0.84 | 0.37 | 0.61 |
| letin1a | 0.49 | 0.71 | 0.6 | 0.75 |
| getahead | 0.59 | 0.8 | 0.31 | 0.5 |
| aged | 0.53 | 0.75 | 0.15 | 0.31 |
| parsol | 0.36 | 0.68 | 0.46 | 0.71 |
| kidssol | 0.44 | 0.79 | 0.17 | 0.45 |
| spanking | 0.39 | 0.53 | 0.46 | 0.54 |
| divlaw | 0.48 | 0.64 | 0.34 | 0.52 |
| sexeduc | 0.78 | 0.84 | 0.34 | 0.47 |
| pillok | 0.44 | 0.67 | 0.49 | 0.64 |
| xmarsex | 0.56 | 0.8 | 0.22 | 0.37 |
| homosex | 0.68 | 0.86 | 0.61 | 0.74 |
| marhomo | 0.53 | 0.69 | 0.6 | 0.7 |
| discaff | 0.57 | 0.8 | 0.47 | 0.78 |
| abdefect | 0.83 | 0.9 | 0.41 | 0.58 |
| abnomore | 0.8 | 0.9 | 0.55 | 0.72 |
| abhlth | 0.92 | 0.96 | 0.27 | 0.4 |
| abpoor | 0.83 | 0.91 | 0.59 | 0.74 |
| abrape | 0.9 | 0.95 | 0.48 | 0.64 |
| absingle | 0.79 | 0.88 | 0.54 | 0.68 |
| abany | 0.79 | 0.88 | 0.56 | 0.71 |
| letdie1 | 0.74 | 0.82 | 0.35 | 0.48 |
| suicide1 | 0.79 | 0.89 | 0.4 | 0.56 |
| suicide2 | 0.75 | 0.85 | 0.05 | 0.07 |
| suicide3 | 0.75 | 0.85 | 0.14 | 0.21 |
| suicide4 | 0.66 | 0.77 | 0.22 | 0.31 |

| **pornlaw** | 0.7 | 0.85 | 0.48 | 0.68 |
| --- | --- | --- | --- | --- |
| **fair** | 0.43 | 0.63 | 0.18 | 0.3 |
| **helpful** | 0.42 | 0.65 | 0.24 | 0.44 |
| **trust** | 0.47 | 0.66 | 0.24 | 0.41 |
| **tax** | 0.58 | 0.73 | 0.26 | 0.46 |
| **vote16** | 0.84 | 0.92 | 0.76 | 0.88 |
| **pres16** | 0.77 | 0.83 | 0.72 | 0.79 |
| **if16who** | 0.81 | 0.89 | 0.76 | 0.86 |
| **polviews** | 0.55 | 0.66 | 0.84 | 0.88 |
| **partyid** | 0.74 | 0.9 | 0.71 | 0.89 |
| **news** | 0.31 | 0.46 | 0.37 | 0.48 |
| **relig*** | 0.76 | 0.85 | 0.63 | 0.73 |
| **jew** | 0.72 | 0.96 | 0.17 | 0.36 |
| **relig16*** | 0.62 | 0.7 | 0.55 | 0.63 |
| **jew16*** | 0.75 | 1.0 | 0.18 | 0.44 |
| **attend** | 0.56 | 0.75 | 0.75 | 0.84 |
| **pray** | 0.5 | 0.69 | 0.75 | 0.83 |
| **postlife** | 0.82 | 0.9 | 0.59 | 0.75 |
| **bible** | 0.66 | 0.75 | 0.65 | 0.75 |
| **reborn** | 0.68 | 0.83 | 0.48 | 0.64 |
| **savesoul** | 0.82 | 0.9 | 0.57 | 0.72 |
| **relpersn** | 0.63 | 0.79 | 0.79 | 0.91 |
| **sprtprsn** | 0.56 | 0.78 | 0.73 | 0.9 |
| **born** | 0.99 | 1.0 | 0.89 | 1.02 |
| **granborn** | 0.62 | 0.82 | 0.43 | 0.84 |
| **uscitzn*** | 1.0 | 1.0 | 1.0 | 1.27 |
| **fucitzn** | 0.99 | 1.0 | 0.42 | 1.02 |
| **mnthsusa** | 0.77 | 0.8 | 0.85 | 0.88 |
| **educ*** | 0.94 | 0.98 | 0.37 | 0.76 |
| **degree*** | 0.89 | 0.98 | 0.94 | 0.99 |
| **income** | 0.42 | 0.67 | 0.34 | 0.56 |
| **visitors** | 0.86 | 0.93 | 0.25 | 0.61 |
| **rvisitor** | 0.98 | 1.0 | 0.57 | 1.1 |
| **dwelown** | 0.94 | 1.0 | 0.92 | 1.02 |
| **zodiac** | 0.9 | 0.92 | 0.89 | 0.92 |
| **othlang** | 0.85 | 0.92 | 0.55 | 0.71 |
| **sex*** | 0.99 | 1.0 | 0.97 | 1.0 |
| **hispanic** | 0.98 | 1.01 | 0.92 | 1.03 |
| **health** | 0.6 | 0.75 | 0.64 | 0.81 |
| **compuse*** | 0.97 | 1.01 | 0.12 | 0.31 |
| **webmob** | 0.98 | 1.01 | 0.15 | 0.62 |
| **xmovie** | 0.66 | 0.75 | 0.23 | 0.31 |
| **usewww*** | 0.98 | 1.0 | -0.01 | -0.05 |
| **life** | 0.61 | 0.78 | 0.33 | 0.5 |
| **richwork** | 0.33 | 0.82 | 0.34 | 0.49 |

### [**Big Five:** Construct-Level Analysis]

| **Big Five** | **MAE** | **-** | **Correlation** | **Correlation-Replication ratio** |
|---|---|---|---|---|
| **extraversion** | 0.72 | - | 0.45 | 0.51 |
| **agreeableness** | 0.6 | - | 0.35 | 0.43 |
| **conscientiousness** | 0.63 | - | 0.52 | 0.59 |
| **neuroticism** | 0.75 | - | 0.68 | 0.77 |
| **openness** | 0.62 | - | 0.39 | 0.46 |

### [**Economic Games:** Construct-Level Analysis]

| **Economic Games** | **MAE** | **-** | **Correlation** | **Correlation-Replication ratio** |
|---|---|---|---|---|
| **game1_DG** | 0.23 | - | 0.11 | 0.24 |
| **game2_TF1** | 0.35 | - | 0.08 | 0.16 |
| **game3_TF2** | 0.17 | - | 0.03 | 0.06 |
| **game4_PG** | 0.52 | - | -0.05 | -0.71 |
| **game5_PD** | 0.36 | - | 0.1 | 0.32 |

**Table 3.** The construct-level analysis of predictive accuracy across the General Social Survey, Big Five personality traits, and economic games. For each construct, we provide accuracy and correlation metrics, along with replication ratios. The analysis highlights the performance of generative agents in predicting specific dimensions within these constructs, with metrics showing varied levels of predictive accuracy across different social, personality, and game-based items. This evaluation complements the individual-level analysis, which is of our primary interest in this work, by offering a detailed look at the accuracy of agents for specific constructs and items.

# [General Social Survey: Regression]

## Age

| Variable | Coefficient | Std. Error | t-value | p-value |
|---|---|---|---|---|
| const | 0.6957 | 0.0039 | 177.365 | <0.001 |
| affiliation_18 - 24 | 0.0029 | 0.0067 | 0.427 | 0.669 |
| affiliation_25 - 34 | -0.0147 | 0.0063 | -2.343 | 0.019 |
| affiliation_35 - 44 | -0.0205 | 0.0059 | -3.510 | <0.001 |
| affiliation_45 - 54 | -0.0199 | 0.0057 | -3.517 | <0.001 |
| affiliation_65 - 74 | 0.0108 | 0.0063 | 1.717 | 0.086 |
| affiliation_75 or more | 0.0224 | 0.0115 | 1.954 | 0.051 |

**Note: $R^2 = 0.045$, $F(6.0, 1045.0) = 8.18$, $p < 0.001$. Reference category: 55 - 64.**

## Census Division

| Variable | Coefficient | Std. Error | t-value | p-value |
|---|---|---|---|---|
| const | 0.6892 | 0.0043 | 161.841 | <0.001 |
| affiliation_e. sou. central | -0.0095 | 0.0069 | -1.370 | 0.171 |
| affiliation_foreign | -0.0060 | 0.0123 | -0.485 | 0.627 |
| affiliation_middle atlantic | 0.0176 | 0.0067 | 2.643 | 0.008 |
| affiliation_mountain | -0.0139 | 0.0092 | -1.515 | 0.130 |
| affiliation_new england | 0.0073 | 0.0083 | 0.878 | 0.380 |
| affiliation_pacific | -0.0075 | 0.0063 | -1.187 | 0.235 |
| affiliation_south atlantic | 0.0032 | 0.0072 | 0.438 | 0.662 |
| affiliation_w. nor. central | 0.0004 | 0.0078 | 0.049 | 0.961 |
| affiliation_w. sou. central | -0.0067 | 0.0076 | -0.878 | 0.380 |

**Note: $R^2 = 0.022$, $F(9.0, 1042.0) = 2.55$, $p < 0.001$. Reference category: e. nor. central.**

## Political Ideology

| Variable | Coefficient | Std. Error | t-value | p-value |
|---|---|---|---|---|
| const | 0.6703 | 0.0031 | 218.263 | <0.001 |
| affiliation_conservative | -0.0081 | 0.0051 | -1.605 | 0.109 |
| affiliation_extremely conservative | -0.0007 | 0.0075 | -0.099 | 0.921 |
| affiliation_extremely liberal | 0.0704 | 0.0058 | 12.181 | <0.001 |
| affiliation_liberal | 0.0497 | 0.0049 | 10.208 | <0.001 |
| affiliation_slightly conservative | -0.0030 | 0.0063 | -0.468 | 0.640 |
| affiliation_slightly liberal | 0.0269 | 0.0062 | 4.336 | <0.001 |

**Note: $R^2 = 0.215$, $F(6.0, 1045.0) = 47.70$, $p < 0.001$. Reference category: moderate.**

## Political Party

| Variable | Coefficient | Std. Error | t-value | p-value |
|---|---|---|---|---|
| const | 0.7165 | 0.0037 | 196.221 | <0.001 |
| affiliation_independent (neither) | -0.0397 | 0.0057 | -7.006 | <0.001 |
| affiliation_independent, close to democrat | 0.0022 | 0.0062 | 0.349 | 0.727 |
| affiliation_independent, close to republican | -0.0516 | 0.0069 | -7.459 | <0.001 |
| affiliation_not very strong democrat | -0.0171 | 0.0059 | -2.884 | 0.004 |

| affiliation_not very strong republican | -0.0508 | 0.0062 | -8.178 | <0.001 |
|---|---|---|---|---|
| affiliation_other party | -0.0408 | 0.0117 | -3.494 | <0.001 |
| affiliation_strong republican | -0.0577 | 0.0058 | -10.024 | <0.001 |

**Note: $R^2 = 0.155$, $F(7.0, 1044.0) = 27.28$, $p < 0.001$. Reference category: strong democrat.**

## Education

| Variable | Coefficient | Std. Error | t-value | p-value |
|---|---|---|---|---|
| const | 0.6828 | 0.0029 | 233.029 | <0.001 |
| affiliation_associate/junior college | -0.0024 | 0.0053 | -0.448 | 0.655 |
| affiliation_bachelor's | 0.0140 | 0.0046 | 3.063 | 0.002 |
| affiliation_graduate | 0.0225 | 0.0056 | 3.985 | <0.001 |
| affiliation_less than high school | -0.0354 | 0.0124 | -2.844 | 0.005 |

**Note: $R^2 = 0.034$, $F(4.0, 1047.0) = 9.11$, $p < 0.001$. Reference category: high school.**

## Race

| Variable | Coefficient | Std. Error | t-value | p-value |
|---|---|---|---|---|
| const | 0.6922 | 0.0021 | 327.316 | <0.001 |
| affiliation_black | -0.0208 | 0.0053 | -3.917 | <0.001 |
| affiliation_other | -0.0067 | 0.0063 | -1.078 | 0.281 |

**Note: $R^2 = 0.015$, $F(2.0, 1049.0) = 7.83$, $p < 0.001$. Reference category: white.**

## Ethnicity

| Variable | Coefficient | Std. Error | t-value | p-value |
|---|---|---|---|---|
| const | 0.6926 | 0.0021 | 334.816 | <0.001 |
| affiliation_American Indian or Alaskan native | -0.0055 | 0.0113 | -0.484 | 0.628 |
| affiliation_Asian | -0.0027 | 0.0085 | -0.317 | 0.752 |
| affiliation_Black/African American | -0.0183 | 0.0052 | -3.492 | <0.001 |
| affiliation_Native Hawaiian or Pacific Islander | -0.0417 | 0.0268 | -1.558 | 0.119 |
| affiliation_Other race or ethnicity | -0.0103 | 0.0084 | -1.232 | 0.218 |

**Note: $R^2 = 0.013$, $F(5.0, 1122.0) = 3.02$, $p < 0.001$. Reference category: White/Caucasian.**

## Gender

| Variable | Coefficient | Std. Error | t-value | p-value |
|---|---|---|---|---|
| const | 0.6909 | 0.0025 | 279.783 | <0.001 |
| affiliation_male | -0.0054 | 0.0037 | -1.432 | 0.153 |

**Note: $R^2 = 0.002$, $F(1.0, 1050.0) = 2.05$, $p < 0.001$. Reference category: female.**

## Income

| Variable | Coefficient | Std. Error | t-value | p-value |
|---|---|---|---|---|
| const | 0.6837 | 0.0046 | 149.055 | <0.001 |
| affiliation_$100,000 to $124,999 | -0.0023 | 0.0086 | -0.263 | 0.792 |
| affiliation_$125,000 to $149,999 | -0.0023 | 0.0103 | -0.221 | 0.825 |
| affiliation_$150,000 to $174,999 | 0.0389 | 0.0148 | 2.631 | 0.009 |
| affiliation_$175,000 to $199,999 | -0.0106 | 0.0170 | -0.623 | 0.533 |
| affiliation_$200,000 to $249,999 | 0.0329 | 0.0183 | 1.800 | 0.072 |

| affiliation_$25,000 to $34,999 | -0.0067 | 0.0076 | -0.889 | 0.374 |
|---|---|---|---|---|
| affiliation_$250,000 or more | 0.0038 | 0.0148 | 0.255 | 0.799 |
| affiliation_$35,000 to $49,999 | 0.0109 | 0.0072 | 1.515 | 0.130 |
| affiliation_$75,000 to $99,999 | 0.0039 | 0.0071 | 0.547 | 0.585 |
| affiliation_Less than $25,000 | -0.0026 | 0.0066 | -0.394 | 0.694 |

**Note: $R^2 = 0.019$, $F(10.0, 861.0) = 1.70$, $p < 0.001$. Reference category: $50,000 to $74,999.**

### Neighborhood

| Variable | Coefficient | Std. Error | t-value | p-value |
|---|---|---|---|---|
| const | 0.6908 | 0.0030 | 230.902 | <0.001 |
| affiliation_Rural | -0.0122 | 0.0054 | -2.255 | 0.024 |
| affiliation_Urban | -0.0076 | 0.0048 | -1.586 | 0.113 |

**Note: $R^2 = 0.007$, $F(2.0, 869.0) = 2.91$, $p < 0.001$. Reference category: Suburban.**

### Sexual Orientation

| Variable | Coefficient | Std. Error | t-value | p-value |
|---|---|---|---|---|
| const | 0.6783 | 0.0022 | 305.815 | <0.001 |
| affiliation_Asexual | 0.0708 | 0.0155 | 4.574 | <0.001 |
| affiliation_Bisexual | 0.0328 | 0.0074 | 4.415 | <0.001 |
| affiliation_Gay or lesbian | 0.0316 | 0.0099 | 3.199 | 0.001 |
| affiliation_Other sexual orientation | 0.0375 | 0.0189 | 1.984 | 0.048 |
| affiliation_Pansexual | 0.0596 | 0.0131 | 4.539 | <0.001 |

**Note: $R^2 = 0.071$, $F(5.0, 864.0) = 13.25$, $p < 0.001$. Reference category: Heterosexual/straight.**

## [**Big Five:** Regression]

### Age

| Variable | Coefficient | Std. Error | t-value | p-value |
|---|---|---|---|---|
| const | 0.6712 | 0.0178 | 37.639 | <0.001 |
| affiliation_18 - 24 | 0.0291 | 0.0306 | 0.949 | 0.343 |
| affiliation_25 - 34 | 0.0167 | 0.0285 | 0.588 | 0.557 |
| affiliation_35 - 44 | -0.0047 | 0.0266 | -0.178 | 0.858 |
| affiliation_45 - 54 | -0.0030 | 0.0258 | -0.118 | 0.906 |
| affiliation_65 - 74 | -0.0516 | 0.0287 | -1.798 | 0.072 |
| affiliation_75 or more | -0.1028 | 0.0521 | -1.973 | 0.049 |

**Note: $R^2 = 0.011$, $F(6.0, 1045.0) = 1.87$, $p < 0.001$. Reference category: 55 - 64.**

### Census Division

| Variable | Coefficient | Std. Error | t-value | p-value |
|---|---|---|---|---|
| const | 0.6520 | 0.0191 | 34.093 | <0.001 |
| affiliation_e. sou. central | 0.0585 | 0.0310 | 1.888 | 0.059 |
| affiliation_foreign | -0.0095 | 0.0551 | -0.172 | 0.863 |
| affiliation_middle atlantic | -0.0208 | 0.0300 | -0.694 | 0.488 |

| **affiliation_mountain** | 0.0361 | 0.0412 | 0.877 | 0.381 |
|---|---|---|---|---|
| **affiliation_new england** | -0.0181 | 0.0374 | -0.486 | 0.627 |
| **affiliation_pacific** | 0.0213 | 0.0283 | 0.753 | 0.452 |
| **affiliation_south atlantic** | 0.0313 | 0.0323 | 0.968 | 0.333 |
| **affiliation_w. nor. central** | 0.0524 | 0.0348 | 1.503 | 0.133 |
| **affiliation_w. sou. central** | -0.0209 | 0.0340 | -0.613 | 0.540 |

**Note: $R^2 = 0.011$, $F(9.0, 1042.0) = 1.28$, $p < 0.001$. Reference category: e. nor. central.**

## Political Ideology

| Variable | Coefficient | Std. Error | t-value | p-value |
|---|---|---|---|---|
| **const** | 0.6805 | 0.0154 | 44.052 | <0.001 |
| **affiliation_conservative** | -0.0068 | 0.0254 | -0.267 | 0.790 |
| **affiliation_extremely conservative** | 0.0151 | 0.0377 | 0.400 | 0.689 |
| **affiliation_extremely liberal** | -0.0241 | 0.0291 | -0.828 | 0.408 |
| **affiliation_liberal** | -0.0436 | 0.0245 | -1.779 | 0.076 |
| **affiliation_slightly conservative** | -0.0041 | 0.0317 | -0.130 | 0.897 |
| **affiliation_slightly liberal** | -0.0363 | 0.0312 | -1.162 | 0.246 |

**Note: $R^2 = 0.005$, $F(6.0, 1045.0) = 0.84$, $p < 0.001$. Reference category: moderate.**

## Political Party

| Variable | Coefficient | Std. Error | t-value | p-value |
|---|---|---|---|---|
| **const** | 0.6656 | 0.0176 | 37.744 | <0.001 |
| **affiliation_independent (neither)** | 0.0365 | 0.0274 | 1.333 | 0.183 |
| **affiliation_independent, close to democrat** | -0.0464 | 0.0298 | -1.558 | 0.120 |
| **affiliation_independent, close to republican** | 0.0105 | 0.0334 | 0.315 | 0.753 |
| **affiliation_not very strong democrat** | -0.0379 | 0.0287 | -1.321 | 0.187 |
| **affiliation_not very strong republican** | -0.0101 | 0.0300 | -0.338 | 0.736 |
| **affiliation_other party** | 0.0701 | 0.0564 | 1.242 | 0.215 |
| **affiliation_strong republican** | 0.0222 | 0.0278 | 0.799 | 0.424 |

**Note: $R^2 = 0.012$, $F(7.0, 1044.0) = 1.81$, $p < 0.001$. Reference category: strong democrat.**

## Education

| Variable | Coefficient | Std. Error | t-value | p-value |
|---|---|---|---|---|
| **const** | 0.6368 | 0.0132 | 48.170 | <0.001 |
| **affiliation_associate/junior college** | 0.0504 | 0.0237 | 2.127 | 0.034 |
| **affiliation_bachelor's** | 0.0353 | 0.0207 | 1.708 | 0.088 |
| **affiliation_graduate** | 0.0416 | 0.0255 | 1.635 | 0.102 |
| **affiliation_less than high school** | 0.1890 | 0.0562 | 3.366 | <0.001 |

**Note: $R^2 = 0.014$, $F(4.0, 1047.0) = 3.77$, $p < 0.001$. Reference category: high school.**

## Race

| Variable | Coefficient | Std. Error | t-value | p-value |
|---|---|---|---|---|
| **const** | 0.6742 | 0.0095 | 71.125 | <0.001 |
| **affiliation_black** | -0.0676 | 0.0238 | -2.834 | 0.005 |
| **affiliation_other** | 0.0093 | 0.0280 | 0.332 | 0.740 |

**Note: $R^2 = 0.008$, $F(2.0, 1049.0) = 4.27$, $p < 0.001$. Reference category: white.**

### Ethnicity

| Variable | Coefficient | Std. Error | t-value | p-value |
|---|---|---|---|---|
| **const** | 0.6740 | 0.0093 | 72.481 | <0.001 |
| **affiliation_American Indian or Alaskan native** | 0.0728 | 0.0507 | 1.436 | 0.151 |
| **affiliation_Asian** | -0.0134 | 0.0380 | -0.353 | 0.724 |
| **affiliation_Black/African American** | -0.0650 | 0.0235 | -2.761 | 0.006 |
| **affiliation_Native Hawaiian or Pacific Islander** | 0.1443 | 0.1204 | 1.199 | 0.231 |
| **affiliation_Other race or ethnicity** | 0.0464 | 0.0377 | 1.230 | 0.219 |

**Note: $R^2 = 0.012$, $F(5.0, 1122.0) = 2.80$, $p < 0.001$. Reference category: White/Caucasian.**

### Gender

| Variable | Coefficient | Std. Error | t-value | p-value |
|---|---|---|---|---|
| **const** | 0.6490 | 0.0110 | 58.917 | <0.001 |
| **affiliation_male** | 0.0377 | 0.0167 | 2.264 | 0.024 |

**Note: $R^2 = 0.005$, $F(1.0, 1050.0) = 5.12$, $p < 0.001$. Reference category: female.**

### Income

| Variable | Coefficient | Std. Error | t-value | p-value |
|---|---|---|---|---|
| **const** | 0.6991 | 0.0204 | 34.276 | <0.001 |
| **affiliation_$100,000 to $124,999** | 0.0170 | 0.0384 | 0.443 | 0.658 |
| **affiliation_$125,000 to $149,999** | -0.0581 | 0.0458 | -1.268 | 0.205 |
| **affiliation_$150,000 to $174,999** | -0.0526 | 0.0657 | -0.801 | 0.423 |
| **affiliation_$175,000 to $199,999** | -0.0166 | 0.0755 | -0.220 | 0.826 |
| **affiliation_$200,000 to $249,999** | 0.0787 | 0.0812 | 0.970 | 0.333 |
| **affiliation_$25,000 to $34,999** | -0.0181 | 0.0337 | -0.537 | 0.591 |
| **affiliation_$250,000 or more** | -0.0505 | 0.0657 | -0.769 | 0.442 |
| **affiliation_$35,000 to $49,999** | -0.0496 | 0.0321 | -1.546 | 0.123 |
| **affiliation_$75,000 to $99,999** | -0.0571 | 0.0315 | -1.812 | 0.070 |
| **affiliation_Less than $25,000** | -0.0210 | 0.0295 | -0.714 | 0.476 |

**Note: $R^2 = 0.010$, $F(10.0, 861.0) = 0.88$, $p < 0.001$. Reference category: $50,000 to $74,999.**

### Neighborhood

| Variable | Coefficient | Std. Error | t-value | p-value |
|---|---|---|---|---|
| **const** | 0.6619 | 0.0132 | 49.984 | <0.001 |
| **affiliation_Rural** | 0.0273 | 0.0240 | 1.137 | 0.256 |
| **affiliation_Urban** | 0.0163 | 0.0212 | 0.770 | 0.442 |

**Note: $R^2 = 0.002$, $F(2.0, 869.0) = 0.73$, $p < 0.001$. Reference category: Suburban.**

### Sexual Orientation

| Variable | Coefficient | Std. Error | t-value | p-value |
|---|---|---|---|---|
| **const** | 0.6798 | 0.0102 | 66.621 | <0.001 |
| **affiliation_Asexual** | -0.0499 | 0.0712 | -0.700 | 0.484 |

| affiliation_Bisexual | -0.0359 | 0.0342 | -1.050 | 0.294 |
|---|---|---|---|---|
| affiliation_Gay or lesbian | 0.0150 | 0.0455 | 0.329 | 0.742 |
| affiliation_Other sexual orientation | -0.0853 | 0.0869 | -0.981 | 0.327 |
| affiliation_Pansexual | -0.0603 | 0.0605 | -0.997 | 0.319 |

**Note: $R^2$ = 0.004, F(5.0, 864.0) = 0.69, p < 0.001. Reference category: Heterosexual/straight.**

## [**Economic Games:** Regression]

### Age

| Variable | Coefficient | Std. Error | t-value | p-value |
|---|---|---|---|---|
| const | 0.2859 | 0.0593 | 4.822 | <0.001 |
| affiliation_18 - 24 | -0.0007 | 0.1017 | -0.007 | 0.994 |
| affiliation_25 - 34 | -0.0081 | 0.0945 | -0.085 | 0.932 |
| affiliation_35 - 44 | 0.1149 | 0.0884 | 1.300 | 0.194 |
| affiliation_45 - 54 | 0.0898 | 0.0857 | 1.049 | 0.295 |
| affiliation_65 - 74 | 0.0104 | 0.0955 | 0.109 | 0.913 |
| affiliation_75 or more | 0.0504 | 0.1755 | 0.287 | 0.774 |

**Note: $R^2$ = 0.003, F(6.0, 1041.0) = 0.54, p < 0.001. Reference category: 55 - 64.**

### Census Division

| Variable | Coefficient | Std. Error | t-value | p-value |
|---|---|---|---|---|
| const | 0.2977 | 0.0636 | 4.683 | <0.001 |
| affiliation_e. sou. central | 0.1580 | 0.1032 | 1.531 | 0.126 |
| affiliation_foreign | -0.0672 | 0.1827 | -0.368 | 0.713 |
| affiliation_middle atlantic | -0.0062 | 0.0998 | -0.062 | 0.950 |
| affiliation_mountain | 0.0058 | 0.1368 | 0.042 | 0.966 |
| affiliation_new england | 0.0033 | 0.1239 | 0.026 | 0.979 |
| affiliation_pacific | 0.0994 | 0.0939 | 1.059 | 0.290 |
| affiliation_south atlantic | -0.0280 | 0.1073 | -0.261 | 0.794 |
| affiliation_w. nor. central | -0.0012 | 0.1156 | -0.010 | 0.992 |
| affiliation_w. sou. central | -0.0131 | 0.1133 | -0.116 | 0.908 |

**Note: $R^2$ = 0.005, F(9.0, 1038.0) = 0.57, p < 0.001. Reference category: e. nor. central.**

### Political Ideology

| Variable | Coefficient | Std. Error | t-value | p-value |
|---|---|---|---|---|
| const | 0.3642 | 0.0509 | 7.153 | <0.001 |
| affiliation_conservative | -0.0534 | 0.0840 | -0.635 | 0.526 |
| affiliation_extremely conservative | 0.2830 | 0.1252 | 2.259 | 0.024 |
| affiliation_extremely liberal | -0.0917 | 0.0959 | -0.956 | 0.339 |
| affiliation_liberal | -0.0976 | 0.0810 | -1.205 | 0.228 |
| affiliation_slightly conservative | -0.0875 | 0.1046 | -0.836 | 0.403 |
| affiliation_slightly liberal | -0.0995 | 0.1030 | -0.966 | 0.334 |

**Note: $R^2$ = 0.010, F(6.0, 1041.0) = 1.76, p < 0.001. Reference category: moderate.**

### Political Party

| Variable | Coefficient | Std. Error | t-value | p-value |
|---|---|---|---|---|
| const | 0.2616 | 0.0587 | 4.456 | <0.001 |
| affiliation_independent (neither) | 0.1738 | 0.0909 | 1.913 | 0.056 |
| affiliation_independent, close to democrat | 0.0344 | 0.0988 | 0.348 | 0.728 |
| affiliation_independent, close to republican | 0.0215 | 0.1110 | 0.193 | 0.847 |
| affiliation_not very strong democrat | 0.0117 | 0.0953 | 0.123 | 0.903 |
| affiliation_not very strong republican | 0.0204 | 0.0996 | 0.205 | 0.838 |
| affiliation_other party | 0.0712 | 0.1871 | 0.381 | 0.703 |
| affiliation_strong republican | 0.1708 | 0.0926 | 1.846 | 0.065 |

**Note: $R^2 = 0.007$, $F(7.0, 1040.0) = 1.01$, $p < 0.001$. Reference category: strong democrat.**

### Education

| Variable | Coefficient | Std. Error | t-value | p-value |
|---|---|---|---|---|
| const | 0.3909 | 0.0440 | 8.887 | <0.001 |
| affiliation_associate/junior college | -0.0941 | 0.0787 | -1.195 | 0.232 |
| affiliation_bachelor's | -0.1202 | 0.0688 | -1.748 | 0.081 |
| affiliation_graduate | -0.1055 | 0.0850 | -1.241 | 0.215 |
| affiliation_less than high school | -0.0636 | 0.1866 | -0.341 | 0.733 |

**Note: $R^2 = 0.004$, $F(4.0, 1043.0) = 0.95$, $p < 0.001$. Reference category: high school.**

### Race

| Variable | Coefficient | Std. Error | t-value | p-value |
|---|---|---|---|---|
| const | 0.3365 | 0.0315 | 10.686 | <0.001 |
| affiliation_black | -0.0358 | 0.0793 | -0.452 | 0.652 |
| affiliation_other | -0.0610 | 0.0935 | -0.653 | 0.514 |

**Note: $R^2 = 0.001$, $F(2.0, 1045.0) = 0.28$, $p < 0.001$. Reference category: white.**

### Ethnicity

| Variable | Coefficient | Std. Error | t-value | p-value |
|---|---|---|---|---|
| const | 0.3335 | 0.0298 | 11.173 | <0.001 |
| affiliation_American Indian or Alaskan native | -0.0601 | 0.1625 | -0.370 | 0.712 |
| affiliation_Asian | -0.0909 | 0.1219 | -0.746 | 0.456 |
| affiliation_Black/African American | -0.0342 | 0.0757 | -0.452 | 0.651 |
| affiliation_Native Hawaiian or Pacific Islander | -0.0499 | 0.3859 | -0.129 | 0.897 |
| affiliation_Other race or ethnicity | -0.0491 | 0.1219 | -0.403 | 0.687 |

**Note: $R^2 = 0.001$, $F(5.0, 1118.0) = 0.18$, $p < 0.001$. Reference category: White/Caucasian.**

### Gender

| Variable | Coefficient | Std. Error | t-value | p-value |
|---|---|---|---|---|
| const | 0.3190 | 0.0365 | 8.742 | <0.001 |
| affiliation_male | 0.0148 | 0.0554 | 0.266 | 0.790 |

**Note: $R^2 = 0.000$, $F(1.0, 1046.0) = 0.07$, $p < 0.001$. Reference category: female.**

### Income

| Variable | Coefficient | Std. Error | t-value | p-value |
|---|---|---|---|---|
| const | 0.2884 | 0.0731 | 3.947 | <0.001 |
| affiliation_$100,000 to $124,999 | -0.0201 | 0.1373 | -0.147 | 0.883 |
| affiliation_$125,000 to $149,999 | -0.0204 | 0.1638 | -0.125 | 0.901 |
| affiliation_$150,000 to $174,999 | -0.0917 | 0.2347 | -0.391 | 0.696 |
| affiliation_$175,000 to $199,999 | -0.0945 | 0.2699 | -0.350 | 0.726 |
| affiliation_$200,000 to $249,999 | 0.0108 | 0.2900 | 0.037 | 0.970 |
| affiliation_$25,000 to $34,999 | 0.0166 | 0.1205 | 0.138 | 0.890 |
| affiliation_$250,000 or more | 0.0266 | 0.2347 | 0.113 | 0.910 |
| affiliation_$35,000 to $49,999 | 0.0252 | 0.1147 | 0.220 | 0.826 |
| affiliation_$75,000 to $99,999 | -0.0138 | 0.1128 | -0.122 | 0.903 |
| affiliation_Less than $25,000 | 0.2508 | 0.1054 | 2.381 | 0.018 |

**Note: $R^2 = 0.011$, $F(10.0, 860.0) = 0.95$, $p < 0.001$. Reference category: $50,000 to $74,999.**

### Neighborhood

| Variable | Coefficient | Std. Error | t-value | p-value |
|---|---|---|---|---|
| const | 0.3270 | 0.0475 | 6.884 | <0.001 |
| affiliation_Rural | 0.0676 | 0.0862 | 0.785 | 0.433 |
| affiliation_Urban | -0.0300 | 0.0760 | -0.394 | 0.693 |

**Note: $R^2 = 0.001$, $F(2.0, 868.0) = 0.56$, $p < 0.001$. Reference category: Suburban.**

### Sexual Orientation

| Variable | Coefficient | Std. Error | t-value | p-value |
|---|---|---|---|---|
| const | 0.3417 | 0.0365 | 9.366 | <0.001 |
| affiliation_Asexual | -0.0639 | 0.2545 | -0.251 | 0.802 |
| affiliation_Bisexual | -0.0330 | 0.1222 | -0.270 | 0.787 |
| affiliation_Gay or lesbian | -0.0567 | 0.1624 | -0.349 | 0.727 |
| affiliation_Other sexual orientation | -0.0750 | 0.3107 | -0.241 | 0.809 |
| affiliation_Pansexual | -0.0866 | 0.2160 | -0.401 | 0.689 |

**Note: $R^2 = 0.000$, $F(5.0, 863.0) = 0.08$, $p < 0.001$. Reference category: Heterosexual/straight.**

**Table 4.** Regression analyses of predictive performance across demographic variables. This table presents the results of regression analyses examining the influence of demographic variables on the predictive performance of generative agents, across the General Social Survey, Big Five Personality Traits, and economic games. Separate models were run for each demographic factor, with predictive performance as the dependent variable. Coefficients represent the relative difference in performance for each demographic group, with positive values indicating higher predictive accuracy. Significant discrepancies are observed for political ideology, party affiliation, and sexual orientation, with stronger performance for participants identifying as strong liberals, strong Democrats, and non-heterosexual/straight individuals.

## [**General Social Survey:** Demographic Parity Difference]

| | **Agents w/ Interview** | **Agents w/ Demog. Info.** | **Agents w/ Persona Desc.** | **Agents w/ Survey Info** | **Agents w/ Survey + Interview** |
|---|---|---|---|---|---|
| **Age** | DPD=4.48<br>Min: 35 - 44 (64.23%)<br>Max: 75 or more (68.71%) | DPD=6.44<br>Min: 45 - 54 (54.42%)<br>Max: 75 or more (60.87%) | DPD=3.43<br>Min: 45 - 54 (55.30%)<br>Max: 18 - 24 (58.73%) | DPD=3.12<br>Min: 45 - 54 (63.79%)<br>Max: 75 or more (66.91%) | DPD=5.44<br>Min: 25 - 34 (66.82%)<br>Max: 75 or more (72.27%) |
| **Census Division** | DPD=2.82<br>Min: mountain (64.58%)<br>Max: middle atlantic (67.40%) | DPD=2.19<br>Min: e. sou. central (55.50%)<br>Max: middle atlantic (57.68%) | DPD=3.81<br>Min: mountain (54.28%)<br>Max: middle atlantic (58.10%) | DPD=3.83<br>Min: mountain (63.05%)<br>Max: middle atlantic (66.88%) | DPD=4.14<br>Min: mountain (66.09%)<br>Max: middle atlantic (70.22%) |
| **Political Ideology** | DPD=8.60<br>Min: conservative (62.69%)<br>Max: extremely liberal (71.29%) | DPD=13.75<br>Min: conservative (49.19%)<br>Max: extremely liberal (62.94%) | DPD=13.00<br>Min: extremely conservative (49.17%)<br>Max: extremely liberal (62.17%) | DPD=6.22<br>Min: extremely conservative (62.68%)<br>Max: extremely liberal (68.90%) | DPD=7.09<br>Min: extremely conservative (66.09%)<br>Max: extremely liberal (73.18%) |
| **Political Party** | DPD=6.98<br>Min: strong republican (62.12%)<br>Max: independent, close to democrat (69.10%) | DPD=11.26<br>Min: strong republican (50.43%)<br>Max: independent, close to democrat (61.69%) | DPD=10.27<br>Min: strong republican (50.84%)<br>Max: independent, close to democrat (61.11%) | DPD=5.16<br>Min: other party (62.37%)<br>Max: strong democrat (67.53%) | DPD=5.38<br>Min: strong republican (65.53%)<br>Max: strong democrat (70.91%) |
| **Education** | DPD=5.32<br>Min: less than high school (62.00%)<br>Max: graduate (67.32%) | DPD=5.09<br>Min: less than high school (53.14%)<br>Max: graduate (58.23%) | DPD=6.30<br>Min: less than high school (52.31%)<br>Max: graduate (58.61%) | DPD=4.63<br>Min: less than high school (61.69%)<br>Max: graduate (66.32%) | DPD=4.25<br>Min: less than high school (65.50%)<br>Max: graduate (69.75%) |
| **Race** | DPD=2.22<br>Min: black (63.81%)<br>Max: white (66.03%) | DPD=3.39<br>Min: black (53.40%)<br>Max: white (56.79%) | DPD=3.04<br>Min: black (53.63%)<br>Max: white (56.67%) | DPD=1.95<br>Min: black (63.40%)<br>Max: white (65.35%) | DPD=2.49<br>Min: black (66.27%)<br>Max: white (68.77% |
| **Ethnicity** | DPD=2.26<br>Min: Black/African American (63.83%)<br>Max: White/Caucasian (66.09%) | DPD=3.81<br>Min: Other race or ethnicity (53.19%)<br>Max: Native Hawaiian or Pacific Islander (57.00%) | DPD=3.63<br>Min: Other race or ethnicity (53.13%)<br>Max: White/Caucasian (56.76%) | DPD=2.59<br>Min: Other race or ethnicity (62.78%)<br>Max: White/Caucasian (65.37%) | DPD=2.84<br>Min: Other race or ethnicity (65.94%)<br>Max: White/Caucasian (68.78%) |
| **Gender** | DPD=0.71<br>Min: male (65.27%)<br>Max: female (65.98%) | DPD=0.86<br>Min: male (55.79%)<br>Max: female (56.66%) | DPD=1.01<br>Min: male (55.64%)<br>Max: female (56.65%) | DPD=0.15<br>Min: female (64.91%)<br>Max: male (65.06%) | DPD=0.63<br>Min: male (67.94%)<br>Max: female (68.57%) |
| **Income** | DPD=4.47<br>Min: $175,000 to $199,999 (64.48%)<br>Max: $150,000 to $174,999 (68.95%) | DPD=6.43<br>Min: $250,000 or more (54.07%)<br>Max: $200,000 to $249,999 (60.50%) | DPD=7.12<br>Min: Less than $25,000 (54.27%)<br>Max: $200,000 to $249,999 (61.39%) | DPD=6.40<br>Min: $175,000 to $199,999 (61.43%)<br>Max: $200,000 to $249,999 (67.83%) | DPD=6.40<br>Min: $175,000 to $199,999 (64.43%)<br>Max: $200,000 to $249,999 (70.83%) |
| **Neighborhood** | DPD=1.14<br>Min: Rural (64.74%)<br>Max: Suburban (65.87%) | DPD=1.81<br>Min: Rural (55.06%)<br>Max: Suburban (56.88%) | DPD=3.00<br>Min: Rural (53.92%)<br>Max: Suburban (56.92%) | DPD=1.11<br>Min: Rural (64.02%)<br>Max: Suburban (65.12%) | DPD=1.40<br>Min: Rural (67.23%)<br>Max: Suburban (68.63%) |
| **Sexual Orientation** | DPD=7.44<br>Min: Heterosexual/straight (64.56%)<br>Max: Asexual (72.00%) | DPD=8.27<br>Min: Heterosexual/straight (54.97%)<br>Max: Pansexual (63.24%) | DPD=9.68<br>Min: Heterosexual/straight (55.03%)<br>Max: Asexual (64.71%) | DPD=5.39<br>Min: Heterosexual/straight (63.94%)<br>Max: Pansexual (69.33%) | DPD=6.50<br>Min: Heterosexual/straight (67.06%)<br>Max: Asexual (73.56%) |

## [**Big Five:** Demographic Parity Difference]

| | **Agents w/ Interview** | **Agents w/ Demog. Info.** | **Agents w/ Persona Desc.** | **Agents w/ Survey Info** | **Agents w/ Survey + Interview** |
|---|---|---|---|---|---|
| **Age** | DPD=0.22<br>Min: 18 - 24 (0.56)<br>Max: 75 or more (0.78) | DPD=0.37<br>Min: 18 - 24 (0.23)<br>Max: 65 - 74 (0.6) | DPD=0.33<br>Min: 18 - 24 (0.37)<br>Max: 65 - 74 (0.7) | DPD=0.3<br>Min: 18- 24 (0.31)<br>Max: 75 or more (0.61 | DPD=0.24<br>Min: 18 - 24 (0.52)<br>Max: 75 or more (0.76) |
| **Census Division** | DPD=0.19<br>Min: mountain (0.49)<br>Max: w. sou. central (0.68) | DPD=0.24<br>Min: mountain (0.36)<br>Max: foreign (0.6) | DPD=0.23<br>Min: mountain (0.39)<br>Max: foreign (0.63) | DPD=0.15<br>Min: mountain (0.45)<br>Max: new england (0.6) | DPD=0.2<br>Min: mountain (0.5)<br>Max: new england (0.7) |
| **Political Ideology** | DPD=0.06<br>Min: moderate (0.6)<br>Max: extremely conservative (0.67) | DPD=0.17<br>Min: extremely liberal (0.34)<br>Max: moderate (0.51) | DPD=0.14<br>Min: extremely liberal (0.49)<br>Max: slightly conservative (0.63) | DPD=0.18<br>Min: extremely liberal (0.4)<br>Max: conservative (0.58) | DPD=0.05<br>Min: extremely liberal (0.6)<br>Max: conservative (0.64) |
| **Political Party** | DPD=0.17<br>Min: other party (0.49)<br>Max: independent, close to democrat (0.66) | DPD=0.22<br>Min: other party (0.3)<br>Max: independent, close to democrat (0.53) | DPD=0.22<br>Min: other party (0.39)<br>Max: strong republican (0.61) | DPD=0.3<br>Min: other party (0.3)<br>Max: not very strong republican (0.6) | DPD=0.23<br>Min: other party (0.42)<br>Max: strong republican (0.65) |
| **Education** | DPD=0.23<br>Min: less than high school (0.44)<br>Max: graduate (0.68) | DPD=0.24<br>Min: less than high school (0.3)<br>Max: graduate (0.54) | DPD=0.3<br>Min: less than high school (0.33)<br>Max: graduate (0.63) | DPD=0.08<br>Min: less than high school (0.47)<br>Max: graduate (0.55) | DPD=0.17<br>Min: less than high school (0.49)<br>Max: graduate (0.65) |
| **Race** | DPD=0.11<br>Min: other (0.59)<br>Max: black (0.7) | DPD=0.17<br>Min: other (0.4)<br>Max: black (0.57) | DPD=0.08<br>Min: other (0.52)<br>Max: black (0.6) | DPD=0.09<br>Min: white (0.5)<br>Max: black (0.58) | DPD=0.08<br>Min: white (0.61)<br>Max: black (0.69) |
| **Ethnicity** | DPD=0.21<br>Min: Native Hawaiian or Pacific Islander (0.49)<br>Max: Black/African American (0.71) | DPD=0.62<br>Min: Native Hawaiian or Pacific Islander (-0.05)<br>Max: Black/African American (0.57) | DPD=0.42<br>Min: Native Hawaiian or Pacific Islander (0.31)<br>Max: American Indian or Alaskan native (0.73) | DPD=0.21<br>Min: Native Hawaiian or Pacific Islander (0.38)<br>Max: Black/African American (0.59) | DPD=0.26<br>Min: Native Hawaiian or Pacific Islander (0.43)<br>Max: Black/African American (0.69) |
| **Gender** | DPD=0.01<br>Min: male (0.62)<br>Max: female (0.63) | DPD=0.01<br>Min: female (0.46)<br>Max: male (0.46) | DPD=0.02<br>Min: female (0.54)<br>Max: male (0.55) | DPD=0.02<br>Min: female (0.5)<br>Max: male (0.53) | DPD=0.0<br>Min: male (0.62)<br>Max: female (0.62) |
| **Income** | DPD=0.32<br>Min: $200,000 to $249,999 (0.41)<br>Max: $250,000 or more (0.73) | DPD=0.29<br>Min: $200,000 to $249,999 (0.31)<br>Max: $250,000 or more (0.6) | DPD=0.38<br>Min: $200,000 to $249,999 (0.35)<br>Max: $150,000 to $174,999 (0.73) | DPD=0.33<br>Min: $200,000 to $249,999 (0.31)<br>Max: $125,000 to $149,999 (0.64) | DPD=0.36<br>Min: $200,000 to $249,999 (0.39)<br>Max: $250,000 or more (0.75) |
| **Neighborhood** | DPD=0.03<br>Min: Urban (0.58)<br>Max: Rural (0.62) | DPD=0.05<br>Min: Rural (0.43)<br>Max: Urban (0.48) | DPD=0.04<br>Min: Suburban (0.51)<br>Max: Urban (0.54) | DPD=0.04<br>Min: Suburban (0.48)<br>Max: Urban (0.52) | DPD=0.02<br>Min: Urban (0.58)<br>Max: Rural (0.61) |
| **Sexual Orientation** | DPD=0.23<br>Min: Other sexual orientation (0.43)<br>Max: Pansexual (0.66) | DPD=0.29<br>Min: Other sexual orientation (0.17)<br>Max: Heterosexual/straight (0.46) | DPD=0.45<br>Min: Other sexual orientation (0.1)<br>Max: Heterosexual/straight (0.55) | DPD=0.27<br>Min: Asexual (0.24)<br>Max: Heterosexual/straight (0.52) | DPD=0.29<br>Min: Other sexual orientation (0.39)<br>Max: Pansexual (0.68) |

## [**Economic Games:** Demographic Parity Difference]

| | Agents w/ Interview | Agents w/ Demog. Info. | Agents w/ Persona Desc. | Agents w/ Survey Info | Agents w/ Survey + Interview |
|---|---|---|---|---|---|
| **Age** | DPD=0.18<br>Min: 75 or more (0.21)<br>Max: 45 - 54 (0.39) | DPD=0.19<br>Min: 35 - 44 (0.18)<br>Max: 55 - 64 (0.36) | DPD=0.17<br>Min: 18 -24 (0.21)<br>Max: 45 - 54 (0.37) | DPD=0.09<br>Min: 75 or more (0.23)<br>Max: 45 - 54 (0.32) | DPD=0.13<br>Min: 75 or more (0.2)<br>Max: 25 - 34 (0.34) |
| **Census Division** | DPD=0.2<br>Min: mountain (0.25)<br>Max: foreign (0.45) | DPD=0.21<br>Min: e. nor. central (0.19)<br>Max: new england (0.4) | DPD=0.21<br>Min: mountain (0.21)<br>Max: pacific (0.42) | DPD=0.23<br>Min: mountain (0.12)<br>Max: new england (0.35) | DPD=0.22<br>Min: e. nor. central (0.16)<br>Max: pacific (0.38) |
| **Political Ideology** | DPD=0.19<br>Min: conservative (0.24)<br>Max: extremely liberal (0.44) | DPD=0.5<br>Min: extremely conservative (-0.03)<br>Max: extremely liberal (0.47) | DPD=0.17<br>Min: moderate (0.23)<br>Max: slightly liberal (0.41) | DPD=0.22<br>Min: slightly conservative (0.2)<br>Max: extremely liberal (0.42) | DPD=0.21<br>Min: conservative (0.18)<br>Max: slightly liberal (0.39) |
| **Political Party** | DPD=0.22<br>Min: independent (neither) (0.21)<br>Max: strong democrat (0.43) | DPD=0.4<br>Min: strong republican (0.03)<br>Max: strong democrat (0.43) | DPD=0.23<br>Min: other party (0.12)<br>Max: strong democrat (0.36) | DPD=0.23<br>Min: independent (neither) (0.17)<br>Max: strong democrat (0.4) | DPD=0.29<br>Min: independent (neither) (0.11)<br>Max: strong democrat (0.41) |
| **Education** | DPD=0.14<br>Min: high school (0.29)<br>Max: less than high school (0.43) | DPD=0.12<br>Min: associate/junior college (0.21)<br>Max: graduate (0.33) | DPD=0.08<br>Min: high school (0.28)<br>Max: graduate (0.36) | DPD=0.1<br>Min: high school (0.25)<br>Max: less than high school (0.35) | DPD=0.12<br>Min: high school (0.23)<br>Max: bachelor's (0.35) |
| **Race** | DPD=0.04<br>Min: black (0.3)<br>Max: white (0.34) | DPD=0.04<br>Min: other (0.24)<br>Max: white (0.29) | DPD=0.03<br>Min: black (0.28)<br>Max: white (0.31) | DPD=0.1<br>Min: black (0.2)<br>Max: other (0.3) | DPD=0.05<br>Min: other (0.23)<br>Max: white (0.28) |
| **Ethnicity** | DPD=0.52<br>Min: Other race or ethnicity (0.06)<br>Max: Native Hawaiian or Pacific Islander (0.58) | DPD=0.63<br>Min: Other race or ethnicity (0.07)<br>Max: Native Hawaiian or Pacific Islander (0.7) | DPD=0.5<br>Min: Other race or ethnicity (0.11)<br>Max: American Indian or Alaskan native (0.61) | DPD=0.45<br>Min: Other race or ethnicity (0.07)<br>Max: American Indian or Alaskan native (0.52) | DPD=0.36<br>Min: Other race or ethnicity (0.04)<br>Max: Native Hawaiian or Pacific Islander (0.4) |
| **Gender** | DPD=0.03<br>Min: female (0.32)<br>Max: male (0.35) | DPD=0.03<br>Min: female (0.27)<br>Max: male (0.29) | DPD=0.03<br>Min: female (0.3)<br>Max: male (0.33) | DPD=0.03<br>Min: female (0.27)<br>Max: male (0.29) | DPD=0.04<br>Min: female (0.26)<br>Max: male (0.3) |
| **Income** | DPD=0.51<br>Min: $200,000 to $249,999 (0.21)<br>Max: $175,000 to $199,999 (0.72) | DPD=0.46<br>Min: $200,000 to $249,999 (0.19)<br>Max: $150,000 to $174,999 (0.65) | DPD=0.5<br>Min: $200,000 to $249,999 (0.21)<br>Max: $175,000 to $199,999 (0.71) | DPD=0.48<br>Min: $200,000 to $249,999 (0.15)<br>Max: $175,000 to $199,999 (0.63) | DPD=0.63<br>Min: $200,000 to $249,999 (0.09)<br>Max: $150,000 to $174,999 (0.71) |
| **Neighborhood** | DPD=0.14<br>Min: Urban (0.25)<br>Max: Suburban (0.38) | DPD=0.11<br>Min: Urban (0.2)<br>Max: Suburban (0.32) | DPD=0.13<br>Min: Urban (0.23)<br>Max: Suburban (0.35) | DPD=0.12<br>Min: Urban (0.21)<br>Max: Suburban (0.32) | DPD=0.09<br>Min: Urban (0.23)<br>Max: Suburban (0.31) |
| **Sexual Orientation** | DPD=0.32<br>Min: Asexual (0.23)<br>Max: Other sexual orientation (0.55) | DPD=0.41<br>Min: Heterosexual/straight (0.24)<br>Max: Other sexual orientation (0.66) | DPD=0.31<br>Min: Bisexual (0.23)<br>Max: Other sexual orientation (0.54) | DPD=0.48<br>Min: Asexual (0.18)<br>Max: Other sexual orientation (0.66) | DPD=0.36<br>Min: Other sexual orientation (0.19)<br>Max: Gay or lesbian (0.55) |

**Table 5.** Demographic Parity Difference (DPD) Results. This table summarizes the results of regression analyses measuring demographic parity differences (DPD) across three tasks (GSS, Big Five, and economic games) for agents using demographic information, interview data, and persona-based profiles. Interview-based agents consistently reduced bias compared to demographic-based agents across political ideology, race, and gender.

## [Robustness Analysis]

| | GSS | Big Five | Economic Games | GSS Num. |
|---|---|---|---|---|
| ***Interview*** | Acc.=64.52 (std =7.85)<br>Nrm. Acc.=0.82 (std = 0.13)<br>--<br>Correl.=0.72 (z std = 0.24)<br>Nrm. Correl.=0.84 (std = 0.13) | MAE=0.67 (std=0.27)<br><br>--<br>Correl.=0.79 (z std=0.73)<br>Nrm. Correl.=0.77 (std=0.76) | MAE=0.51 (std=2.01)<br><br>--<br>Correl.=0.52 (z std=0.9)<br>Nrm. Correl.=0.33 (std=1.22) | MAE=0.14 (std=0.15)<br><br>--<br>Correl.=0.97 (z std=0.82)<br>Nrm. Correl.=2.16 (std=9.17) |
| **Survey Agents**<br><br>(GSS+Big5)<br>Remove only GSS outcome question out of input | Acc.=64.25 (std = 12.11)<br>Nrm. Acc.=0.82 (std = 0.18)<br>--<br>Correl.=0.70 (z std = 0.24)<br>Nrm. Correl.=0.82 (std = 0.14) | MAE=0.47 (std=2.0)<br><br>--<br>Correl.=0.52 (z std=0.93)<br>Nrm. Correl.=0.31 (std=1.22) | MAE=0.47 (std=2.0)<br><br>--<br>Correl.=0.52 (z std=0.93)<br>Nrm. Correl.=0.31 (std=1.22) | MAE=0.25 (std=0.17)<br><br>--<br>Correl.=0.89 (z std=0.62)<br>Nrm. Correl.=2.05 (std=9.07) |
| **Survey Agents**<br><br>(GSS+Big5)<br>Remove whole GSS question block from input when predicting on question in that block | Acc.=60.57 (std = 7.82)<br>Nrm. Acc.=0.77 (std = 0.12)<br>--<br>Correl.=0.70 (z std = 0.22)<br>Nrm. Correl.=0.82 (std = 0.14) | MAE=0.68 (std=0.25)<br><br>--<br>Correl.=0.72 (z std=0.75)<br>Nrm. Correl.=0.64 (std=0.61) | MAE=0.47 (std=2.0)<br><br>--<br>Correl.=0.52 (z std=0.93)<br>Nrm. Correl.=0.31 (std=1.22) | MAE=0.32 (std=0.19)<br><br>--<br>Correl.=0.75 (z std=0.62)<br>Nrm. Correl.=1.74 (std=8.94) |
| **Survey agents with interview**<br><br>(GSS+big5+interview)<br>Remove only GSS outcome question out of input | Acc.=66.81 (std = 12.91)<br>Nrm. Acc.=0.85 (std = 0.19)<br>--<br>Correl.=0.74 (z std = 0.21)<br>Nrm. Correl.=0.87 (std = 0.14) | MAE=0.63 (std=0.26)<br><br>--<br>Correl.=0.79 (z std=0.75)<br>Nrm. Correl.=0.76 (std=0.74) | MAE=0.48 (std=2.01)<br><br>--<br>Correl.=0.57 (z std=0.87)<br>Nrm. Correl.=0.43 (std=1.34) | MAE=0.24 (std=0.20)<br><br>--<br>Correl.=0.87 (z std=0.63)<br>Nrm. Correl.=2.03 (std=8.88) |
| **Maximal agents**<br><br>(GSS+big5+econ+interview)<br>Remove only GSS outcome question out of input | Acc.=68.42 (std = 7.02)<br>Nrm. Acc.=0.87 (std = 0.13)<br>--<br>Correl.=0.75 (z std = 0.24)<br>Nrm. Correl.=0.87 (std = 0.14) | MAE=0.68 (std=0.26)<br><br>--<br>Correl.=0.77 (z std=0.75)<br>Nrm. Correl.=0.72 (std=0.78) | MAE=0.47 (std=2.0)<br><br>--<br>Correl.=0.55 (z std=0.93)<br>Nrm. Correl.=0.30 (std=1.29) | MAE=0.28 (std=0.21)<br><br>--<br>Correl.=0.78 (z std=0.71)<br>Nrm. Correl.=1.85 (std=9.19) |
| **Summary** | Acc.=63.61 (std=7.09)<br>Nrm. Acc.=0.81 (std=0.13)<br>--<br>Correl.=0.72 (z std=0.23)<br>Nrm. Correl.=0.84 (std=0.15) | MAE=0.69 (std=0.3)<br><br>--<br>Correl.=0.76 (z std=0.76)<br>Nrm. Correl.=0.7 (std=0.67) | MAE=0.5 (std=2.0)<br><br>--<br>Correl.=0.49 (z std=0.85)<br>Nrm. Correl.=0.41 (std=1.1) | MAE=0.19 (std=0.18)<br><br>--<br>Correl.=0.93 (z std=0.82)<br>Nrm. Correl.=0.92 (std=2.77) |
| **Random lesion (0% removal)** | Acc.=64.52 (std =7.85)<br>Nrm. Acc.=0.82 (std = 0.13)<br>--<br>Correl.=0.72 (z std = 0.24)<br>Nrm. Correl.=0.84 (std = 0.13) | MAE=0.67 (std=0.27)<br><br>--<br>Correl.=0.79 (z std=0.73)<br>Nrm. Correl.=0.77 (std=0.76) | MAE=0.51 (std=2.01)<br><br>--<br>Correl.=0.52 (z std=0.9)<br>Nrm. Correl.=0.33 (std=1.22) | MAE=0.14 (std=0.15)<br><br>--<br>Correl.=0.97 (z std=0.82)<br>Nrm. Correl.=2.16 (std=9.17) |
| **Random lesion (20% removal)** | Acc.=64.20 (std = 7.58)<br>Nrm. Acc.=0.82 (std = 0.14)<br>--<br>Correl.=0.71 (z std = 0.22)<br>Nrm. Correl.=0.84 (std = 0.14) | MAE=0.66 (std=0.28)<br><br>--<br>Correl.=0.79 (z std=0.74)<br>Nrm. Correl.=0.75 (std=0.72) | MAE=0.52 (std=2.02)<br><br>--<br>Correl.=0.58 (z std=0.89)<br>Nrm. Correl.=0.46 (std=1.18) | MAE=0.16 (std=0.13)<br><br>--<br>Correl.=0.95 (z std=0.81)<br>Nrm. Correl.=1.46 (std=3.71) |
| **Random lesion (40% removal)** | Acc.=63.77 (std = 7.30)<br>Nrm. Acc.=0.81 (std = 0.12)<br>--<br>Correl.=0.71 (z std = 0.21)<br>Nrm. Correl.=0.84 (std = 0.13) | MAE=0.65 (std=0.28)<br><br>--<br>Correl.=0.78 (z std=0.74)<br>Nrm. Correl.=0.75 (std=0.74) | MAE=0.52 (std=2.02)<br><br>--<br>Correl.=0.63 (z std=0.9)<br>Nrm. Correl.=0.5 (std=1.21) | MAE=0.16 (std=0.14)<br><br>--<br>Correl.=0.96 (z std=0.75)<br>Nrm. Correl.=2.14 (std=9.15) |
| **Random lesion (60% removal)** | Acc.=64.35 (std = 7.24)<br>Nrm. Acc.=0.82 (std = 0.13)<br>--<br>Correl.=0.72 (z std = 0.20)<br>Nrm. Correl.=0.85 (std = 0.14) | MAE=0.65 (std=0.29)<br><br>--<br>Correl.=0.77 (z std=0.74)<br>Nrm. Correl.=0.73 (std=0.68) | MAE=0.48 (std=2.0)<br><br>--<br>Correl.=0.57 (z std=0.91)<br>Nrm. Correl.=0.41 (std=1.33) | MAE=0.16 (std=0.12)<br><br>--<br>Correl.=0.95 (z std=0.81)<br>Nrm. Correl.=1.96 (std=9.02) |
| **Random lesion (80% removal)** | Acc.=62.03 (std = 7.37)<br>Nrm. Acc.=0.79 (std = 0.12)<br>--<br>Correl.=0.71 (z std = 0.21)<br>Nrm. Correl.=0.84 (std = 0.12) | MAE=0.64 (std=0.26)<br>--<br>Correl.=0.75 (z std=0.72)<br>Nrm. Correl.=0.73 (std=0.74) | MAE=0.52 (std=2.01)<br>--<br>Correl.=0.62 (z std=0.93)<br>Nrm. Correl.=0.46 (std=1.19) | MAE=0.18 (std=0.15)<br>--<br>Correl.=0.96 (z std=0.76)<br>Nrm. Correl.=2.0 (std=9.09) |

**Table 6.** Robustness analysis. This table presents the results of an exploratory robustness analysis comparing different agent architectures informed by various data sources, including interviews, surveys, experiments, and summaries. All analyses are conducted on a random subsample of the same 100 agents. Performance is evaluated across four constructs: the General Social Survey (GSS), GSS Numeric (GSS Num.), Big Five Personality Traits (Big Five), and economic games. We use two 'hold-out' strategies for the GSS results when using GSS data in the input (indicated in the far left column): we either remove just the outcome question we are predicting on from the input, or we are removing the whole question block a given outcome question is in from the input. The latter ensures the input has few related questions as the outcome question, whereas the former will have related—but not identical—questions in input and output. For the big-5 and economic games, we always hold-out the whole question block a given outcome question is in. Results indicate that agents with interview data only perform well, achieving high normalized accuracy (0.82, std=0.13) on the GSS when just using the interviews. Maximal agents, which integrate data from all sources, show even better performance (0.87, std=0.13). Summary agents perform slightly below interview-based agents, with minor losses in accuracy (0.81, std = 0.13). A progressive decline is observed for random lesion agents as portions of interview data are removed, with accuracy dropping from 0.82 to 0.79 as 80% of the utterances are excluded, suggesting that even shortened interviews retain valuable insights.

0.0843

| | **Five social-science studies** |
|---|---|
| ***Human replication rate*** | Accuracy=0.48 (std=0.14)<br>Correl.=0.61 (z std=0.34) |
| ***Interview*** | Acc.=0.24 (std =0.15)<br>Nrm. Acc.=0.50 (std=0.12)<br><br>--<br>Correl.=0.34 (std=0.16)<br>Nrm. Correl.=0.56 (std=0.28)<br><br>–<br>Correlation with treatment effects: 0.93 |
| **Survey Agents**<br><br>(GSS+Big5) | Acc.=0.24 (std = 0.15)<br>Nrm. Acc.=0.52 (std = 0.11)<br><br>--<br>Correl.=0.33 (std = 0.14)<br>Nrm. Correl.=0.38 (std = 0.25)<br><br><br>--<br>Correlation with treatment effects: 0.91 |
| **Survey agents with interview**<br><br>(GSS+big5+interview) | Acc.=0.25 (std = 0.16)<br>Nrm. Acc.=0.53 (std = 0.12)<br><br>--<br>Correl.=0.34 (std = 0.15)<br>Nrm. Correl.=0.38 (std = 0.29)<br><br>--<br>Correlation with treatment effects: 0.99 |
| ***Demographic*** | Acc.=0.21 (std =0.16)<br>Nrm. Acc.=0.45 (std=0.14)<br><br>--<br>Correl.=0.31 (std=0.17)<br>Nrm. Correl.=0.32 (std=0.34)<br><br>–<br>Correlation with treatment effects: 0.93 |
| ***Persona*** | Acc.=0.23 (std =0.16)<br>Nrm. Acc.=0.48 (std=0.14)<br><br>--<br>Correl.=0.33 (std=0.17)<br>Nrm. Correl.=0.54 (std=0.35)<br><br>–<br>Correlation with treatment effects: 0.94 |

**Table 6 (continued).** Robustness analysis. This table presents the results of an exploratory robustness analysis comparing different agent architectures informed by various data sources. Performance is evaluated for the five social science experiments. All analyses are conducted on the full agent bank. The correlation and accuracy between predicted and actual treatment effects is calculated by comparing the vector of predicted treatment effects with actual treatment effects. The correlation between sample-level predicted and actual means is calculated by comparing the vector of the means for all outcomes in all five studies with the actual means. Accuracy is assessed on all the outcomes in these studies and so are correlations. Human replication rates are assessed by comparing the humans in wave 1 and wave 2.

# [Interview Script]

| Script | Time Limit (sec) |
| --- | --- |
| Hi, <participant’s name>! My name is Isabella, and I'm an AI assistant who will be conducting today's interview with you. Thank you so much for choosing to participate in our study!<br><br>Before we start, please take a moment to familiarize yourself with the controls on your screen. The button labeled 'Show Isabella's Subtitles' will display subtitles for my dialogue, while the pause button will take you back to the homepage that led you to this interview screen. Please note, however, that this interview is intended to be completed in one sitting. While it's okay to pause the interview if unexpected events occur or if you need a break, we recommend trying to complete the interview in one sitting.<br><br>Please also be aware that if you pause the interview, you will be asked to resume from the last saved checkpoint. You will know when you have passed a checkpoint by a flash of green light around the circle displaying my avatar, accompanied by a message that says 'Saved.' If we lose connection for more than a minute during the interview, a red button will appear, allowing you to refresh the page and resume from the last saved checkpoint.<br><br>This is a semi-structured interview expected to take roughly two hours. For completeness, I may ask a few questions that might seem repetitive along the way. As described in the consent form, this interview will cover your life experiences as well as your views on various social topics. The content of this interview will be a part of our dataset that will be shared with scientific researchers to better understand the lived experiences of those being interviewed.<br><br>Finally, you may speak when a microphone icon appears in the circle displaying my avatar. Please stop the interview now if any of this is not ok.<br><br>If all this sounds good, let’s get started! | 0 |
| To start, I would like to begin with a big question: tell me the story of your life. Start from the beginning -- from your childhood, to education, to family and relationships, and to any major life events you may have had. | 625 |
| Some people tell us that they've reached a crossroads at some points in their life where multiple paths were available, and their choice then made a significant difference in defining who they are. What about you? Was there a moment like that for you, and if so, could you tell me the whole story about that from start to finish? | 235 |
| Some people tell us they made a conscious choice or decision in moments like these, while others say it 'just happened'. What about you? | 50 |
| Do you think another person or an organization could have lent a helping hand during moments like this? | 40 |
| Moving to present time, tell me more about family who are important to you. Do you have a partner, or children? | 310 |
| Are there anyone else in your immediate family whom you have not mentioned? Who are they, and what is your relationship with them like? | 25 |
| Tell me about anyone else in your life we haven’t discussed (like friends or romantic partners). Are there people outside of your family who are important to you? | 80 |
| Now let’s talk about your current neighborhood. Tell me all about the neighborhood and area in which you are living now. | 105 |
| Some people say they feel really safe in their neighborhoods, others not so much. How about for you? | 40 |
| Living any place has its ups and downs. Tell me about what it’s been like for you living here. | 40 |
| Tell me about the people who live with you right now, even people who are staying here temporarily. | 80 |
| Is anyone in your household a temporary member, or is everyone a permanent member of the household? | 10 |
| Tell me about anyone else who stays here from time to time, if there is anyone like that | 10 |
| Is there anyone who usually lives here but is away – traveling, at school, or in a hospital? | 10 |

| Right now, across a typical week, how do your days vary? | 105 |
| --- | --- |
| At what kind of job or jobs do you work, and when do you work? | 105 |
| Do you have other routines or responsibilities that you did not already share? | 80 |
| Tell me about any recent changes to your daily routine. | 80 |
| If you have children, tell me about a typical weekday during the school year for your children. What is their daily routine? And what are after school activities your children participate in? | 80 |
| Some people we’ve talked to tell us about experiences with law enforcement. How about for you? | 80 |
| What other experiences with law enforcement stand out in your mind? | 40 |
| Some people tell us about experiences of arrest – of loved ones, family members, friends, or themselves. How about for you? | 40 |
| Some people say they vote in every election, some tell us they don’t vote at all. How about you? | 80 |
| How would you describe your political views? | 310 |
| Tell me about any recent changes in your political views. | 115 |
| One topic a lot of people have been talking about recently is race and/or racism and policing. Some tell us issues raised by the Black Lives Matter movement have affected them a lot, some say they’ve affected them somewhat, others say they haven’t affected them at all. How about for you? | 105 |
| How have you been thinking about the issues Black Lives Matter raises? | 105 |
| How have you been thinking about race in the U.S. recently? | 105 |
| How have you responded to the increased focus on race and/or racism and policing? Some people tell us they’re aware of the issue but keep their thoughts to themselves, others say they’ve talked to family and friends, others have joined protests. What about you? | 50 |
| What about people you’re close to? How have they responded to the increased focus on race and/or racism and policing? | 80 |
| Now we’d like to learn more about your health. First, tell me all about your health. | 80 |
| For you, what makes it easy or hard to stay healthy? | 65 |
| Tell me about anything big that has happened in the past two years related to your health: any medical diagnoses, flare-ups of chronic conditions, broken bones, pain – anything like that. | 80 |
| Sometimes, health problems get in the way. They can even affect people’s ability to work or care for their children. How about you? | 50 |
| Sometimes, it’s not your health problem, but the health of a loved one. Has this been an issue for you? | 80 |
| Tell me what it has been like trying to get the health care you or your immediate family need. Have you ever had to forgo getting the health care you need? | 80 |
| Have you or your immediate family ever used alternative forms of medicine? This might include indigenous, non-western, or informal forms of care. | 80 |
| During tough times, some people tell us they cope by smoking or drinking. How about for you? | 80 |
| Other people say they cope by relying on prescriptions, pain medications, marijuana, or other substances. How about for you and can you describe your most recent experience of using them, if any? | 80 |
| How are you and your family coping with, or paying for, your health care needs right now? | 80 |
| Some people are excited about medical vaccination, and others, not so much. How about you? | 50 |
| What are your trusted sources of information about the vaccine? | 80 |
| Now we’re going to talk a bit more about what life was like for you over the past year. Tell me all about how | 115 |

| you have been feeling. | |
|---|---|
| Tell me a story about a time in the last year when you were in a rough place or struggling emotionally. | 80 |
| Some people say they struggle with depression, anxiety, or something else like that. How about for you? | 80 |
| How has it been for your family? | 50 |
| Some people say that religion or spirituality is important in their lives, for others not so much. How about you? | 155 |
| Some people tell us they use Facebook, Instagram, or other social media to stay connected with the wider world. How about for you? Tell me all about how you use these types of platforms. | 115 |
| Some people say they use Facebook or Twitter to get (emotional or financial) support during tough times. What about you? | 50 |
| Tell me about any recent changes in your level of stress, worry, and your emotional coping strategies. | 50 |
| This might be repetitive, but before we go on to the next section, I want to quickly make sure I have this right: who are you living with right now? | 25 |
| Any babies or small children? | 10 |
| Anyone who usually lives here but is away – traveling, at school, or in a hospital? | 10 |
| Any lodgers, boarders, or employees who live here? | 10 |
| Who shares the responsibility for the rent, mortgage, or other household expenses? | 5 |
| Now we'd like to talk about how you make ends meet and what the monthly budget looks like for you and your family. What were your biggest expenses last month? | 50 |
| How much did your household spend in the past month? Is that the usual amount? And if not, how much do you usually spend and on what? | 50 |
| Does your household own or rent your home? | 50 |
| Bills can fluctuate over the course of a year. Seasons change, there are holidays and special events, school starts and ends, and so on. Tell me all about how your bills have fluctuated over the past year. | 50 |
| Tell me all about how you coped with any extra expenses in the past months. | 40 |
| Some people have a savings account, some people save in a different way, and some people say they don't save. How about you? | 80 |
| Some people save for big things, like a home, while others save for a rainy day. How about for you (in the last year)? | 80 |
| Some people have student loans or credit card debt. Others take out loans from family or friends or find other ways of borrowing money. Tell me about all the debts you're paying on right now. | 50 |
| Tell me about any time during the past year that you haven't had enough money to buy something that you needed or pay a bill that was due. | 65 |
| What is or was your occupation? In this occupation, what kind of work do you do and what are the most important activities or duties? | 155 |
| Was your occupation covered by a union or employee association contract? | 40 |
| In the past year, how many weeks did you work (for a few hours, including paid vacation, paid sick leave, and military service)? | 40 |
| How many hours each week did you usually work? | 15 |
| In the past week, did you work for pay (even for 1 hour) at a job or business? | 15 |
| Did you have more than one job or business, including part time, evening, or weekend work? | 25 |

| Now, in the past month, how many jobs did you work, including part time, evening, or weekend work? | 25 |
|---|---|
| For your job or occupations, how much money did you receive in the past month? | 50 |
| How often do you get paid? | 25 |
| In total, how much did your household make in the past month? | 50 |
| Switching gears a bit, let’s talk a little about tax time. Did you file taxes last year (the most recent year)? | 40 |
| Tell me more about your workplace. How long have you been at your current job? How would you describe the benefits that come with your job? (This could include things like health benefits, paid time off, vacation, and sick time.) | 15 |
| How would you describe your relationships at work? (How is your relationship with your manager or boss? How are your relationships with your coworkers?) | 15 |
| Tell me about how predictable your work schedule is. How would you describe your job in terms of flexibility with your hours? | 80 |
| Some people say it’s hard to take or keep a job because of child care. How about for you? | 15 |
| Do you receive any payments or benefits from SNAP, food stamps, housing voucher payments, supplemental security income, or any other programs like that? | 25 |
| Tell me about times over the last year when your income was especially low. And tell me about all the things you did to make ends meet during that time. | 50 |
| What would it be like for you if you had to spend $400 for an emergency? Would you have the money, and if not, how would you get it? | 50 |
| Overall, how do you feel about your financial situation? | 105 |
| Are you now married, widowed, divorced, separated, or have you never been married? If you are not currently married, are you currently living with a romantic partner? | 40 |
| For our records, we also need your date of birth. Could you please provide that? | 25 |
| Were you born in the United States? If you were not born in the U.S., what country were you born, and what year did you first come to the U.S. to live? | 25 |
| What city and state were you born in? | 25 |
| Are you of Hispanic, Latinx, or Spanish origin? | 15 |
| What race or races do you identify with? | 15 |
| What is the highest degree or grade you’ve completed? | 15 |
| Are you enrolled in school? | 15 |
| Have you been enrolled in school during the past 3 months? | 15 |
| Have you ever served on active duty in the U.S. Armed Forces, Reserves, or National Guard? | 15 |
| What religion do you identify with, if any? | 15 |
| Generally speaking, do you usually think of yourself as a Democrat, a Republican, an Independent, or what? And how strongly do you associate with that party? | 80 |
| Do you think of yourself as closer to the Democratic Party or to the Republican Party? | 50 |
| What was the city and state that you lived in when you were 16 years old? | 25 |
| Did you live with both your own mother and father when you were 16? | 25 |
| Who else did you live with? | 25 |
| What’s the highest degree or grade your dad completed? | 15 |

| What’s the highest degree or grade your mom completed? | 15 |
| --- | --- |
| Did your mom work for pay for at least a year while you were growing up? | 15 |
| What was her job? What kind of work did she do? What were her most important activities or duties? | 50 |
| What kind of place did she work for? What kind of business was it? What did they make or do where she worked? | 50 |
| Did your dad work for pay for at least a year while you were growing up? | 15 |
| What was his job? What kind of work did he do? What were his most important activities or duties? | 50 |
| What kind of place did he work for? What kind of business was it? What did they make or do where he worked? | 50 |
| We all have hopes about what our future will look like. Imagine yourself a few years from now. Maybe you want your life to be the same in some ways as it is now. Maybe you want it to be different in some ways. What do you hope for? | 155 |
| What do you value the most in your life? | 80 |
| And that was the last question I wanted to ask today. Thank you so much again for your time <participant’s name>. It was really wonderful getting to know you through this interview. Now, we will take you back to the Home Screen so that you may finish the rest of the study! | 0 |

**Table 7.** The interview script used to guide two-hour conversations with participants, adapted from the American Voices Project's interview. The script covers a wide range of topics, from participants' life stories to their views on social, political, and personal values. Select portions were abbreviated to ensure a manageable session length, while still capturing the breadth of experiences and perspectives essential for building nuanced generative agents.

| General Social Survey | Accuracy | Normalized Accuracy | Correlation | Normalized Correlation |
|---|---|---|---|---|
| ***Participant Replication*** | *79.53% (std=8.65)* | *1.00 (std=0.00)* | *0.87 (std=0.50)* | *1.00 (std=0.00)* |
| **Agents w/ Interview** | 65.67% (std=6.51) | 0.83 (std=0.11) | 0.74 (std=0.23) | 0.87 (std=0.15) |
| **Agents w/ Demog. Info.** | 58.12% (std=7.02) | 0.74 (std=0.12) | 0.65 (std=0.22) | 0.76 (std=0.16) |
| **Agents w/ Persona Desc.** | 56.21% (std=8.29) | 0.71 (std=0.12) | 0.64 (std=0.26) | 0.73 (std=0.19) |
| **Agents w/ Survey Resp.** | 64.98% (std=6.40) | 0.82 (std=0.11) | 0.71(std=0.25) | 0.83 (std=0.16) |
| **Agents w/ Survey + Interview** | 68.30% (std=6.73) | 0.86 (std=0.09) | 0.75 (std=0.24) | 0.87 (std=0.14) |

| Big Five | Mean Absolute Error | - | Correlation | Normalized Correlation |
|---|---|---|---|---|
| ***Participant Replication*** | *0.30 (std=0.17)* | - | *0.95 (std=0.76)* | *1.00 (std=0.00)* |
| **Agents w/ Interview** | 0.67 (std=0.27) | - | 0.78 (std=0.70) | 0.80 (std=1.88) |
| **Agents w/ Demog. Info.** | 0.68 (std=0.28) | - | 0.63 (std=0.70) | 0.61 (std=1.85) |
| **Agents w/ Persona Desc.** | 0.74 (std=0.32) | - | 0.71 (std=0.73) | 0.75 (std=2.59) |
| **Agents w/ Survey Resp.** | 0.62 (std=0.26) | - | 0.68 (std=0.71) | 0.65 (std=1.88) |
| **Agents w/ Survey + Interview** | 0.62 (std=0.25) | - | 0.77 (std=0.69) | 0.77 (std=1.79) |

| Economic Games | Mean Absolute Error | - | Correlation | Normalized Correlation |
|---|---|---|---|---|
| ***Participant Replication*** | *0.25 (std=0.88)* | - | *0.95 (std=1.00)* | *1.00 (std=0.00)* |
| **Agents w/ Interview** | 0.32 (std=0.89) | - | 0.64 (std=0.94) | 0.66 (std=2.83) |
| **Agents w/ Demog. Info.** | 0.33 (std=0.88) | - | 0.54 (std=0.91) | 0.47 (std=3.00) |
| **Agents w/ Persona Desc.** | 0.33 (std=0.88) | - | 0.60 (std=0.93) | 0.57 (std=2.84) |
| **Agents w/ Survey Resp.** | 0.34 (std=0.89) | - | 0.56 (std=0.92) | 0.38 (std=3.01) |
| **Agents w/ Survey + Interview** | 0.34 (std=0.88) | - | 0.54 (std=0.92) | 0.49 (std=2.79) |

**Table 8.** Generative agents' predictive performance. The consistency rate between participants and the predictive performance of generative agents is evaluated across various constructs and averaged across individuals. For the General Social Survey (GSS), accuracy is reported due to its categorical response types, while the Big Five personality traits and economic games report mean absolute error (MAE) due to their numerical response types. Correlation is reported for all constructs. Normalized accuracy is provided for all metrics, except for MAE, which cannot be calculated for individuals whose MAE is 0 (i.e., those who responded the same way in both phases). We find that generative agents predict participants' behavior and attitudes well, especially when compared to participants' own rate of internal consistency. Additionally, using interviews to inform agent behavior significantly improves the predictive performance of agents for both GSS and Big Five constructs, outperforming other commonly used methods in the literature.

| Dimension | Category | Sample Percentage | Quota Target |
|---|---|---|---|
| Age | 18–24 | 11.0% | 11.9% |
| Age | 25–34 | 13.9% | 17.8% |
| Age | 35–44 | 17.5% | 16.6% |
| Age | 45–54 | 19.8% | 16.3% |
| Age | 55–64 | 21.5% | 16.8% |
| Age | 65+ | 16.4% | 20.7% |
| Gender | Male | 43.4% | 49.1% |
| Gender | Female | 56.4% | 50.9% |
| Region | Northeast | 19.5% | 17.4% |
| Region | Midwest | 26.8% | 20.9% |
| Region | South | 30.2% | 38.0% |
| Region | West | 20.9% | 23.7% |
| Race/ethnicity | White/Caucasian | 79.2% | 74.5% |
| Race/ethnicity | Black/African American | 14.6% | 14.3% |
| Race/ethnicity | Asian | 5.0% | 6.9% |
| Race/ethnicity | Native Hawaiian or Pacific Islander | 0.5% | 0.4% |
| Race/ethnicity | American Indian or Alaskan Native | 2.8% | 1.9% |
| Race/ethnicity | Other race or ethnicity | 5.8% | 9.4% |
| Hispanic ethnicity | Hispanic | 16.9% | 18.4% |
| Hispanic ethnicity | Non-Hispanic | 83.1% | 81.6% |
| Education | Less than high school | 2.3% | 10.6% |
| Education | High school / some college, no degree | 38.9% | 45.6% |
| Education | Associate/junior college | 17.6% | 8.8% |
| Education | Bachelor's | 26.9% | 21.3% |
| Education | Graduate | 14.4% | 13.7% |

| Partisanship | Democrat | 35.3% | 27.0% |
| --- | --- | --- | --- |
| Partisanship | Republican | 26.4% | 27.0% |

**Table 9.** Quota targets and sample percentages for the experimental study based on vendor supplied quotas. Quota targets were derived by Bovitz from the U.S. Census Bureau, 2023 American Community Survey 5-Year Estimates